\documentclass[english, data-science]{agh-wi} 
\titlePL{Charakteryzacja struktur topologicznych w wybranych architekturach sieci neuronowych}  
\titleEN{Characterization of topological structures in different neural network architectures} 
\author{inż. Paweł Świder}
\supervisor{dr hab. inż. Marcin Kurdziel}
\usepackage{polski}                           

\usepackage{natbib}
\usepackage[listings=false]{scrhack}
\usepackage{amsmath}                          
\usepackage{amssymb}                          
\usepackage[polish, intoc]{nomencl}           
\usepackage{graphicx}                         
\usepackage{xcolor}                           
\usepackage{tabularx}                         
\usepackage{longtable}                        
\usepackage{caption}                          
\usepackage{hyperref}                         
\usepackage{subfig}
\usepackage{float}
\usepackage{cleveref}
\usepackage{rotating}
\usepackage{tabularray}
\usepackage{array}
\usepackage{multirow}
\usepackage{longtable}
\usepackage{booktabs}
\usepackage{pifont}
\usepackage{hyperref}

\definecolor{shadecolor}{gray}{0.9}
\hypersetup{
    colorlinks=true,
    linkcolor=blue,
    filecolor=magenta,
    urlcolor=cyan
}
\makenomenclature

\crefrangelabelformat{figure}{#3#1#4-#5#2#6}

\begin{document}
\frontmatter 
\maketitle 
\cleardoublepage
\thispagestyle{empty}
\vspace*{\fill}
\begin{flushright}
    \em
    \begin{minipage}{0.75\textwidth}
        We gratefully acknowledge Polish high-performance computing infrastructure PLGrid (HPC Center: ACK Cyfronet AGH) for providing computer facilities and support within computational grant no. PLG/2024/017160
    \end{minipage}
\end{flushright}
\begin{abstractPL}
Jednym z najważniejszych zadań w przyszłości będzie zrozumienie tego jak działają sieci neuronowe, ponieważ staną się one jeszcze potężniejsze i wszechobecne.
Niniejsza praca ma na celu wykorzystanie metod TDA do analizy reprezentacji neuronowych.
Opracowano w niej metody analizy reprezentacji dla różnych architektur oraz sprawdzono jak należy ich używać aby uzyskać prawidłowe wyniki.
Odkryte zależności wskazują, że usuwanie wartości odstających nie ma większego wpływu na wyniki oraz że powinno się porównywać reprezentacje o tej samej liczbie elementów.
Stworzone metody zostały użyte dla różnych architektur, jak ResNet, VGG19 i ViT, pomiędzy którymi znaleziono znaczące różnice, jak i pewne podobieństwa.
Dodatkowo ustalono, że modele o podobnej architekturze mają podobną topologię, a modele z większą liczbą warstw zmieniają swoją topologię bardziej płynnie. 
Co więcej, odkryto, że topologia pretrenowanych i finetunowanych modeli zaczyna się różnić w środkowej i końcowej warstwie, pozostając dość podobną w początkowych warstwach.
Odkrycia te dowodzą skuteczności TDA w analizie zachowania sieci neuronowych.
\end{abstractPL}

\begin{abstractEN}
One of the most crucial tasks in the future will be to understand what is going on in neural networks, as they will become even more powerful and widely deployed.
This work aims to use TDA methods to analyze neural representations.
    We develop methods for analyzing representations from different architectures and check how one should use them to obtain valid results.
Our findings indicate that removing outliers does not have much impact on the results and that we should compare representations with the same number of elements.
We applied these methods for ResNet, VGG19, and ViT architectures and found substantial differences along with some similarities.
Additionally, we determined that models with similar architecture tend to have a similar topology of representations and models with a larger number of layers change their topology more smoothly. 
Furthermore, we found that the topology of pre-trained and finetuned models starts to differ in the middle and final layers while remaining quite similar in the initial layers.
    These findings demonstrate the efficacy of TDA in the analysis of neural network behavior.

\end{abstractEN}
\tableofcontents   
\mainmatter 
\chapter{Introduction}
In recent years, neural networks have become very popular learning models and have found numerous applications. They are used, e.g., in image recognition, natural language processing, prediction tasks, medical diagnosis, or treatment planning. Researchers are constantly developing new ideas to improve neural network capabilities. These improvements have made neural networks more powerful and popular than ever before. That said, neural networks have important limitations. Firstly, they are vulnerable to overfitting.
Furthermore, neural models, especially large ones, require lots of data and computational power during training.  Another drawback of neural networks is that they are black-box models. That is, we do not understand how they make decisions or why they produce specific outputs. This translates to surprising phenomena, such as susceptibility to adversarial attacks.

Understanding the decision processes in neural networks is crucial for several reasons. When using neural networks in medicine, law, or other high-risk applications, we want to know why the trained model made a particular decision. Such transparency helps create trust and simplifies detecting model biases. Knowing the mechanism behind predictions also helps to improve the model and gives more information to the final user, which he can use to make better decisions. In fact, in many application domains, the law requires, or may soon require, that users of AI systems be familiar with the decision-making process.

Explaining deep neural networks is, therefore, a subject of vigorous research efforts.
One significant area of research concentrates on explainability, where we try to explain why the model made a specific prediction.
This class of methods is widely discussed in \citet{saleem2022explaining} and \citet{liang2021explaining}.
However, one drawback of this approach is that it only explains specific decisions and does not concentrate on finding the underlying general mechanisms responsible for decision-making. 

Another area of research focuses on analyzing neural network behavior from the perspective of neural representations.
We can treat neural representations as activation vectors of given network layers.
However, analysis of such vectors is challenging for several reasons.
One of the reasons is that two networks that solve the same task could potentially have totally different activation vectors, even if only from having different initial weights.
Therefore, dedicated methods were developed for comparing neural representations.
Examples of these methods are singular value canonical correlation analysis (SVCCA) \citep{raghu2017svcca} or centered kernel alignment (CKA) \citep{kornblith2019similarity}.
An alternative approach is to analyze the topology of neural representations.
To this end, methods of Topological Data Analysis (TDA) \citep{boissonnat2022topological} were used.
Some of the research on neural networks that used TDA tools concentrated on finding early stopping methods  \citep{watanabe2020deep} or detecting trojaned models \citep{gebhart2017adversary}.
That said, this area is relatively unexplored, as most research works with simple networks or with synthetic data.

Our goal in this thesis is to investigate and compare representations learned by selected neural architectures from the topology perspective. To this end, we will use TDA methods, specifically persistent homology \citep{edelsbrunner2013persistent}.
The implementation of this task consists of several research themes.
Firstly, we will develop methods for analyzing neural representations across different architectures.
We will also evaluate proposed approaches and identify their strengths and weaknesses in the investigation of neural network internals.
Next, we will conduct a detailed examination of representations in various layers within selected network architectures.
Our goal will be to identify how topology changes across different models and to identify similarities and differences between multiple architectures. Departing from some previous works \citep{naitzat2020topology} \citep{wheeler2021activation}, we will also abandon synthetic datasets and use real-world data.
By doing so, we can better understand how neural networks operate and what factors contribute to their performance.
Additionally, we will examine neural representations across different classes of input to see how they differ.
This will allow us to determine if neural models process distinct classes differently in the topological sense. Lastly, we will check how finetuning impacts the neural representations of selected network architectures.

By pursuing the research goals outlined above, we hope to find new methods for investigating how neural networks process data. We also want to uncover to what extent network architecture influences the topology of learned representations and, if it does, whether the uncovered differences correlate with important properties of the network. With this, we intend to increase our body of knowledge about internal representations of data in neural networks.

\chapter{Method}

This chapter introduces Topological Data Analysis methods that we use to examine representations learned by neural networks. We will operate on neural representations understood as activation coming from intermediate layers of the networks. This concept is defined in Equations \eqref{eq:network} and \eqref{eq:reprensetation}. 

\begin{eqnarray} \label{eq:network}
   F:&& \mathbb{R}^n \rightarrow \mathbb{R^d}, \\
   f_k:&& \mathbb{R}^{n_{k-1}} \rightarrow \mathbb{R}^{n_i}, \nonumber \\
   F_l:&& f_l \circ f_{l-1} \circ ... \circ f_2 \circ f_1. \nonumber
\end{eqnarray}
Equation \eqref{eq:network} defines a network $F$ that consists of multiple layers $f_i$. A subset of the network up to the $l-th$ layer, denoted by $F_l$ is the composition of network layers from 1 to $l$.

\begin{equation} \label{eq:reprensetation}
    x_i^l = (f_l \circ f_{l-1} \circ ... \circ f_2 \circ f_1)(x_l^0) = F_l(x_i^0).\\
\end{equation}
\Cref{eq:reprensetation} defines a neural activation $x^l_i$ as the activation vector (or volume, in convolutional networks) coming from $l$-th layer, given example $x^0_i$ on the network input. Note that by the index 0, we mean the original input: $x_i^0 = x_i$. Also, we will denote the original dataset by $X^0$, and its neural representation at layer $l$ by $X^l$.

In this work, we employ the Vietoris-Rips complex to analyze neural representations, which are an often-used construction from a family of so-called simplicial complexes \cite[Chapter~1 \S2]{munkres2018elements}.
The Vietoris-Rips complex, defined in \Cref{eq:vietoris} is a convenient way of forming a topological space from a set of points within a certain distance from each other.
The Vietoris-Rips complex consists of sets of points from $X^l$ that satisfy the condition that the distance ($d$) between any two points within the set is less than some arbitrarily selected cut-off ($\epsilon$).
We will call each set meeting this condition a simplex.
In our work, we will be using Euclidean distance, but in general each metric can be used. 
The index of $m$ in the definition determines the geometric interpretation of the simplex.
For instance, $m=0$ relates to points, $m=1$ to lines, $m=2$ to triangles, and for bigger $m$ we have $m$-dimensional simplexes. 
In real applications it is typically necessary to limit the maximum dimensionality of simplexes, in order to manage the computational cost of experiments.
Another reason for limiting simplexes' dimensionality is the difficulty of interpreting high-dimensional simplexes.

\begin{equation} \label{eq:vietoris}
    VR_{\epsilon}(X^l) := \{[x_0^l,...,x_m^l] : d(x_i^l, x_j^l) \leq 2\epsilon, x_0^l,...x_k^l \in X^l, m=0,1,...,n \}.
\end{equation}
Having defined Vietoris-Rips complexes, we will denote by $H_k(X^l)$ a corresponding $k$-th homology group [\citeauthor[Chapter~1 \S5]{munkres2018elements}]. Informally, $H_k(X^l)$ captures k-dimensional holes in the space created by the Vietoris-Rips complex \citep{vietoris_rips}. For example, $H_0(X^l)$ corresponds to connected components, $H_k(X^l)$ to loops, etc. The count of the k-dimensional holes in the constructed space is the so-called $k$-th  Betti number, formally defined as the rank of the homology group:
\begin{equation} \label{eq:betti}
    \beta_k(X) = rank(H_k(X)) \text{,  } k \in \mathbb{N}.
\end{equation}
Consequently, $\beta_0$ is the number of connected components, $\beta_1$ is the number of one-dimensional holes (i.e. loops), and $\beta_1$ counts the number of cavities (two-dimensional holes).

Another tool closely related to simplicial complexes is persistent homology. Its main idea is to investigate topological features uncovered by simplicial complexes across a range of distance cut-offs $\varepsilon$. Such an approach gives a dynamic, detailed view of how the shape of data changes across distance scales, and how different points connect. Consequently, we obtain more information about the data than is revealed by Betti numbers with fixed $\varepsilon$.

\begin{figure}[!ht]
    \centering
    \includegraphics[width=\linewidth]{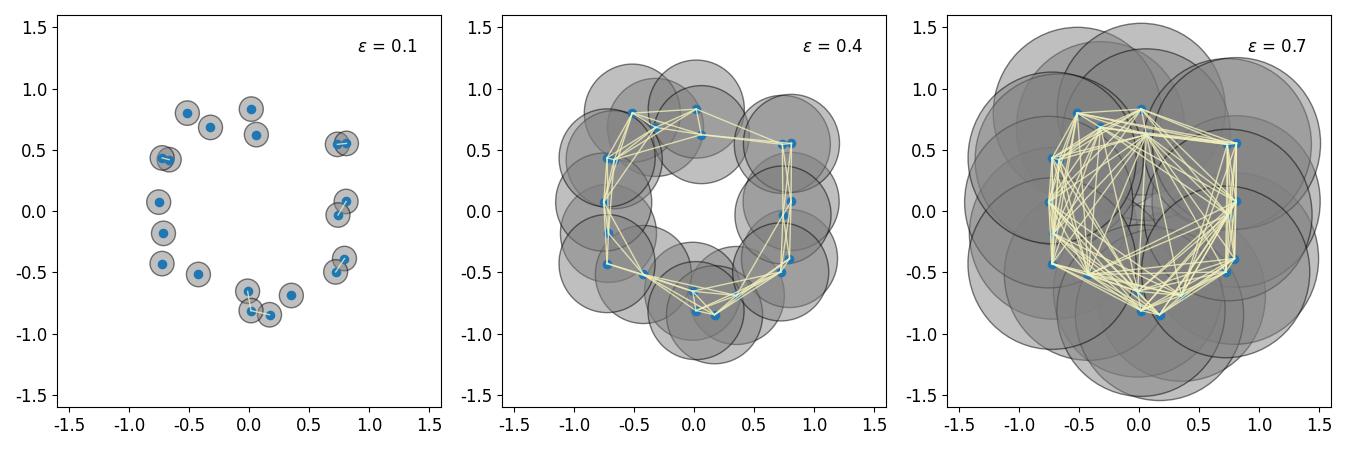}
    \caption{Vietoris-Rips complexes on toy data, with the different $\varepsilon$ value.}
    \label{fig:methods-complex}
\end{figure}
To better illustrate the persistent homology concept, we created a visualization demonstrating how the number of topological features in a toy dataset changes with increasing $\varepsilon$ in the Vietoris-Rips complex. The visualization, pictured in \Cref{fig:methods-complex}, displays three snapshots of the persistent homology. Circles mark distance cut-offs ($\varepsilon$) for points considered close to each other. Lines connect points that belong to the same Vietoris-Rips complex. Note that each of the three charts displays a different topological situation. On the first chart, with $\varepsilon = 0.1 $, we see points that, in most cases, are not connected to each other. Only some of them are close enough to other points to be in the same complex.
Consequently, we observe many connected components, mostly single points and no one-dimensional holes (loops). The situation gradually changes with increasing $\varepsilon$ values, and connected components disappear. The middle chart shows a topology with only one connected component. Furthermore, at this cut-off distance, data created a single $H_1$ topological feature (loop). This loop exists for only a limited cut-off span; it is not visible on the last chart, where cut-off $\varepsilon = 0.7$ removes all topological features. At this scale, we observe only one connected component containing all points.

\begin{figure}[!ht]
    \centering
    \includegraphics[width=0.95\linewidth]{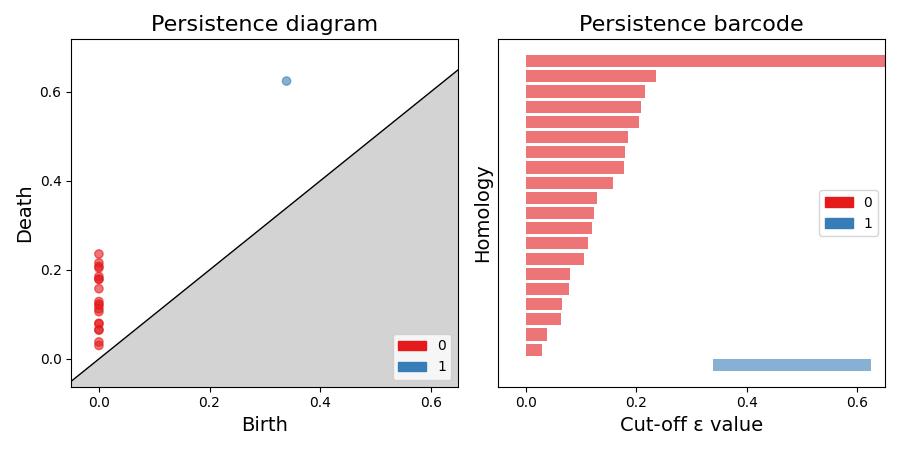}
    \caption{Persistence diagram and persistence barcodes for data from \Cref{fig:methods-complex}.}
    \label{fig:methods-barcode}
\end{figure}
Information uncovered by persistent homology can be visualized on so-called persistence diagrams or persistence barcodes.
These summaries of persistence homology have an advantage over analysis of specific snapshots in that they visualize topological features across the entire span of $\varepsilon$ values.
The persistence diagram and the persistent barcode for our toy data are presented in \Cref{fig:methods-barcode}.
The persistence diagram displays the life spans of homological features.
Specifically, each point on the diagram shows the features' time of birth ($\varepsilon$ value) on the x-axis and the time of death on the y-axis. In this construction, points are always located above the $x=y$ diagonal.
The further the point is from this line, the longer the corresponding homological feature lives, and the more significant it is.
Homological features with short lifetimes are often insignificant and can be a manifestation of the noise, rather than real topological features. On the diagram for our toy data, we see multiple features for $H_0$, most of them with a short lifetime. We also see one $H_1$ homological feature, representing a loop visible on \Cref{fig:methods-complex} for $\varepsilon = 0.4$. As we can see, persistent diagrams display substantial information about the data examined in a clear format. The diagram presented in \Cref{fig:methods-barcode}, right, displays the persistent barcodes for the same toy dataset. Barcodes display the same information as persistence diagrams but focus more on the points' lifetime.

\begin{figure}[!ht]
    \centering
    \includegraphics[width=0.9\linewidth]{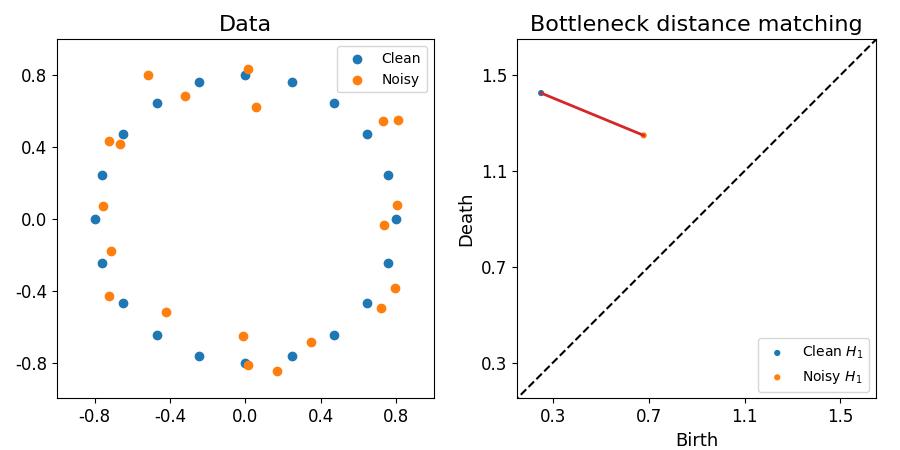}
    \caption{Two datasets with clean and noisy data along with persistence diagrams for $H_1$ homological features with bottleneck distance matching between them.}
    \label{fig:methods-bottleneck_matching}
\end{figure}
In this work, we will also need a method for comparing persistence diagrams and determining their differences. To this end, we will use the so-called bottleneck distance \cite[Chapter~9]{botnantopological}. The bottleneck distance is defined as a minimal distance to match every homological feature from one persistent diagram to a feature (or the diagonal line) from another persistent diagram. The matching (denoted by $\chi$) between two multisets is defined in \Cref{eq:matching}.
\begin{equation} \label{eq:matching}
    \chi := {\{(x_i^l, y_j^l) \mid x_i^l \in X^l, y_j^l \in Y^l, \text{ and each } x_i^l \text{ and } y_j^l \text{ occur in at most one pair}} \}.
\end{equation}
For each matching, we can calculate the cost of it. The cost of the matching is defined as the maximum distance between two points of the matching or the distance of the unmatched points to the diagonal line. This matching cost is formalized in \Cref{eq:cost}
\begin{equation} \label{eq:cost}
    c(\chi) := max \left( \sup_{(x_i^l, y_j^l) \in \chi} d(x_i^l, y_j^l), \sup_{ \text{unmatched } x_i^l \in X^l \cup Y^l} d(x_i^l, x=y) \right).
\end{equation}
Having defined matching and its cost, we can define a bottleneck distance as the matching with minimal cost. The definition of the distance is presented in \Cref{eq:bottleneck}.

\begin{equation} \label{eq:bottleneck}
    d_B(X^l, Y^l) := \inf \{ c(\chi) : \chi \text{ is a matching between } X^l \text{ and } Y^l \}.
\end{equation}
What is worth mentioning, bottleneck distance has some interesting properties.
The bottleneck distance is metric and is stable. The stability of this metric manifests in resilience to small changes in data or small noise.
In particular, this metric is sensitive only to maximum matching, and changes in data that do not change maximal matching leave the distance identical. 
That said, the bottleneck distance has one important limitation: it is not stable against adding extra elements to the set of data points (e.g., an extra point in the middle of a loop).
Consequently, outliers can have a substantial impact on this metric. \Cref{fig:methods-bottleneck_matching} presents a visualization of the bottleneck distance and the bottleneck matching.
We have two datasets in this example: the toy dataset from previous illustrations (the "Noisy" dataset), and its denoised variant (the "Clean" dataset).
We constructed persistent homology for each dataset and then calculated the bottleneck distance between them. The bottleneck matching between two persistence diagrams is visualized on the plot on the right. We see a line that matches only the "Noisy" $H_1$ homology group with the corresponding homology group from the "Clean" dataset.
For more information about this property, see \citep[Chapter~9.1.1]{botnantopological}.

Concluding this chapter, we have presented several TDA methods and tools that we will use in the upcoming chapters.
Specifically, in our work on neural representations we will heavily use Vietoris-Rips complexes and bottleneck distances. We will not pay much attention to Betti numbers, as they give concise but static information about the data. Persistent homology, on the other hand, provides a more complex view of the structure in neural representations. It is worth mentioning that other TDA methods could possibly also be used to analyze neural representations. One prominent example is the Mapper algorithm, which creates a graph based on the shape of data. That said, we do not explore this alternative in our work.

\chapter{Proposed experiments}

In this chapter, we describe the experiments that support the main result of this thesis. 
The results of these experiments will be presented in later chapters. 
Our experiments will analyze the homology of neural representations learned by different network architectures. 
Thus, our first goal is to identify methods that can reveal significant homological features. 
Next, we will apply these methods to selected, commonly used neural network architectures in order to find differences between representations learned by these architectures, as well as differences between models' layers and input classes.
To manage the computational cost of experiments, we narrowed this study to the image classification task using a standard dataset, namely, CIFAR100. 

The methods considered in our work should be architecture agnostic.
That way, they can be used changelessly across various popular architectures, like plain convolutional networks (e.g., VGG19), residual networks (e.g., ResNet18 and ResNet50), and Vision Transformers (ViT).
We will collect and process neural activations independently for each class, treating the neural representation of each input image as a single element for persistent homology.
For each architecture, we will collect neural representations from selected layers distributed across the network depth.
Generally, we want to examine activations from the initial, middle, and final layers of the network.
Activations collected from different layers give us a broader picture of the whole network.
To obtain neural representation with a size reasonable for subsequent processing, we try to keep input images at original resolutions when possible, and decided to increase resolution only when investigating a pre-trained model, or if the model requires bigger images (mainly ViT).
To make activations coming from various models and layers scale-invariant, we decided to normalize the activation vector of each example to zero mean and unit variance.

In experiments with VGG19 architecture we calculate neural activations directly after the selected convolution operator and before the activation function. 
We will denote activation from the $n$-th convolutional layer (counting from the network input) as \texttt{Conv n}. 
Another family of models analyzed in this work is the residual networks. 
They comprise several residual blocks arranged into \emph{stages}.
Different ResNet networks have different numbers of blocks, and blocks can vary in a number of parameters in convolution operators.
That said, each ResNet typically consists of four stages.
We will collect ResNet activations at the output of selected residual blocks and denote activation from the $n$-th stage and the $k$-th block as \texttt{Stage n Block k}.
To reduce the size of the intermediate representation and handle bigger images, the original ResNet architecture begins with a two-pixel stride convolution operator, followed by a max pooling operator.
We will remove these layers when training ResNet on small images, as otherwise they have a massive impact on the model accuracy.
With ViT we will use a slightly different approach to collecting neural representation.
Vision Transformer consists of a standard Transformer encoder with multiple transformer blocks. 
We will denote the $n$-th block as \texttt{Block n}.
As the neural representation, we will take an extra learnable class token included at the end of encoding blocks.
Because Vision Transformer requires a large dataset for training, we will use pre-trained ViT models and only finetune them on our dataset.
Also, with Vision Transformer we will always increase the resolution of our images as required by this architecture.

We split the experiments into two independent parts. The first part consists of experiments that focus on analyzing topological methods themselves, rather than neural networks.
We perform these experiments to check which factors have an impact on persistent homology, and which have no discernible effect.
The second part concentrates on analyzing neural representations in various neural network architectures.

\section{How we can use TDA for neural reprensetations}
\label{points-impact}
Our goal in the first set of experiments is to explore how our proposed methods behave in the analysis of neural representations, and determine the strengths and weaknesses of persistent homology in this task.

\paragraph{How does the number of points impact persistent homology}  ~\\
In the first experiment, we will check how does the number of inputs from which we collect neural representation impact the final results.
We would like to find the minimum number of points needed to obtain stable results.
The consistency of the outcomes would then imply that even when a larger number of elements is included in a single neural activation batch, the results remain relatively unchanged.
This experiment will also tell us if we can compare two persistent homologies that we created from different numbers of elements.
Analyzing different numbers of inputs is essential to compare neural activations from test and train datasets.
Specifically, we must consider that in CIFAR100 we have, for each class, 500 examples from the training set and 100 examples from the test set.
Therefore, finding a sufficient number of examples is crucial for obtaining reliable results without unnecessarily increasing the computational cost of the experiment.

In this experiment, we will first collect neural activation independently for each class, using increasing subsets of the available class members.
Next, we will compute Vietoris-Rips complexes for each subset.
Finally, we will compare them to complexes created with all elements belonging to the class.
We will perform these comparisons using the bottleneck distance and the number of points on persistence diagrams.
We anticipate discovering a threshold value whose corresponding persistence diagram will exhibit a minimal bottleneck distance to the baseline.
In possession of this threshold, we will be able to operate on a minimal number of inputs from the dataset.

\paragraph{Impact of outliers on persistence homology}  ~\\
The second experiment checks whether it is important to remove outliers from neural representations before calculating persistent homology.
We use the outlier detection step as part of the neural activation processing pipeline, introduced by \citet{naitzat2020topology}.
For the outlier detection algorithm, we use the Local Outlier Factor (LOF) proposed by \citet{breunig2000lof}.
In our case, with real-world data, we can benefit in many ways from outlier removal.
Firstly, we will reduce the size of our dataset, thereby making computation faster.
Secondly, outliers can impact the detection of topological features, and removing them can give more reliable results.
For example, outliers from long-living connected components, may spuriously impact bottleneck distance between diagrams. 

The experiment calculates persistent diagrams for each class with and without the LOF step and then compares these diagrams visually.
Next, we will construct two persistent diagrams for each class using disjoint subsets of inputs and then calculate all pairwise bottleneck distances between diagrams.
We expect that the bottleneck distance will be smaller for diagrams coming from the same class, and that after applying LOF we will obtain similar results with lower variance. 

With these two experiments, we will get more insights into how persistent homology performs in analyzing neural representations, and this knowledge will help us plan and analyze further experiments.

\section{Experiments on topology of neural representations}\label{resnet-cnn}

The second set of experiments consists of analyzing representation in selected neural network architectures.
Here, we focus entirely on topological differences between various architectures and models, as well as between selected layers in those networks.
For example, we check how topology changes in the initial, middle, and final layers of the investigated models.

\paragraph{Topological characterization of selected architectures} ~\\
This experiment focuses on analyzing basic statistics of neural representation.
We compute persistence diagrams for different networks and layers and compare corresponding diagrams to each other.
To this end, we compute basic statistics for persistence diagrams, such as the average number of topological features for every homology group and the average birth, death, and life time.
We also compute standard deviations for the estimated statistics.
These experiments are run independently for each class.
Later, we check how similar the persistence diagrams are by computing histograms of bottleneck distances between them.
We will also check for class clusters.
To this end, we will use matrices of bottleneck distances between persistence diagrams and embed those distances in $\mathbb{R}^2$ using the UMAP algorithm \citep{mcinnes2020umap}.
In principle, any other algorithm for embedding distance matrices could also be suitable.

In another set of experiments, we also check if there are differences between representations of train and test datasets.
To do so, we compute the characteristics outlined above for the test dataset, a subset of the training dataset with the matching number of inputs and all training data. 
This analysis is also partitioned independently for each class.

\paragraph{Effects of finetuning on topology of representations}  ~\\
In this experiment, we investigate differences between models with identical architecture.
We examine three model instances: networks with random weights, networks pre-trained on different data, and networks pre-trained on different data and finetuned on CIFAR100.
Again, we will compare these models in class-conditional settings.
To this end, we will first look into persistence diagrams to find topological differences between their representations.
Next, we will compute bottleneck distances between all persistence diagrams to obtain the distances matrix.
Finally, similar to the previous experiment we will visualize the distribution of these diagrams using the UMAP algorithm.

In this experiment, models should form clusters on the UMAP embedding if their neural activations have similar topological features.
We hypothesize that persistence diagrams corresponding to models with random weights will form a cluster separated from other models, and that differences between pre-trained and finetuned models will be smaller than their difference from the random-weights models.

\paragraph{Where networks change homology most rapidly} ~\\
In the last experiment, we will compare persistence diagrams computed at, respectively, input and output at selected network layers. 
In particular, we will calculate bottleneck distances between diagrams obtained from layers' inputs and their corresponding outputs.
Alternatively, distances can also be computed between persistent diagrams of inputs to any chosen network layers at the persistent diagram of output from the last layer before the classification head.
For example, in ViT we compute bottleneck distance taking neural representation from the input of \texttt{Block 2} and output of \texttt{Block 7}.
We averaged bottleneck distances across classes and plotted a heatmap that shows how similar complexes are between the model's layers.

This experiment will show which layers change persistence homology most and least rapidly.
It will also verify whether bottleneck distance increases when comparing neural representations from distant layers.
We expect this to be the case.
Lastly, we aim to find similarities and differences between various architectures with respect to topology evolution.

\chapter{Results of experiments}

In this chapter we will present the results of our experiments. 
The first two sections report the results of experiments that aim to study our methodology. 
We determine whether we should use the same number of points when topologically comparing two neural representations.
Furthermore, we clarify the impact of outliers on persistent homology and uncover how many of these outliers are in neural representations. 
The results of these experiments will give us valuable insights into how we should design the next ones.

The results of subsequent experiments, reported in the last three sections, are intended to identify the key differences and commonalities between various models and architectures.
We explore differences between the topology of neural representations coming from the train and test datasets.
Additionally, we analyze how finetuning impacts topology and uncover differences between the finetuned model, pre-trained, and model with random weights.
Furthermore, we explore how neural networks change the topology of the data and in which layers the topology changes most rapidly.
We anticipated that the results of these experiments would provide valuable insights into neural representations and the manner in which the model evolves them.

In terms of technical details, we perform our experiments using the PyTorch library to collect neural representations and train models. 
To compute persistence diagrams we use the Ripser$++$ library \citep{zhang2020gpu}, a GPU-enabled software for computing Vietoris-Rips persistence barcodes.
For other topology-related operations we will use the GUDHI library \citep{gudhi:BottleneckDistance}. 
The GUDHI library implements numerous algorithms that we use in this work, including the bottleneck distance. 
All the GPU-related computations (training models, calculating Vietoris-Rips persistence barcodes) were performed on Athena, a supercomputer at AGH Cyfronet equipped with an NVIDIA A100 40GB GPUs. 
The source code for training and executing experiments is available in the GitHub repository. One can access it via: \href{https://github.com/pawlo555/tda-dl}{https://github.com/pawlo555/tda-dl}.

\newpage
\section{How does the number of points impacts persistent homology}
\label{sec:number_homologies}

\begin{figure}[!ht]
    \centering
    \subfloat[\texttt{Conv 4}, $H_2$]{
        \label{fig:optimal_points-points_vgg_h2_conv4}
        \includegraphics[width=0.308\linewidth]{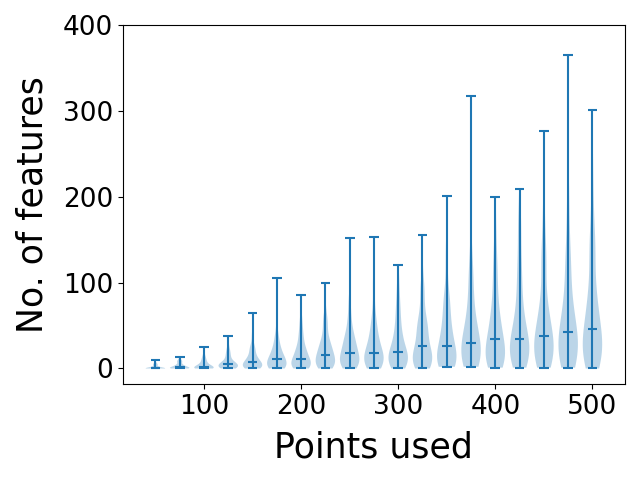}
    }
    \subfloat[\texttt{Conv 12}, $H_0$]{
        \label{fig:optimal_points-points_vgg_h0_conv12}
        \includegraphics[width=0.308\linewidth]{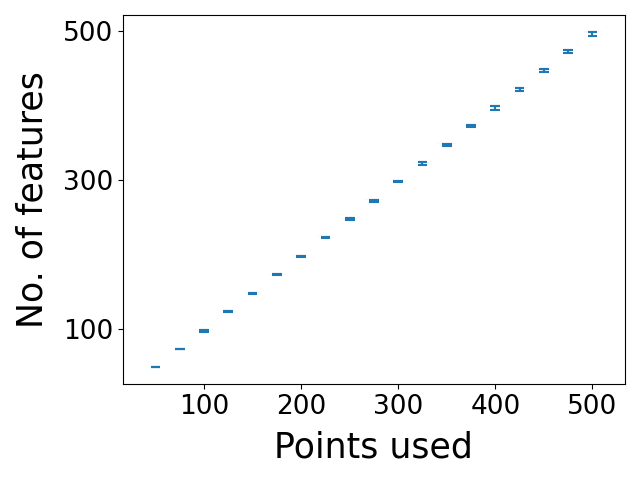}
    }
    \subfloat[\texttt{Conv 12}, $H_1$]{
        \label{fig:optimal_points-points_vgg_h1_conv12}
        \includegraphics[width=0.308\linewidth]{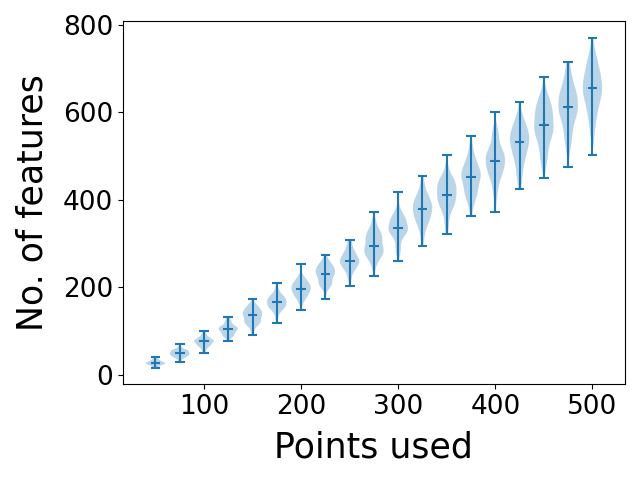}
    }
    \\
    \subfloat[\texttt{Conv 4}, $H_2$]{
        \label{fig:optimal_points-homology_vgg_h2_conv4}
        \includegraphics[width=0.308\linewidth]{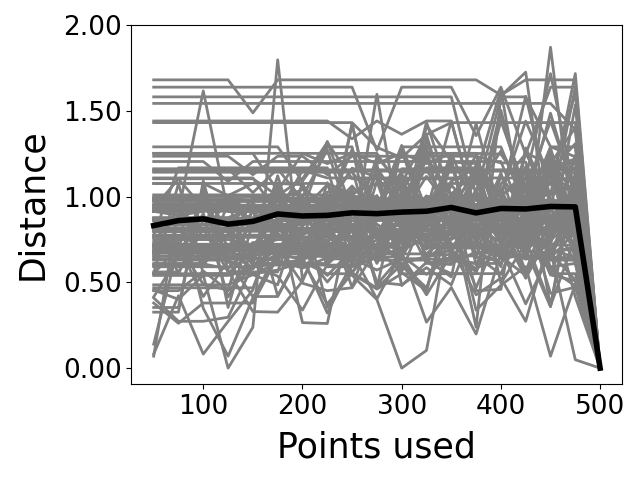}
    }
    \subfloat[\texttt{Conv 12}, $H_0$]{
        \label{fig:optimal_points-homology_vgg_h0_conv12}
        \includegraphics[width=0.308\linewidth]{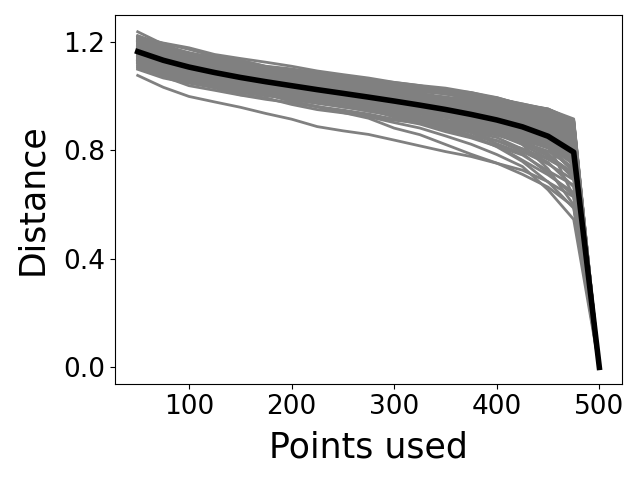}
    }
    \subfloat[\texttt{Conv 12}, $H_1$]{
        \label{fig:optimal_points-homology_vgg_h1_conv12}
        \includegraphics[width=0.308\linewidth]{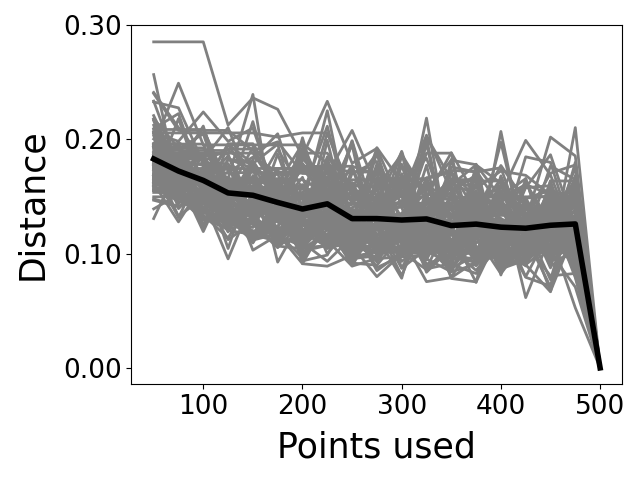}
    }
    \\
    \subfloat[\texttt{Conv 12}, $H_2$]{
        \label{fig:optimal_points-points_vgg_h2_conv12}
        \includegraphics[width=0.308\linewidth]{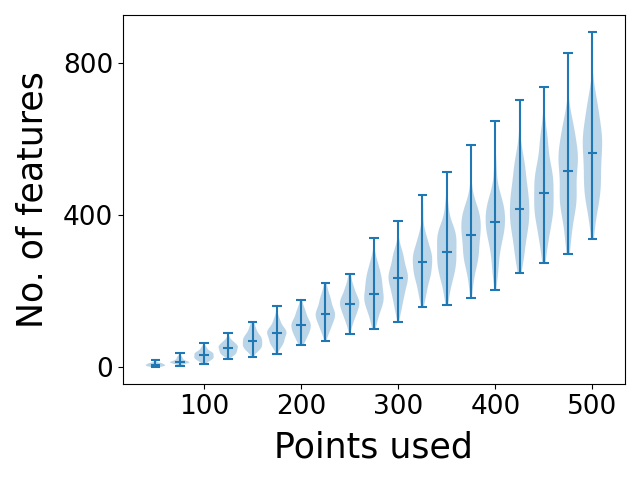}
    }
    \subfloat[\texttt{Conv 12}, $H_2$]{
        \label{fig:optimal_points-homology_vgg_h2_conv12}
        \includegraphics[width=0.308\linewidth]{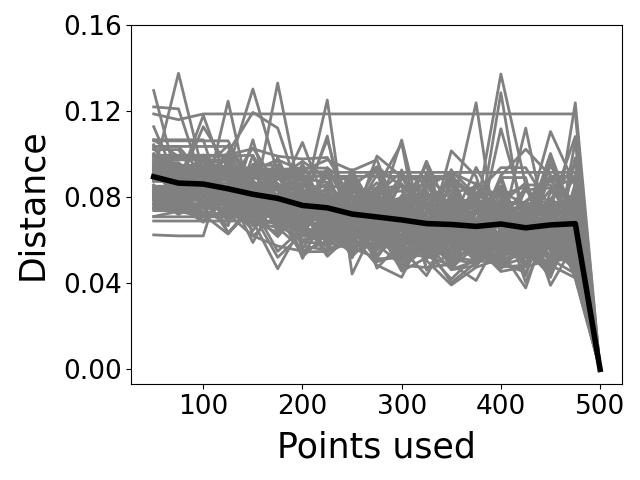}
    }
  \caption{\centering Violin plots of the homological features along with bottleneck distances calculated between diagrams obtained from a subset of points with all available for VGG19.}
  \label{fig:optimal_points-vgg19}
\end{figure}

In this experiment we aim to determine how the number of points used to create persistence diagrams impacts results.
We experimented with the VGG19 network, across a range of random subsets of data, namely, from 50 to 500 inputs, at every 25-point interval.
Next, we counted points on persistent diagrams for each class and each number of inputs taken.
We grouped the results by the number of inputs taken, and plotted their distribution using a violin plot.
Representative results from this experiment are presented on \Cref{fig:optimal_points-points_vgg_h2_conv4,fig:optimal_points-points_vgg_h0_conv12,fig:optimal_points-points_vgg_h1_conv12,fig:optimal_points-points_vgg_h2_conv12}.
Additionally, for each class we computed the bottleneck distances between baseline homology, calculated using 500 points, and homologies computed using subsets described above.
We aggregate these results on \Cref{fig:optimal_points-homology_vgg_h2_conv4,fig:optimal_points-homology_vgg_h0_conv12,fig:optimal_points-homology_vgg_h1_conv12,fig:optimal_points-homology_vgg_h2_conv12}. The charts show the distances for each class together with respective average values. The individual distances were plotted with grey lines, and the average distance was plotted using a black line.

When observing the number of points on persistent diagrams, we notice that it correlates strongly with the number of inputs used to calculate persistent homology.
The number of points generally increases linearly or squarely with the number of inputs used. 
We observed a strong linear correlation for $H_0$ homology (Figure \ref{fig:optimal_points-points_vgg_h0_conv12}).
This correlation occurs because for small $\varepsilon$ cut-offs the number of connected components equals the number of elements in the dataset, unless there are duplicated points.
Nevertheless, connected components aren't engaging in this experiment, and we will not concentrate on this homology group.
Instead, we focus on $H_1$ and $H_2$ groups, and observe two trends for these homological features.
We observe that for 500 inputs the average number of homological features is similar to the number of features with 200 or fewer inputs (\ref{fig:optimal_points-points_vgg_h2_conv4}).
We also notice that for some classes the number of features is considerably larger than the average.
This phenomenon is evident in the initial and intermediate layers.
As the network progresses, we observe different trends (\Cref{fig:optimal_points-points_vgg_h1_conv12,fig:optimal_points-points_vgg_h2_conv12}).
We notice that when using more elements the number of homological features drastically increases for the last layers.

\Cref{fig:optimal_points-homology_vgg_h2_conv4,fig:optimal_points-homology_vgg_h0_conv12,fig:optimal_points-homology_vgg_h1_conv12,fig:optimal_points-homology_vgg_h2_conv12} show bottleneck distances. 
We observe that bottleneck distance remains relatively constant when using almost all possible inputs, except the last one. 
The distance equals 0 when we use 500 inputs, because we compare two identical persistence diagrams. 
For most classes bottleneck distances fluctuate up and down with an increasing number of inputs, but for some it remains at the same value regardless of the number of inputs used. 
We anticipated that the bottleneck distance would decline when using more inputs, but it even increases (\Cref{fig:optimal_points-homology_vgg_h2_conv4}).
On the other hand, we observe a stable decline for \Cref{fig:optimal_points-homology_vgg_h0_conv12}.
However, the $H_0$ homology group on that plot isn't as interesting as other, higher groups.
The results for $H_1$ and $H_2$ are reported on \Cref{fig:optimal_points-homology_vgg_h1_conv12,fig:optimal_points-homology_vgg_h2_conv12}.
Upon analyzing these charts, we noticed that the initial value of the bottleneck distance slightly decreases when we add more inputs. 
This trend continues until the number of points reaches approximately 250, after which the bottleneck distance remains constant without any significant change.
We also noticed that bottleneck distance could increase when the number of points on persistent diagrams remains stable and decline after adding more network inputs, when the number of points on the diagrams increases.

\begin{figure}[!ht]
    \centering
    \subfloat[\texttt{Stage 1}, $H_2$]{
        \label{fig:optimal_points-points_resnet_h2_stage1}
        \includegraphics[width=0.308\linewidth]{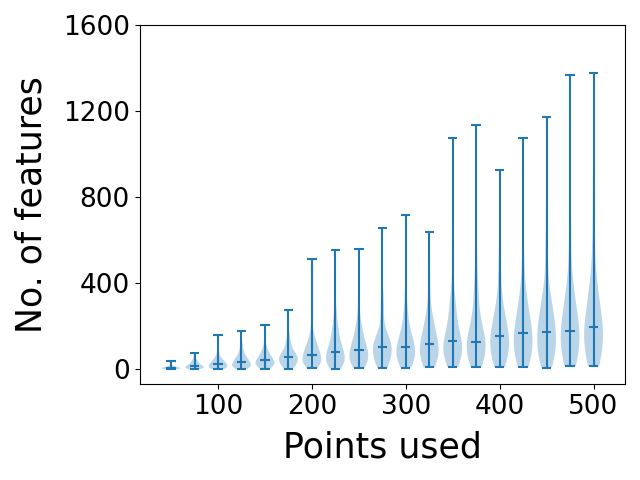}
    }
    \subfloat[\texttt{Stage 4}, $H_2$]{
        \label{fig:optimal_points-points_resnet_h2_stage4}
        \includegraphics[width=0.308\linewidth]{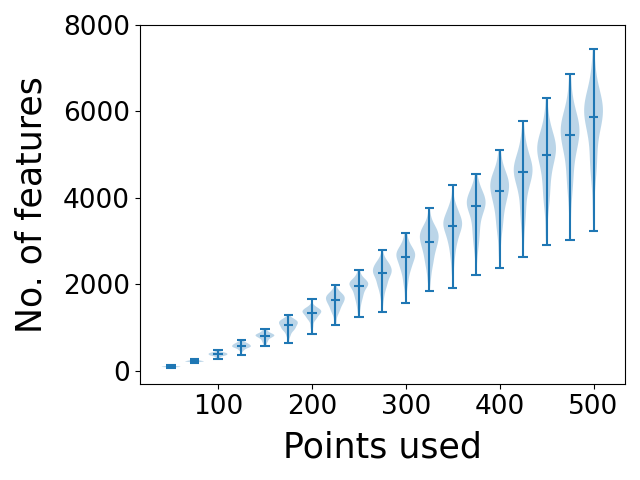}
    }
    \\
    \subfloat[\texttt{Stage 1}, $H_2$]{
        \label{fig:optimal_points-homology_resnet_h2_stage1}
        \includegraphics[width=0.308\linewidth]{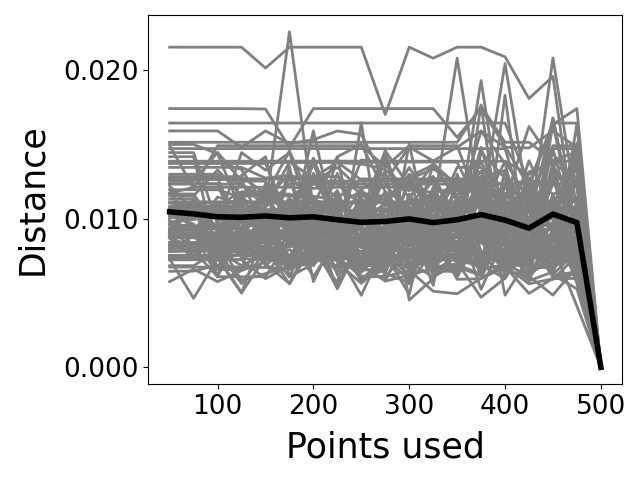}
    }
    \subfloat[\texttt{Stage 4}, $H_2$]{
        \label{fig:optimal_points-homology_resnet_h2_stage4}
        \includegraphics[width=0.308\linewidth]{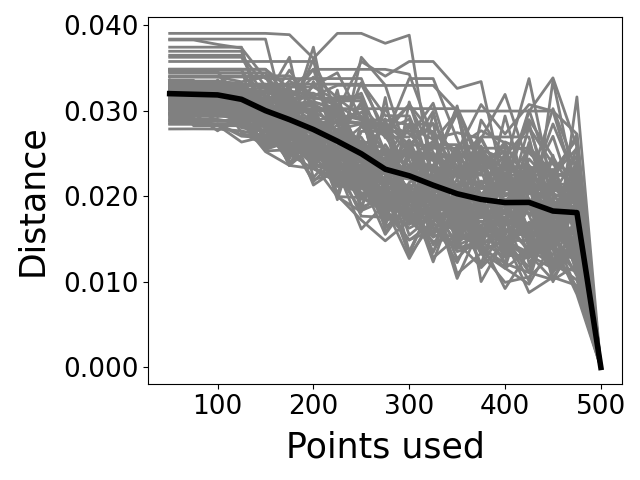}
    }
  \caption{\centering Violin plots of the homological features along with bottleneck distances calculated between diagrams obtained from a subset of points with all available for ResNet18.}
  \label{fig:optimal_points-resnet18}
\end{figure}
We conducted a parallel experiment using ResNet18, and some representative charts are reported in \Cref{fig:optimal_points-resnet18}. 
For the residual network, we observed behavior similar to that seen with VGG19. 
For \Cref{fig:optimal_points-points_resnet_h2_stage1,fig:optimal_points-homology_resnet_h2_stage1}, that correspond to the beginning of the network (\texttt{Stage 1}), we observe that the number of points on persistent diagrams is similar for all subsets. 
Still, for some classes it is significantly bigger than typical distances.
This keeps the average bottleneck distance at the same level as for VGG19.
For the final network layers (\texttt{Stage 4}), reported in \Cref{fig:optimal_points-points_resnet_h2_stage4,fig:optimal_points-homology_resnet_h2_stage4}, there is a significant increase in points present on the persistence diagrams.
There is also a decline in the bottleneck distance values.
This decline was also visible in VGG19. 

The results obtained in this experiment indicate similarities trendy in between different models.
An important observation is that the number of inputs in the dataset has a lesser impact on the results when data come from initial layers. 
To see this compare results from \texttt{Conv 4} or \texttt{Stage 1} with those from \texttt{Conv 12} or \texttt{Stage 4}.
Therefore, we should be very careful when comparing two datasets with different numbers of points, even if they come from the same distribution.
Also, even with small changes in the persistent homology the bottleneck distance increases significantly.
This also should be taken into account in the subsequent experiments.

\newpage
\section{Impact of outliers on persistence homology}
\label{sec:lof}
\begin{figure}[!ht]
\centering
    \subfloat[LOF, bear]{
        \label{fig:lof_vgg19_bear}
        \includegraphics[width=0.22\linewidth]{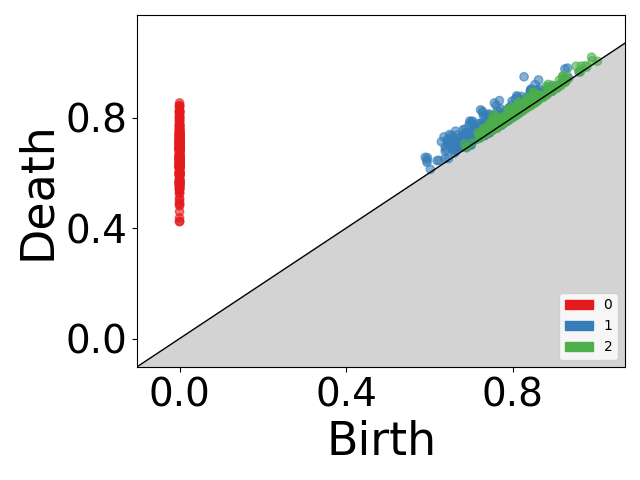}
    }
    \subfloat[no LOF, bear]{
        \label{fig:nolof_vgg19_bear}
        \includegraphics[width=0.22\linewidth]{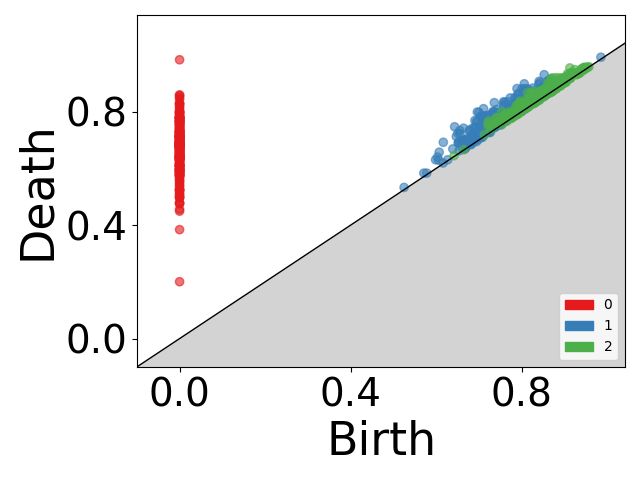}
    }
    \subfloat[LOF, beaver]{
        \label{fig:lof_vgg19_beaver}
        \includegraphics[width=0.22\linewidth]{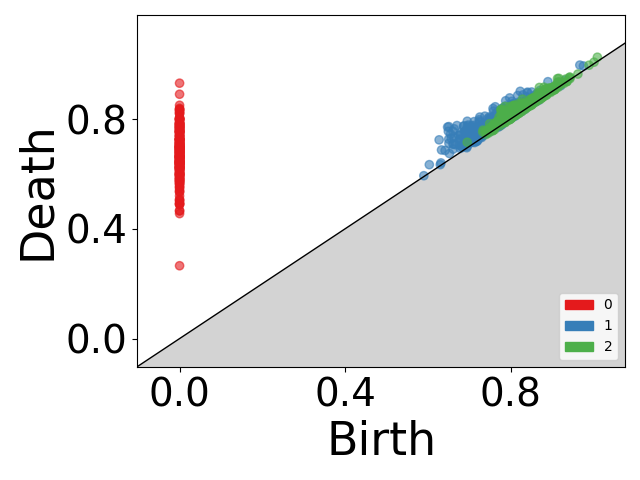}
    }
    \subfloat[no LOF, beaver]{
        \label{fig:nolof_vgg19_beaver}
        \includegraphics[width=0.22\linewidth]{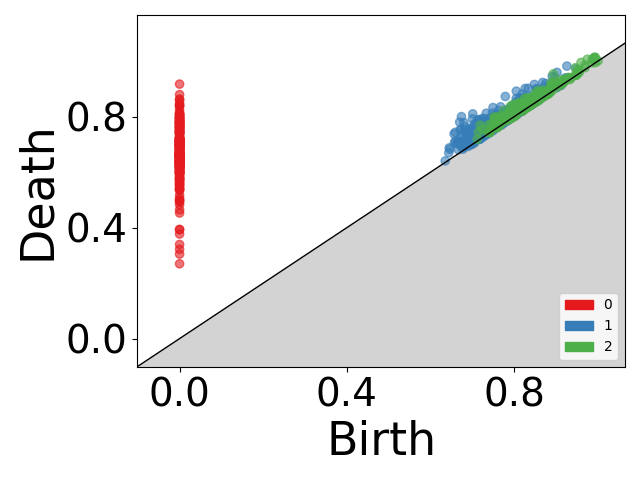}
    }

  \caption{\centering Persistence diagrams for VGG19 \texttt{Conv12} layer and selected classes, with and without LOF.}
  \label{fig:lof}
\end{figure}

The following experiments aim to find the utility of filtering out outliers (using LOF) before computing Vietoris-Rips complexes. 
Illustrative examples of persistence diagrams with and without LOF are displayed in \Cref{fig:lof}. 
We generate diagrams for representations coming from \texttt{Conv 12} layer in VGG19.
However, we obtained similar results also in different layers and network architectures. 
In general, removing local outliers changes persistent diagrams slightly, especially for $H_1$ and $H_2$ homology groups. 
One of the differences in many persistence diagrams appears when we have an element that is further from other components. 
In this case, without LOF we observe a single $H_0$ feature located above other features. 
We don't see these points when we use LOF, as elements responsible for it are outliers, and are therefore removed by the LOF algorithm (\Cref{fig:lof_vgg19_bear,fig:nolof_vgg19_bear}).
Specifically, we see one point for homology $H_0$ with the death point larger than the rest (\Cref{fig:nolof_vgg19_bear}). 
We didn't see such points in the persistent diagram generated from neural activation processed with LOF (\Cref{fig:lof_vgg19_bear}).
That said, persistence diagrams with and without LOF are similar.

\begin{table}[!ht]
\centering
\caption{ \begin{centering} Averaged bottleneck distances with and without LOF for various layers of VGG19 network.\end{centering}}
\label{tab:lof-vgg19}
\begin{tabular}{cccccccccc}
\toprule   
\multicolumn{2}{c}{VGG19} & \multicolumn{2}{c}{\texttt{Conv 4}} & \multicolumn{2}{c}{\texttt{Conv 8}} &  \multicolumn{2}{c}{\texttt{Conv} 12}  &  \multicolumn{2}{c}{\texttt{Conv 16}} \\
Homology & LOF & d & $\sigma$ & d & $\sigma$ & d & $\sigma$ & d & $\sigma$ \\
\midrule
\multicolumn{10}{c}{All distances} \\
\midrule
 $H_0$ & \ding{51} & 0.3   & 0.1   & 0.2   & 0.1   & 0.13  & 0.06  & 0.12  & 0.05  \\
 $H_0$ & \ding{55} & 0.3   & 0.1   & 0.2   & 0.1   & 0.13  & 0.06  & 0.12  & 0.05  \\
 $H_1$ & \ding{51} & 0.03  & 0.01  & 0.03  & 0.01  & 0.04  & 0.01  & 0.06  & 0.01  \\
 $H_1$ & \ding{55} & 0.03  & 0.01  & 0.03  & 0.01  & 0.04  & 0.01  & 0.05  & 0.01  \\
 $H_2$ & \ding{51} & 0.012 & 0.003 & 0.012 & 0.004 & 0.023 & 0.005 & 0.032 & 0.008 \\
 $H_2$ & \ding{55} & 0.012 & 0.003 & 0.012 & 0.003 & 0.023 & 0.005 & 0.032 & 0.007 \\
 \midrule
 \multicolumn{10}{c}{Class distances} \\
 \midrule
 $H_0$ & \ding{51} & 0.2   & 0.1   & 0.17  & 0.09  & 0.09  & 0.05  & 0.09  & 0.03  \\
 $H_0$ & \ding{55} & 0.2   & 0.1   & 0.19  & 0.09  & 0.09  & 0.05  & 0.09  & 0.04  \\
 $H_1$ & \ding{51} & 0.023 & 0.006 & 0.022 & 0.007 & 0.035 & 0.007 & 0.05  & 0.001 \\
 $H_1$ & \ding{55} & 0.023 & 0.008 & 0.023 & 0.008 & 0.035 & 0.008 & 0.048 & 0.001 \\
 $H_2$ & \ding{51} & 0.01  & 0.003 & 0.011 & 0.003 & 0.021 & 0.005 & 0.031 & 0.006 \\
 $H_2$ & \ding{55} & 0.010 & 0.003 & 0.011 & 0.003 & 0.021 & 0.005 & 0.03  & 0.007 \\
\bottomrule
\end{tabular}
\end{table}
The next part of the experiment compares the average bottleneck distance values with and without LOF. 
The distances, along with their standard deviation intervals, are reported in \Cref{tab:lof-vgg19} for all pairs of persistence diagrams (All distances) and for those belonging to the same class (Class distances). 
The table contains distances calculated with and without LOF using neural activations collected from different layers of the VGG19 network. 
We find observations that are both related and unrelated to LOF.
Starting with observations not related to LOF, we observe that for $H_0$ homology, the averaged bottleneck distance reaches lower values with activations coming from later layers. 
The same happens for the standard deviation of $H_0$ homology. 
The decrease is largest between \texttt{Conv 8} and \texttt{Conv 12}. 
That said, the opposite occurs for the two other homology groups, namely, $H_1$ and $H_2$.
We observe an increase in average bottleneck distance values and standard deviation intervals for those groups.
Again, the most substantial change occurs between \texttt{Conv 8} and \texttt{Conv 12} layers.
Another observation after analyzing this table is that the standard deviation is a notable factor compared to the average values, with values ranging from approximately twenty to over fifty percent.
The last observation unrelated to LOF is that the average bottleneck distances between persistence diagrams for Class-specific distances are lower than those for All distances.
The largest differences are visible for $H_0$ homology, with a value of 0.05 for \texttt{Conv 4} representing approximately one-quarter of the distance between classes. In contrast, the discrepancies in $H_2$ are around a few percent.

Regarding LOF, differences in the bottleneck distance with and without outlier removal are minor. 
For All distances, values with and without LOF are almost equal. 
However, we see a slightly smaller standard deviation for most of the layers and homologies. 
The real impact of using LOF is visible in $H_0$ homology for Class distances, where LOF reduces distance and standard deviation intervals by a few percent; for $H_1$, the decrease is negligible; and for $H_2$, there is a minimal increase, instead of a decline. 
The results indicate that the use of LOF is of minimal consequence, and the decision to use it has secondary importance. 
We also need to point out that we did not observe many outliers, so either they do not often occur in neural representations, or we did not find them.

\begin{table}[!ht]
\centering
\caption{ \begin{centering} Averaged bottleneck distances with and without LOF for various layers of ResNet18 network.\end{centering}}
\label{tab:lof-resnet18}
\begin{tabular}{cccccccccc}
\toprule   
\multicolumn{2}{c}{ResNet18} & \multicolumn{2}{c}{\texttt{Stage 1}} & \multicolumn{2}{c}{\texttt{Stage 2}} &  \multicolumn{2}{c}{\texttt{Stage 3}}  &  \multicolumn{2}{c}{\texttt{Stage 4}} \\
Homology & LOF & d & $\sigma$ & d & $\sigma$ & d & $\sigma$ & d & $\sigma$ \\
\midrule
\multicolumn{10}{c}{All distances} \\
\midrule
 $H_0$ & \ding{51} & 0.21  & 0.08  & 0.19  & 0.07  & 0.19  & 0.07  & 0.14 & 0.05  \\
 $H_0$ & \ding{55} & 0.21  & 0.08  & 0.2   & 0.08  & 0.19  & 0.07  & 0.14 & 0.05  \\
 $H_1$ & \ding{51} & 0.06  & 0.02  & 0.05  & 0.01  & 0.06  & 0.02  & 0.06 & 0.02  \\
 $H_1$ & \ding{55} & 0.06  & 0.02  & 0.06  & 0.017 & 0.06  & 0.018 & 0.06 & 0.016 \\
 $H_2$ & \ding{51} & 0.027 & 0.008 & 0.024 & 0.006 & 0.025 & 0.007 & 0.04 & 0.01  \\
 $H_2$ & \ding{55} & 0.028 & 0.008 & 0.025 & 0.007 & 0.025 & 0.006 & 0.04 & 0.01 \\
 \midrule
 \multicolumn{10}{c}{Class distances} \\
 \midrule
 $H_0$ & \ding{51} & 0.15  & 0.7   & 0.15  & 0.07  & 0.15  & 0.056 & 0.12  & 0.043 \\
 $H_0$ & \ding{55} & 0.15  & 0.08  & 0.15  & 0.07  & 0.16  & 0.07  & 0.12  & 0.04  \\
 $H_1$ & \ding{51} & 0.05  & 0.014 & 0.05  & 0.014 & 0.05  & 0.016 & 0.06  & 0.016 \\
 $H_1$ & \ding{55} & 0.05  & 0.01  & 0.05  & 0.01  & 0.05  & 0.014 & 0.06  & 0.014 \\
 $H_2$ & \ding{51} & 0.024 & 0.007 & 0.023 & 0.006 & 0.023 & 0.006 & 0.036 & 0.009 \\
 $H_2$ & \ding{55} & 0.025 & 0.007 & 0.023 & 0.006 & 0.023 & 0.005 & 0.037 & 0.009 \\
\bottomrule
\end{tabular}
\end{table}
The same results as in \Cref{tab:lof-vgg19}, but this time for ResNet18 architecture are reported in \Cref{tab:lof-resnet18}. 
There are some differences compared to VGG19.
First, values for all homology groups remain similar in all Stages. 
The only difference is in \texttt{Stage 4}, where for $H_0$ homology we observe lower values than in the other layers (for all distances and for distances between diagrams for the same class. 
We do not see any differences for $H_1$ homology, and for $H_2$ homology we notice a distance increase. The same trend that we observe for distance applies to the standard deviation intervals. 
Additionally, we observe that the standard deviation is high, at several tens of percent, similar to the VGG19 network. 
Also, the Class distances are lower than the average distance between all persistence diagrams, except for the homology $H_0$ in \texttt{Stage 4}. 
The impact of LOF is negligible, as in most cases we see the same or almost identical distance values.

In conclusion, when analyzing results for both architectures, some standard topological features have high variance. 
Class distances are lower than All distances.
LOF turns out to not have much role in analyzing neural representations using persistent homology, as results with LOF are almost identical to results without it.
This is visible in identical persistence diagrams where no outliers were detected, or in similar diagrams where some were detected. 

\begin{figure}[!ht]
    \centering
    \includegraphics[width=0.53\linewidth]{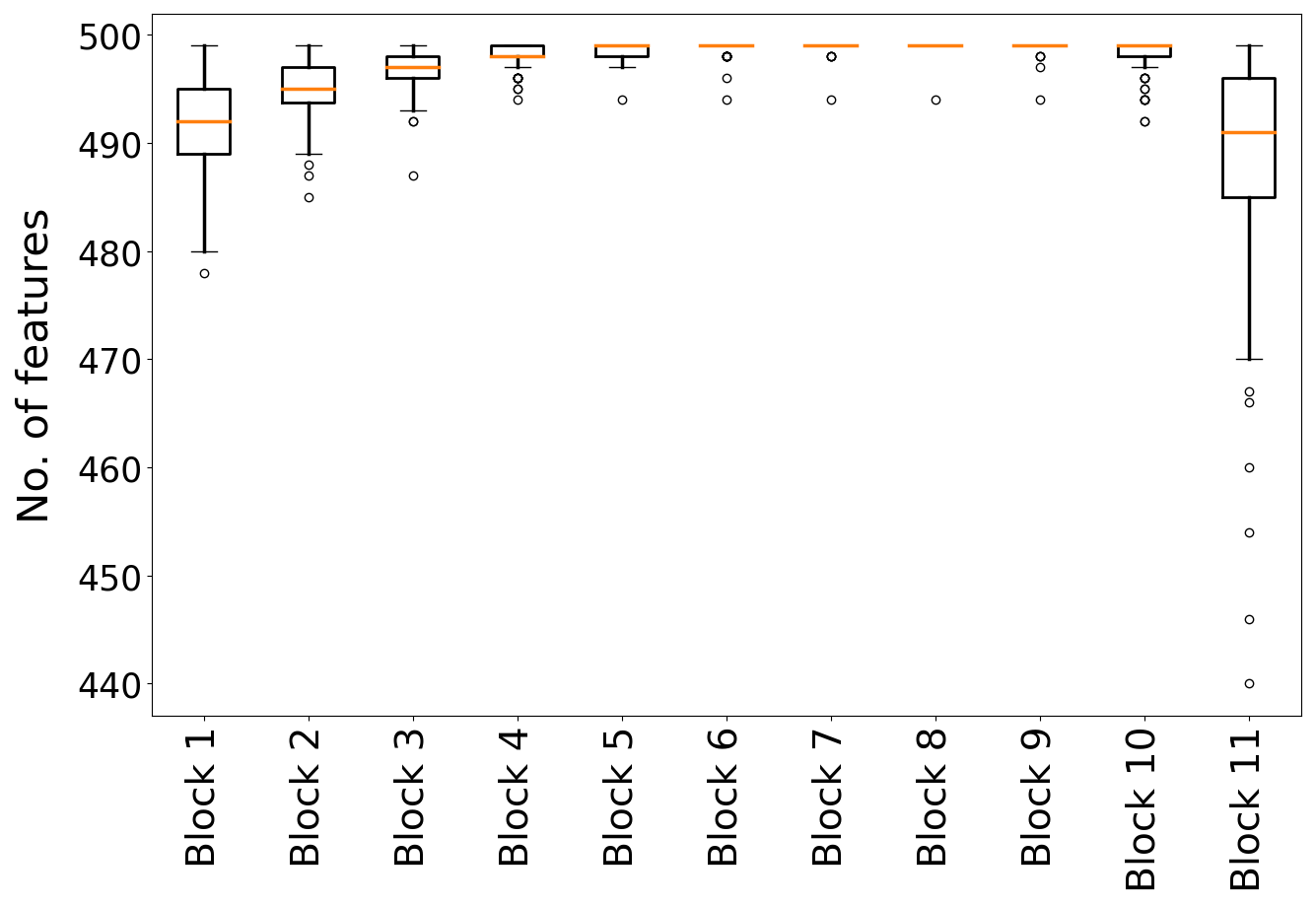}
  \caption{\centering Number of $H_0$ homological features in ViT model.}
  \label{fig:lof-points_number-vit}
\end{figure}
In addition to previous results, we also counted the average number of $H_0$ homological features per layer in all considered network architecture (VGG19, ResNets, and ViT).
This gives us information about the number of outliers detected, as for each removed outlier we observe one point less on the persistent diagram.
It turns out that for ResNets and VGG19 the number of $H_0$ features is almost unchanged in most cases, and is equal to the number of inputs used to collect neural representation.
That said, we observe a distinctive behavior for ViT.
\Cref{fig:lof-points_number-vit} presents the average number of per-class homological features in ViT representations.
The chart contains boxplots with the average number of features for consecutive Blocks. 
We observe that for \texttt{Block 1} and \texttt{Block 2}, the average number of homological features is between 500 and 490.
This number of features equals about ten outliers for each class in the dataset. 
We see that the number of outliers goes down in the successive layers, and for \texttt{Block 7} and \texttt{Block 8} there are almost no outliers. 
An interesting thing happens at \texttt{Block 11}, where we see many detected outliers. 
For some classes, the number of outliers is equal to more than 50 (that is above 10\% of data), but for most classes, it is around ten outliers.

This shows that the Transformer network processes data differently than convolutional models, with many more outliers at certain depths.
At the same time, different convolutional architectures appear to process data similarly.
For CNNs, there are no outliers in most classes.
The few exceptions typically have at most few of them. 
For CNNs, therefore, applying LOF before calculating persistent homology appears to be insignificant. 
Nevertheless, using it is not computationally expensive and does not distort the results, so its application is a matter of experimenter choice.
In ViT, LOF removes a much larger number of outliers.
It is, therefore, worth using, as outliers can perturb the results.
For this reason, we will be using LOF for all models considered in this thesis, especially since removing outliers isn't computationally expensive.

The final observation is that the change in average distances varies from model to model and from layer to layer. 
We will analyze this further in the following experiment.

\newpage
\section{Topological characterization of selected architectures}
\label{characterization}
In this experiment we use persistent homology to calculate various topological features for selected network architectures, namely, plain convolutional network (VGG19), residual networks (ResNet18 and ResNet50), and vision transformer. 
We will start analyzing our results by inspecting persistence diagrams for these architectures.

\begin{figure}[!ht]
    \centering
    \subfloat[VGG19, \texttt{Conv 12}]{
        \label{fig:char-diagram-vgg19}
        \includegraphics[width=0.40\linewidth]{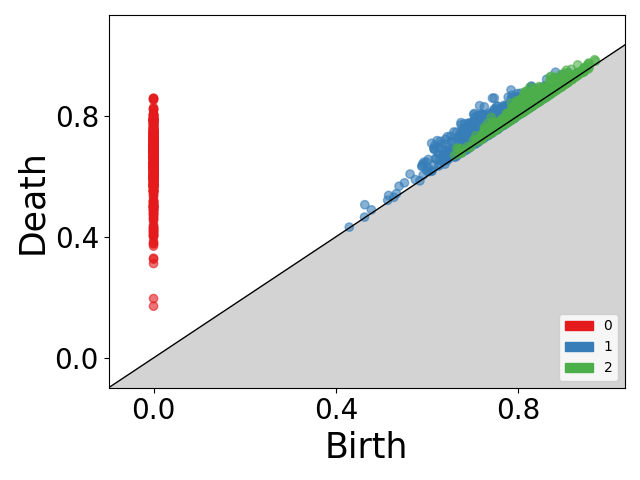}
    }
    \subfloat[ResNet18, \texttt{Stage 3}]{
        \label{fig:char-diagram-resnet18}
        \includegraphics[width=0.40\linewidth]{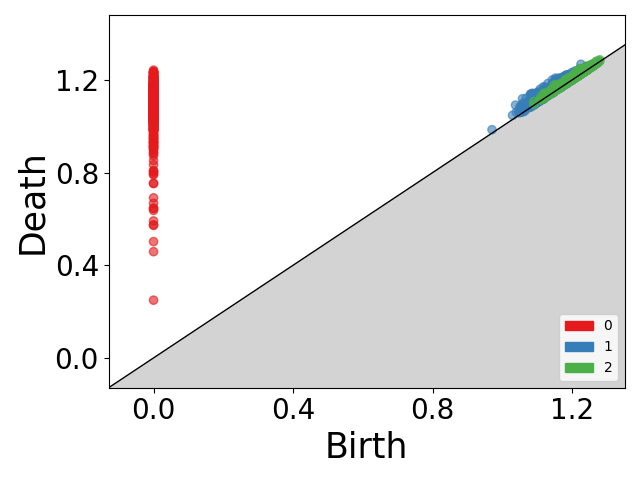}
    }
    \\
    \subfloat[ResNet50, \texttt{Stage 3 Block 3}]{
        \label{fig:char-diagram-resnet50}
        \includegraphics[width=0.40\linewidth]{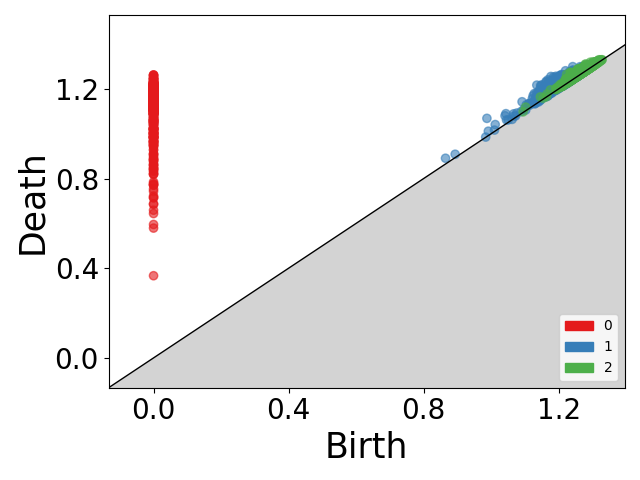}
    }
    \subfloat[ViT, \texttt{Block 7}]{
        \label{fig:char-diagram-vit}
        \includegraphics[width=0.40\linewidth]{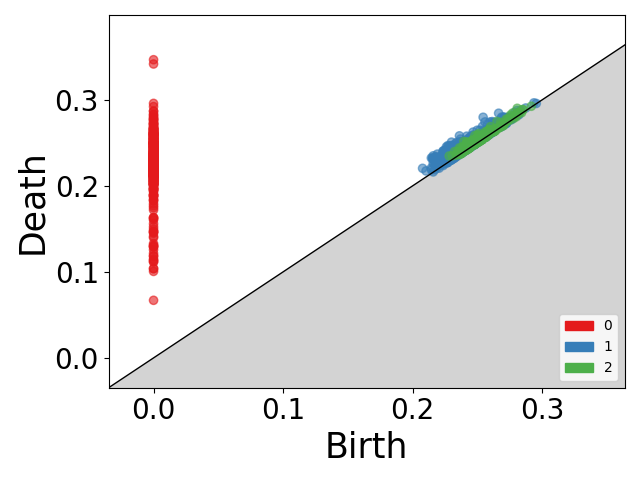}
    }
  \caption{\centering Persistence diagrams for the raccoon class in selected models.}
  \label{fig:char-diagram-raccon}
\end{figure}
\Cref{fig:char-diagram-raccon} shows four persistence diagrams for the raccoon class. They were calculated using representations in layers from the second half of the corresponding networks. 
Even though these diagrams look similar, there are some differences.
Firstly, except for ResNet models on \Cref{fig:char-diagram-resnet18,fig:char-diagram-resnet50}. different networks have different scales. 
This happens despite normalizing neural representations before calculating persistent homology.
Secondly, there is a difference in the number and position of $H_1$ and $H_2$ homological features. 
The largest number of features is in the diagram for the VGG19 network (\Cref{fig:char-diagram-vgg19}).
These homological features are also more relevant than in other models, because they are further away from the diagonal line.
For VGG19, we also observe that $H_2$ features live after the $H_0$ features die, which is much less common in other networks and does not appear in ViT. 
Lastly, only in ViT (\Cref{fig:char-diagram-vit}), there are $H_0$ features above the central cluster of higher homology groups (by cluster we mean a concentration of points for a specific homology group).
Regarding common features for all diagrams, we see one or two $H_0$ features that are present below the cluster. 
Furthermore, the features from $H_1$ and $H_2$ generally live for a short period, especially compared to $H_0$ homology group. 
To summarize this analysis, certain differences between representations make it possible to determine the model family by looking at the respective persistence diagrams.
While not reported here, persistence diagrams for other classes look similar to those presented in this work.

\subsection{Detailed analysis of plain convolutional networks}

\begin{figure}[!ht]
    \subfloat[Train 500]{
        \label{fig:char-vgg19-train500}
        \includegraphics[width=0.306\linewidth]{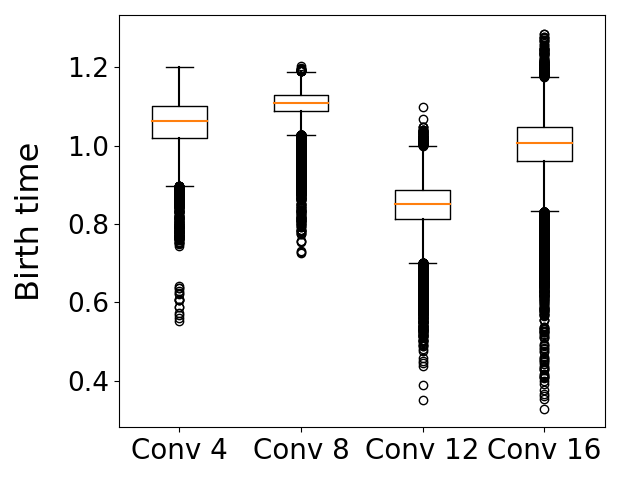}
    }
    \subfloat[Train 100]{
        \label{fig:char-vgg19-train100}
        \includegraphics[width=0.306\linewidth]{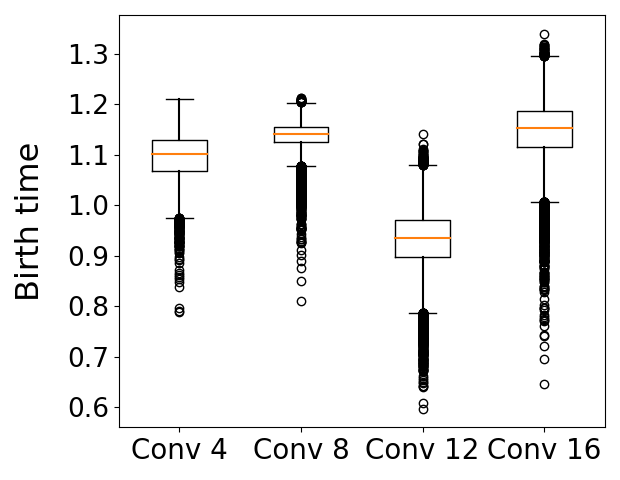}
    }
    \subfloat[Test 100]{
        \label{fig:char-vgg19-test100}
        \includegraphics[width=0.306\linewidth]{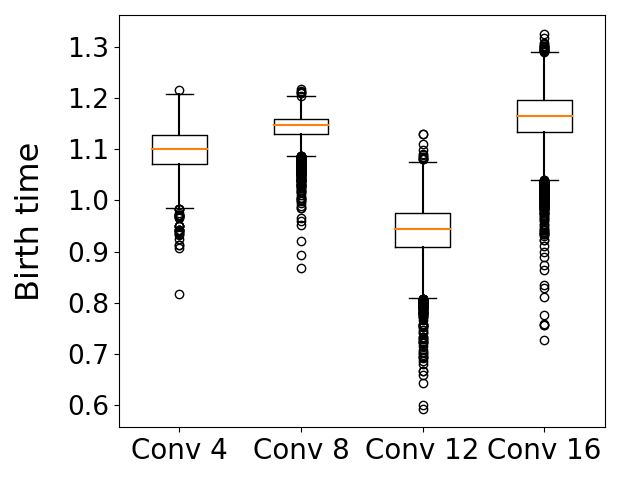}
    }
  \caption{\centering Boxplots of $H_2$ birth times for VGG19 network. Results are reported for the test and training set, using 100 and 500 input elements.}
  \label{fig:char-vgg19-test_train}
\end{figure}

We will now analyze the basic characteristics of persistence diagrams.
We will start with a plain convolutional network, i.e., VGG19. 
First, we compute boxplots of the birth times for the $H_2$ homological features in selected layers and input datasets. 
The results are displayed on \Cref{fig:char-vgg19-test_train}.

We compute the exact statistics for three different neural representations. 
The first one, reported in \Cref{fig:char-vgg19-train500}, was calculated using all 500 elements per class from the training set.
The second chart (\Cref{fig:char-vgg19-train100}) contains the same data, but with a random subset of 100 elements per class.
The last chart (\Cref{fig:char-vgg19-test100}) differs from the other two by using the test set.
All these charts are similar, but with minor differences.
In all charts, the median value drops in the \texttt{Conv 12} layer and then returns to typical levels in \texttt{Conv 16}.
This drop confirms the similarity of all charts and, therefore, topological similarity between different neural representations.
Also, with representations from the \texttt{Conv 4} layer, we didn't observe any outliers above the median, apart from \Cref{fig:char-vgg19-test100} which shows one outlier.
We see many outliers below the median for this layer.
These outliers give us a long-tail distribution.
A similar phenomenon is also present in the latter layers, but is less intensive.

Additionally, we notice a difference in birth times when comparing 100 elements charts with the 500 elements chart; the median birth time is less by about 0.1 (\Cref{fig:char-vgg19-train500}) compared to the two other charts. 
Also, charts for test and train data (with the same number of input examples) are almost identical. 
In particular, there is less difference between persistent diagrams computed from train and test data (with the same number of points) than between diagrams calculated with a different number of train inputs.

\begin{figure}[!ht]
    \centering
    \subfloat[$H_1$]{
        \label{fig:char-vgg19-points_number_1}
        \includegraphics[width=0.40\linewidth]{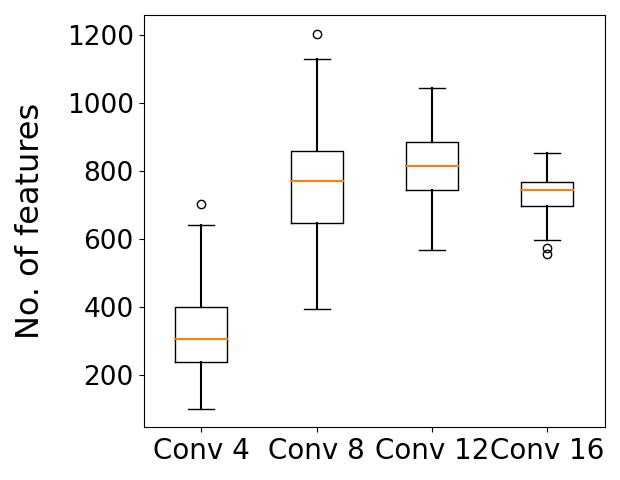}
    }
    \subfloat[$H_2$]{
        \label{fig:char-vgg19-points_number_2}
        \includegraphics[width=0.40\linewidth]{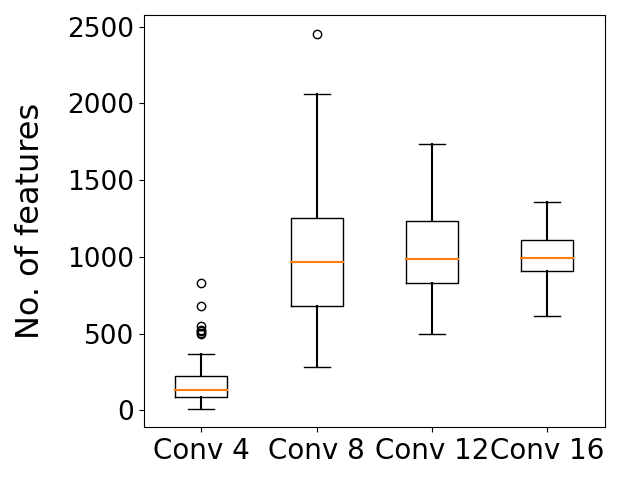}
    }
  \caption{\centering Boxplots of the number of homological features in VGG19 network representations.}
  \label{fig:char-vgg19-points}
\end{figure}
Next, we analyze other persistence homology characteristics.
Because we do not observe differences between the train and test datasets, we only show the train dataset with the maximum available elements per class (500). 
The boxplots with the number of features are reported on \Cref{fig:char-vgg19-points}. 
We observed that for the $H_0$ homology group for all layers, there were 499 homological features in almost every layer (due to a lack of outliers).
We therefore omitted this chart. 
For the $H_1$ homology group (\Cref{fig:char-vgg19-points_number_1}), we observe an increase in the number of homological features in \texttt{Conv 8} layer.
This statistic then oscillates around similar levels in subsequent layers. 
Also, in the deeper layers spread of the homological features is reduced.
A larger number of homological features in \texttt{Conv 8} (compared with \texttt{Conv 4}) is also the case for the $H_2$ homology group (\Cref{fig:char-vgg19-points_number_2}). 
Here, like in $H_1$ homology, the spread of the number of features is also reduced in deeper layers of the network. 
Moreover, we do not see many outliers here, as most of them occur in the \texttt{Conv 4} layer.

\begin{figure}[!ht]
    \centering
    \subfloat[Alive]{
        \label{fig:char-vgg19-alive}
        \includegraphics[width=0.306\linewidth]{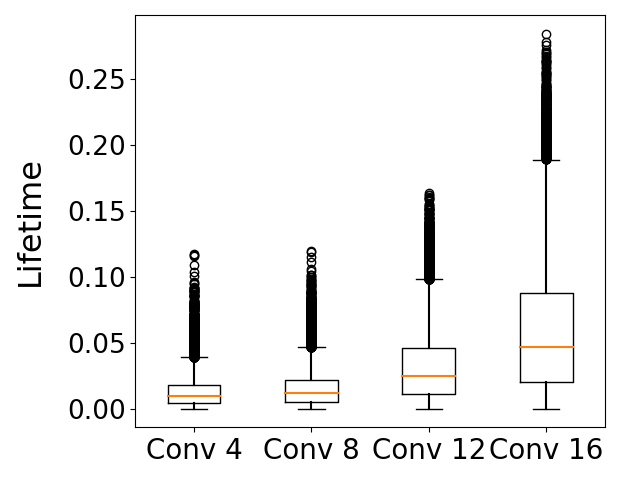}
    }
    \subfloat[Birth]{
        \label{fig:char-vgg19-born}
        \includegraphics[width=0.306\linewidth]{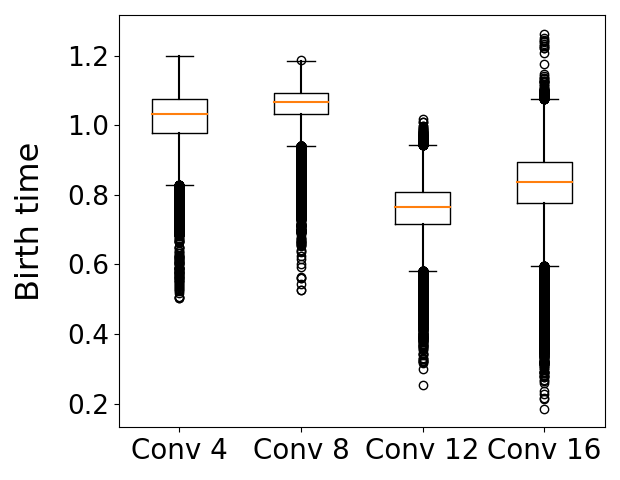}
    }
    \subfloat[Death]{
        \label{fig:char-vgg19-death}
        \includegraphics[width=0.306\linewidth]{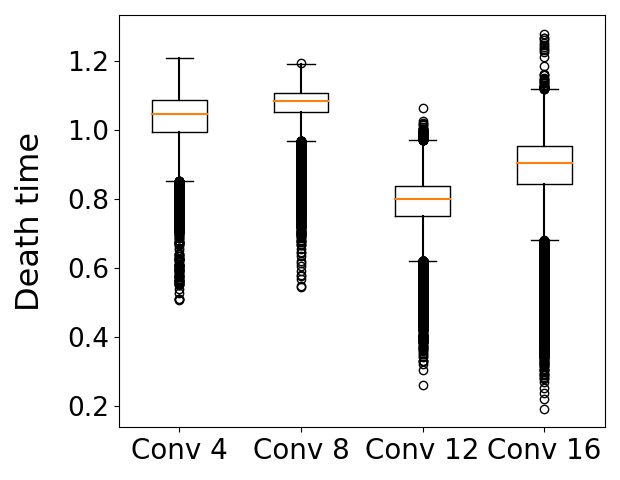}
    }
  \caption{\centering Boxplots for VGG19 network with statistics at homological features for $H_1$ homology group.}
  \label{fig:char-vgg19-alive_birth_death}
\end{figure}
Apart from looking at the number of homological features, we also checked different statistics, such as lifetime, birth, and death times. 
The comparison of these statistics for VGG19 is reported on \Cref{fig:char-vgg19-alive_birth_death}. 
All results are for the $H_1$ homology.
The boxplots for the average lifetime of features (\Cref{fig:char-vgg19-alive}) display a long-tail distribution, which is expected when most features live for a short period, and only some live longer. 
Homological features tend to live longer in further layers, and their lifetime has more variance.
Times of birth are shown on \Cref{fig:char-vgg19-born}.
The median time of birth drops by around 20\% between \texttt{Conv 8} and \texttt{Conv 12}. 
Another observation is that most of the outliers are features that started to live before average birthtime.
In particular, for \texttt{Conv 4} no outliers are above the median.
Boxplots showing death time are displayed on \Cref{fig:char-vgg19-death}.
The distributions are similar to birthtime boxplots. 
We observed similar results with the $H_2$ homology.

To summarize the analysis of VGG19, we observed that the number of inputs used to calculate representations affects the results. 
That said, the differences are typically small, and manifest by slightly different positions with an overall similar variance.
Additionally, we found that the layers starting from the middle of the network's depth exhibit a larger number of $H_1$ and $H_2$ homological features. 
These homologies persist longer and are present earlier in the network.

\subsection{Detailed analysis of ResNet models}

\begin{figure}[!ht]
    \centering
    \subfloat[Features number]{
        \label{fig:char-resnet18-points}
        \includegraphics[width=0.4\linewidth]{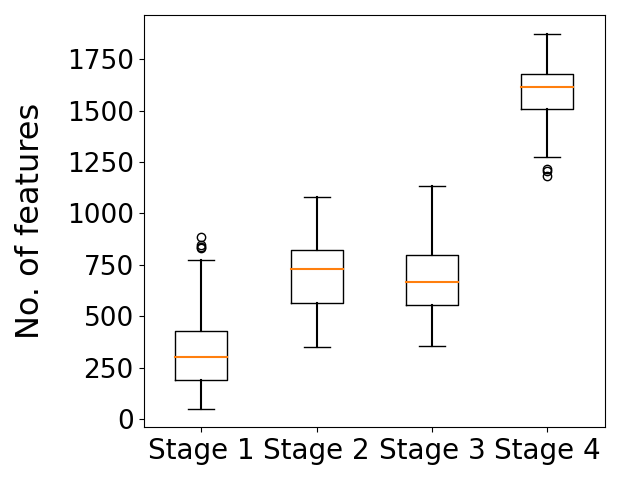}
    }
    \subfloat[Alive time]{
        \label{fig:char-resnet18-alive}
        \includegraphics[width=0.4\linewidth]{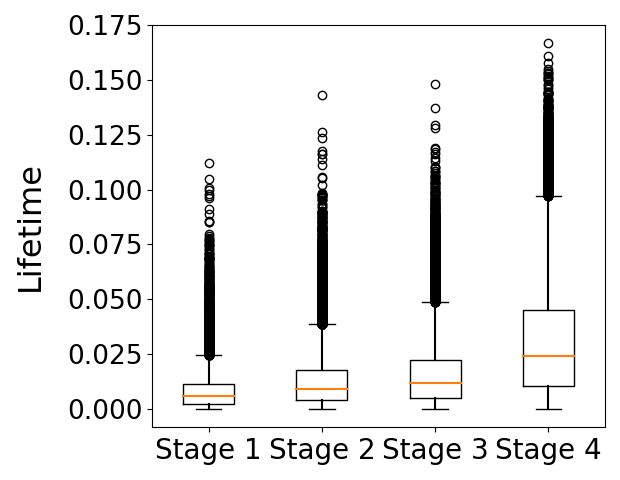}
    }
    \\
    \subfloat[Birth]{
        \label{fig:char-resnet18-born}
        \includegraphics[width=0.4\linewidth]{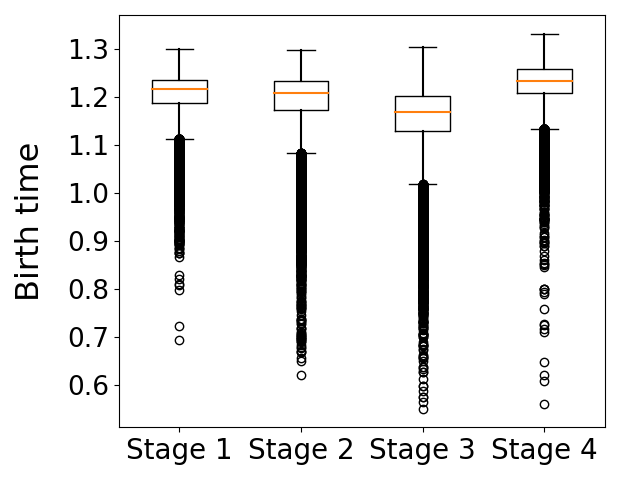}
    }
    \subfloat[Death]{
        \label{fig:char-resnet18-death}
        \includegraphics[width=0.4\linewidth]{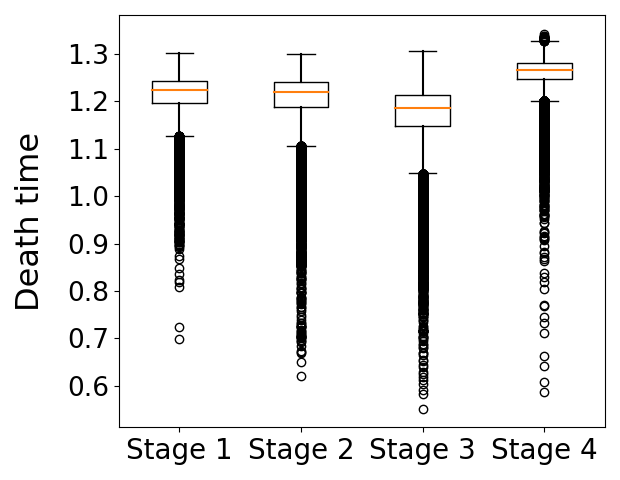}
    }
  \caption{\centering Boxplots for different statistics at $H_1$ homology group in ResNet18 network.}
  \label{fig:char-resnet18}
\end{figure}
Having explored VGG architecture, we will go to the next network. 
The architectures examined in this section are residual networks, represented firstly by ResNet18. 
In our experiments, many results for ResNet18 were similar to that for VGG19.
We will therefore show only some representative results (\Cref{fig:char-resnet18}).
We report the boxplot with the number per stage of $H_1$ homological features (\Cref{fig:char-resnet18-points}). 
Note an increase in feature number after \texttt{Stage 1}, which is analogous to the increase in feature number after \texttt{Conv 4} in VGG.
That said, we also see an even more significant increase in feature count at \texttt{Stage 4}.
When it comes to feature lifetimes (\Cref{fig:char-resnet18-alive}), we observe that the median values become increasingly larger for consecutive stages.
This observation corresponds to a similar phenomenon in VGG19.
The only difference is the scale. 
Also, the average lifetime for this network has a long-tail distribution. 
Boxplots showing birth and death time are displayed on \Cref{fig:char-resnet18-born,fig:char-resnet18-death}, respectively. 
Here, as in VGG19, these two charts are nearly identical.
This is because the time of birth or death is much longer than the lifetime.
The difference with the plain convolutional network is that the median birthtime and deathtime remain at the same level for all stages and lack outliers above the median. 
To summarize, ResNet18 performs similarly to VGG19, but with some minor differences.
These two models, however, have more common topological characteristics than distinct ones.

\begin{figure}[!ht]
    \centering
    \subfloat[$H_1$]{
        \label{fig:char-resnet50-points_number_1}
        \includegraphics[width=0.475\linewidth]{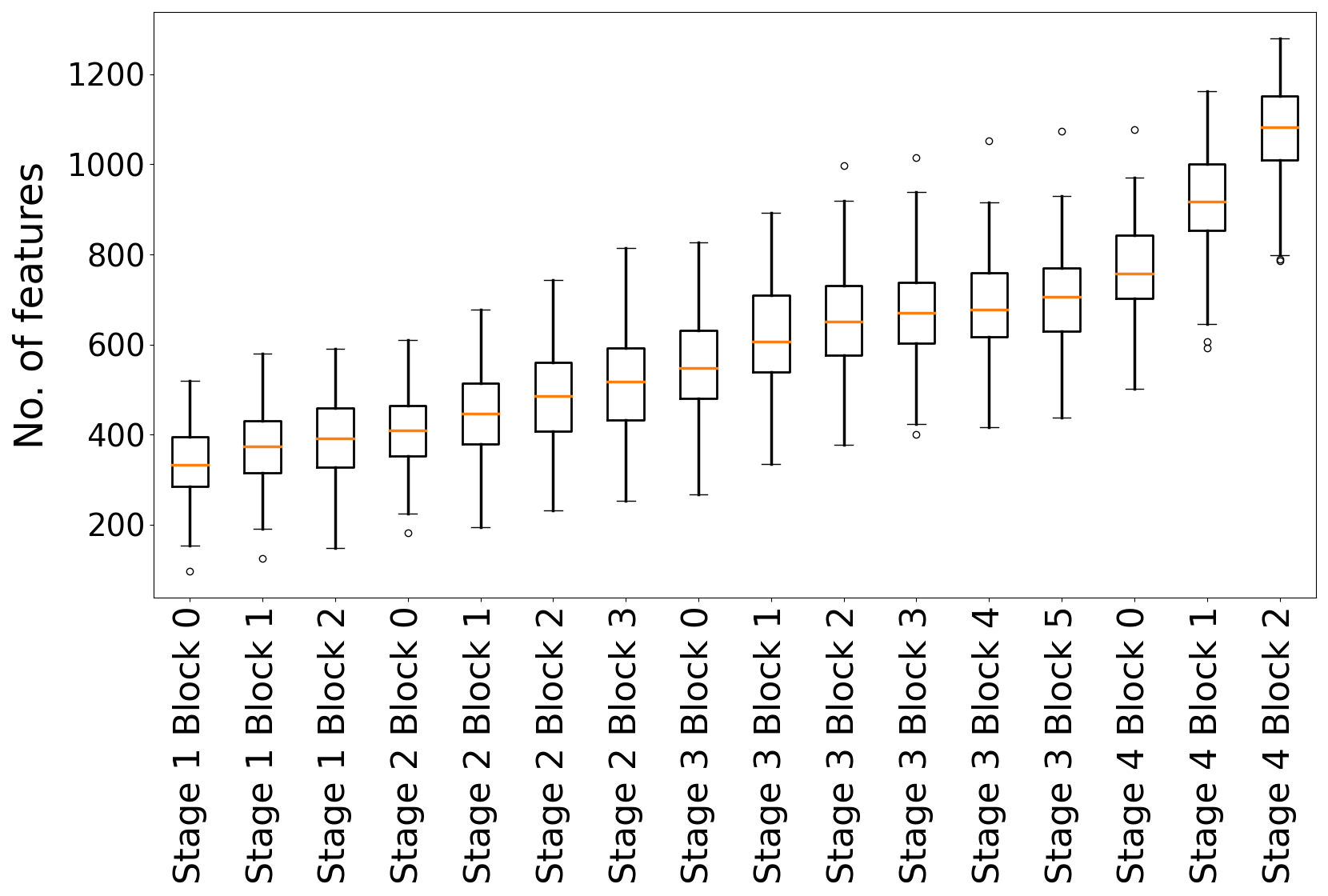}
    }
    \subfloat[$H_2$]{
        \label{fig:char-resnet50-points_number_2}
        \includegraphics[width=0.475\linewidth]{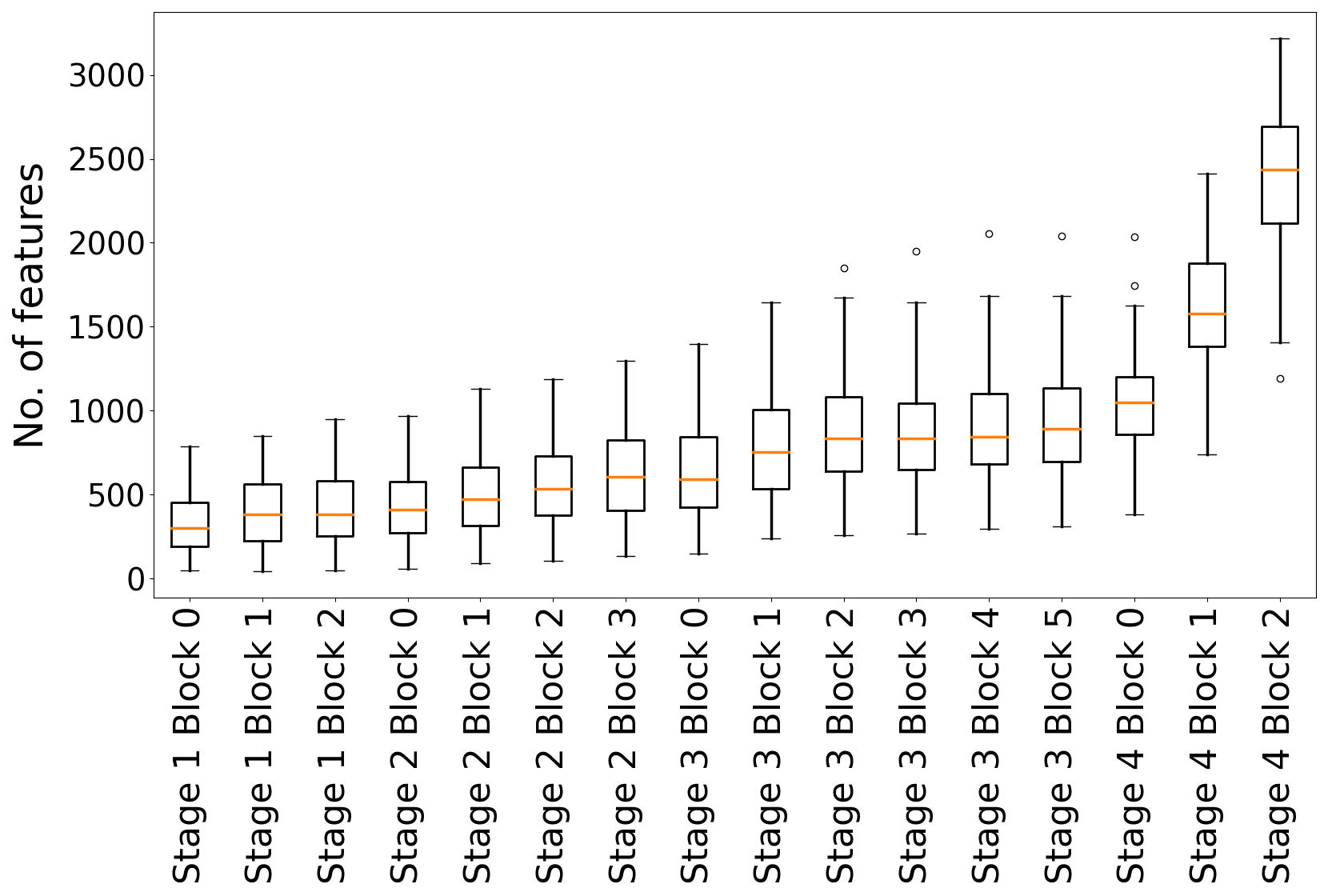}
    }
  \caption{\centering Boxplots of the number of homologies in ResNet50 network.}
  \label{fig:char-resnet50-points}
\end{figure}

Next, we analyze another residual network, namely, ResNet50. 
We will begin with boxplots showing the number of $H_1$ and $H_2$ homological features (\Cref{fig:char-resnet50-points}). For $H_0$ homology, we almost always have the same number of elements.

In the case of for ResNet50 network, we collected data for a larger number of layers, it is a bigger model than ResNet18.
We have already explained the reasons for this in \Cref{sec:lof}.
First, we observe a slow, steady growth in the number of features for $H_1$ homology (\Cref{fig:char-resnet50-points_number_1}). 
We also see that growth rate increases in \texttt{Stage 4}. Similar results are visible for $H_2$ (\Cref{fig:char-resnet50-points_number_2}). 
Here, the number of homological features in \texttt{Stage 4} increases much faster than in previous Stages. The large increase in the number of homological features is visible also in ResNet18 architecture, and this appears to be a common trait for residual convolutional networks. 
We do not observe a similar increase in VGG19 or ViT models.
Additionally, for ResNet50 we did not see many outliers in either homology group.

\begin{figure}[!ht]
    \centering
    \subfloat[$H_0$]{
        \label{fig:char-resnet50-death-0}
        \includegraphics[width=0.475\linewidth]{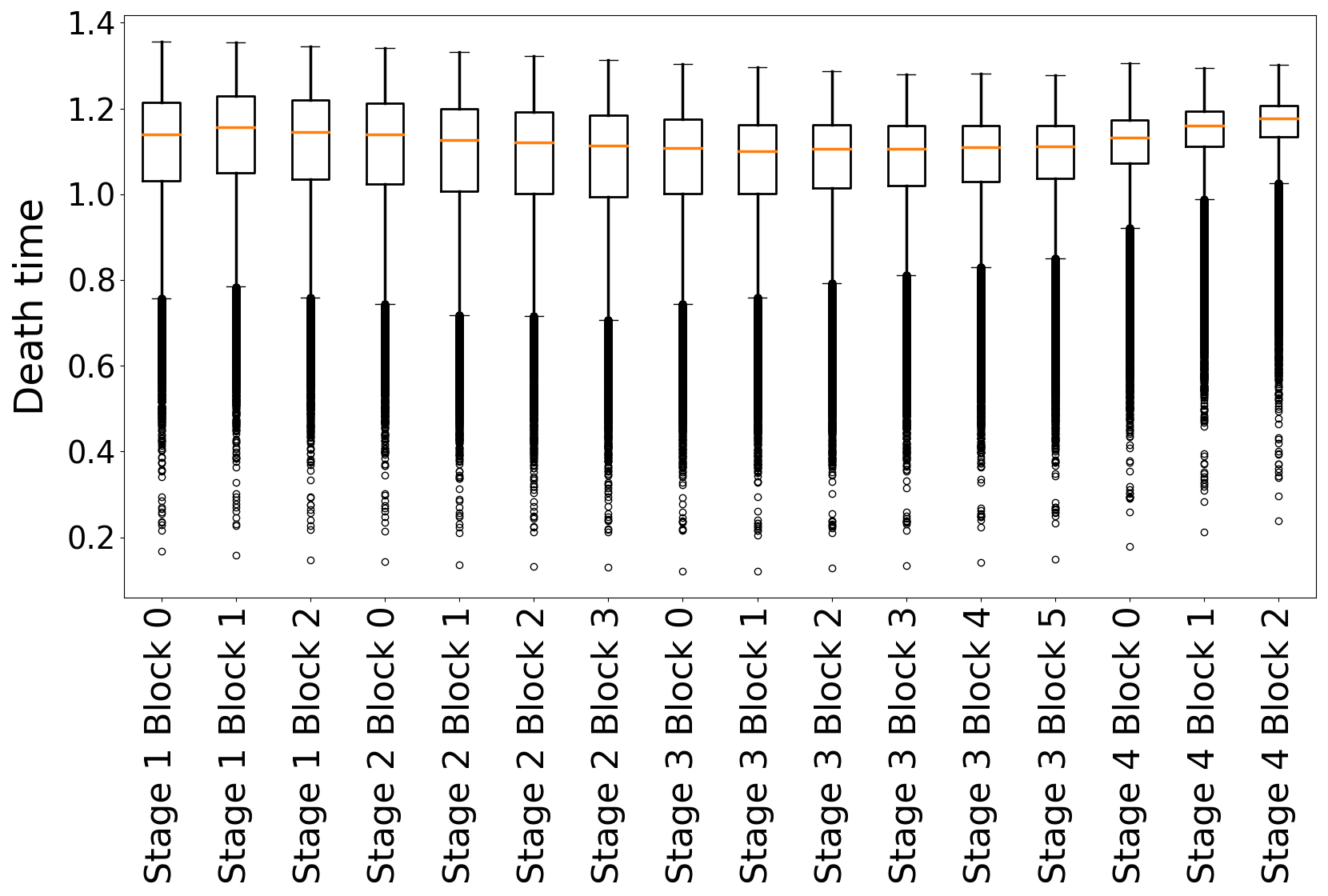}
    }
    \subfloat[$H_2$]{
        \label{fig:char-resnet50-death-2}
        \includegraphics[width=0.475\linewidth]{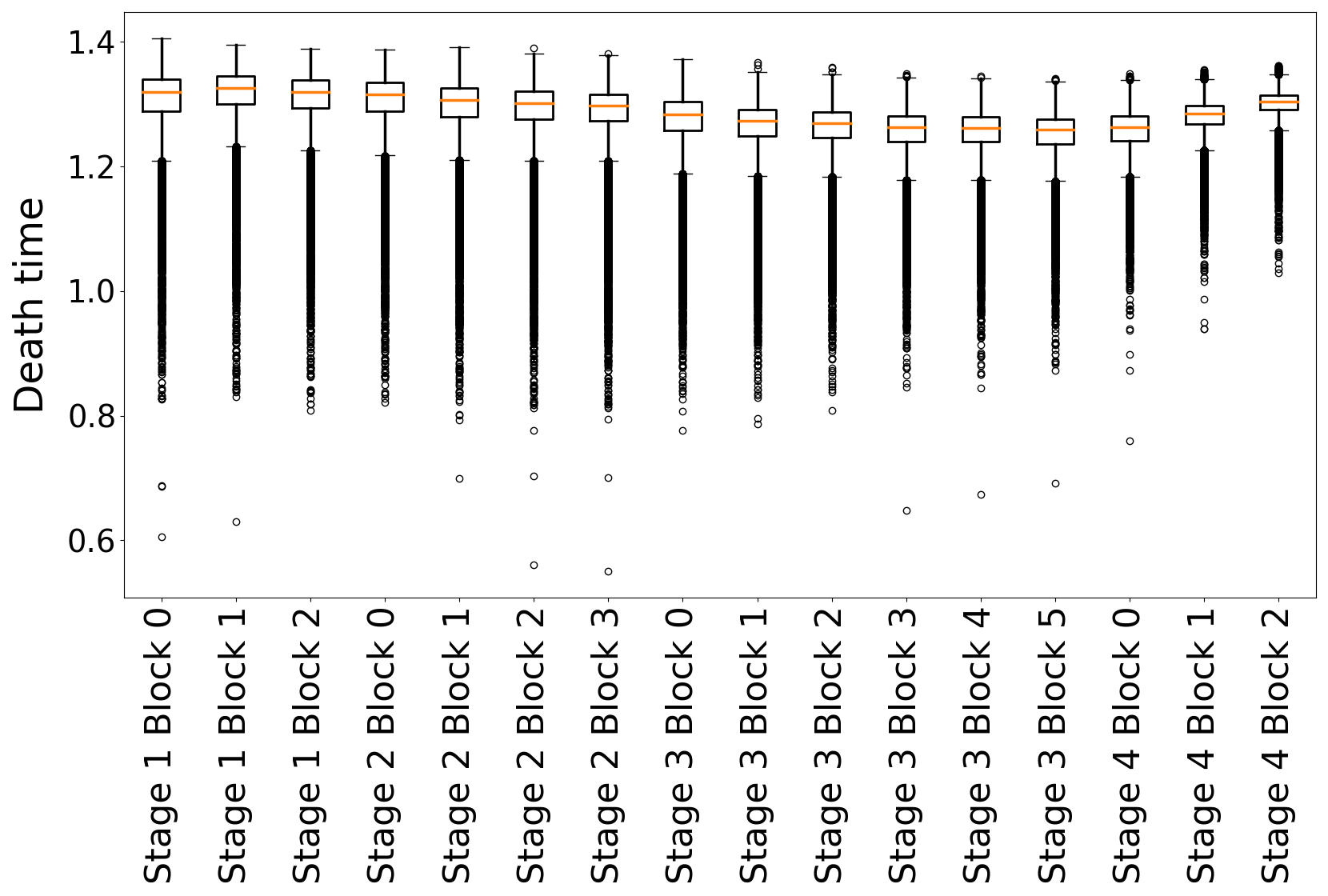}
    }
  \caption{\centering Boxplots of the death times for $H_0$ and $H_2$ homologies in ResNet50 network.}
  \label{fig:char-resnet50-death}
\end{figure}
Next, on \Cref{fig:char-resnet50-death} we report feature death times for $H_0$ and $H_2$ homology. 

We observe that the mean time of death for connected components (\Cref{fig:char-resnet50-death-0}) remains roughly constant in all layers.
Still, we see a smaller interquartile range for later stages, especially the last.
Additionally, the outliers for this homology group are all located below the median values.
For the results for $H_2$ homology (\Cref{fig:char-resnet50-death-2}), we observe that boxplots are located at roughly the same level, as for $H_0$ homology.
The interquartile range is comparable for all stages, except \texttt{Stage 4 Block 2}, for which the interquartile range is smaller.
Also, most of the outliers for this homology are below the median values, except for \texttt{Stage 3} and \texttt{Stage 4}, where a small number of outliers appear above the median.
All these results, e.g., an increase in the number of homological features in \texttt{Stage 4}, are similar to results for ResNet18. 
The results for $H_1$ homology group are not reported here, but are similar to the results for $H_2$ homology.

\begin{figure}[!ht]
    \centering
    \subfloat[$H_0$]{
        \label{fig:char-resnet50-alive-0}
        \includegraphics[width=0.475\linewidth]{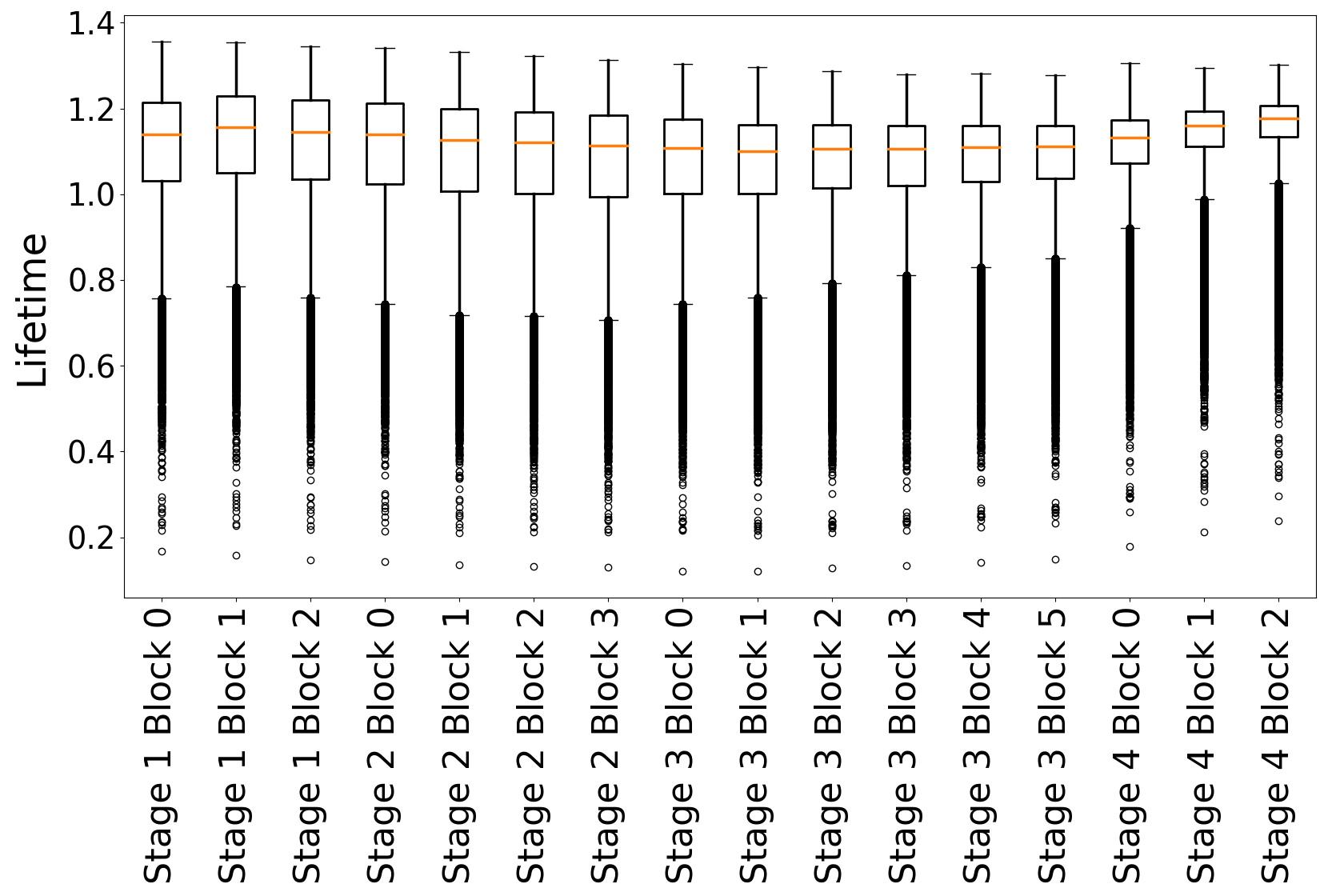}
    }
    \subfloat[$H_2$]{
        \label{fig:char-resnet50-alive-2}
        \includegraphics[width=0.475\linewidth]{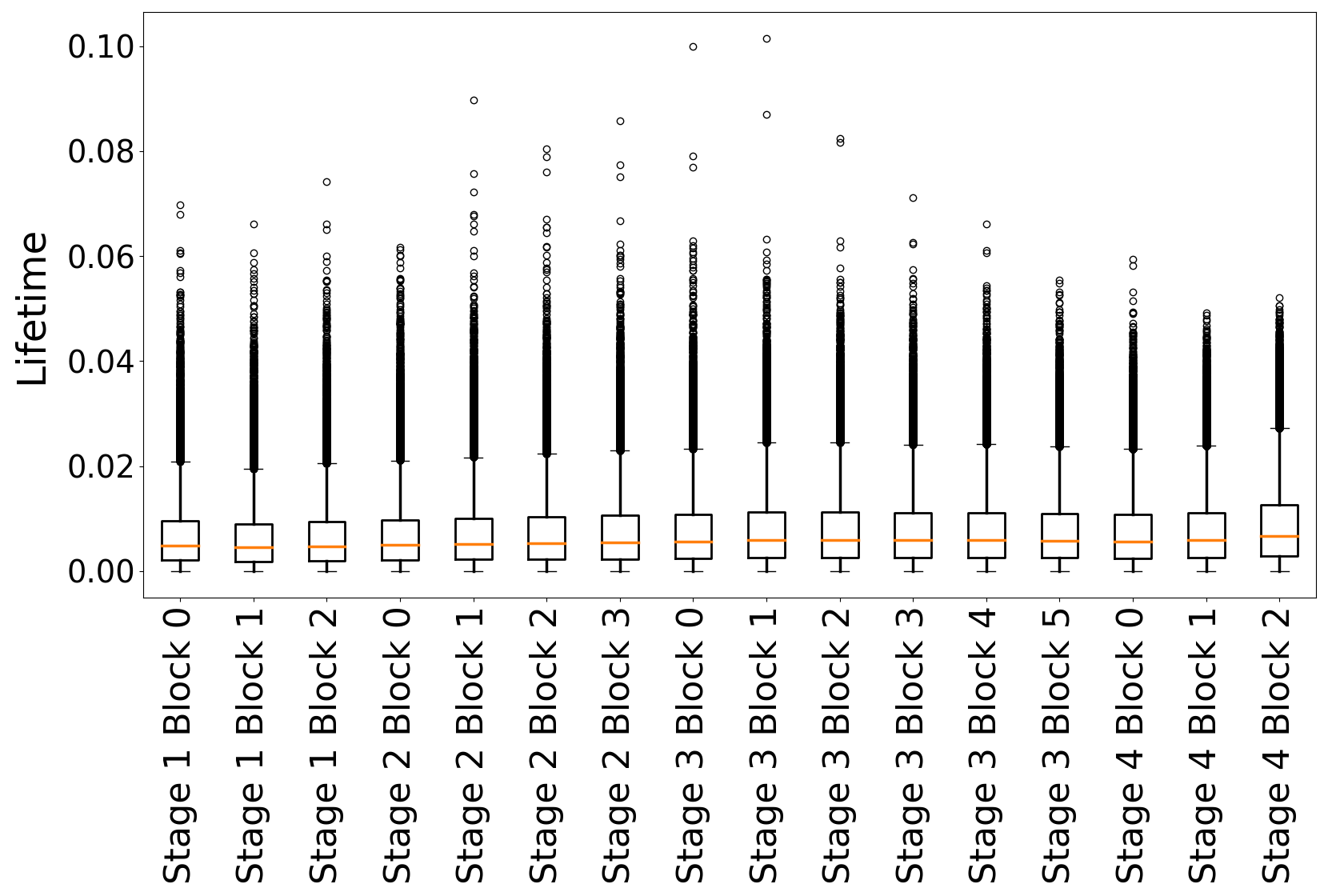}
    }
  \caption{\centering Boxplots of the time of life for $H_0$ and $H_2$ homologies in ResNet50 network.}
  \label{fig:char-resnet50-alive}
\end{figure}
For ResNet50, we also prepared lifetime boxplots reported on \Cref{fig:char-resnet50-alive}.
The results for $H_0$ homology (\Cref{fig:char-resnet50-alive-0}) are similar to those on \Cref{fig:char-resnet18-alive}.
In particular, the interquartile range is smaller for later stages, medians stay at a similar level across network depth, and outliers occur only below the median.
When analyzing the lifetime for $H_2$ homological features (\Cref{fig:char-resnet50-alive-2}), we see that the time of life for all layers remains steady (while not reported here, the same pattern appears for $H_1$ homology).
The variance also remains roughly unchanged across network depth.
This is in contrast to $H_0$ homology, where we observe a smaller interquartile range in deeper layers.
It also differs from the corresponding result for the ResNet18 model, where we observed an increase in average lifetime in later stages.

To conclude the ResNet50 analysis, most of the results for this network do not change much across the whole network depth.
The differences, when they appear, happen mostly in \texttt{Stage 4}, e.g., a rapid increase in the number of homological features.
Also model gives results similar to those of ResNet18.
Most similar are the birthtime and deathtime distribution in both residual networks.
Additionally, on many boxplots we see a long-tail distribution.
Features lifetimes (\Cref{fig:char-resnet50-alive}) are a prominent example.
The results for ResNet50 also demonstrate the outcomes of measuring a deeper network.
In this particular network we did not see significant changes between the successive layers, as we saw for VGG19, e.g., between \texttt{Conv 4} and \texttt{Conv 8} (\Cref{fig:char-vgg19-points}).

\subsection{Detailed analysis of the ViT model}

We will now examine our last model, namely, ViT.
In this model we observe behavior that is quite different from that of convolutional networks.
For example, boxplots with the number of homological features in ViT display distribution markedly different than those in convolutional models.
We have already shown the number of $H_0$ homology features per diagram on \Cref{fig:lof-points_number-vit}.
This chart shows that the number of homological features increases until \texttt{Block 6} or \texttt{Block 8}, and then starts decreasing, especially for in last block.
However, the number of features for $H_0$ homology is highly correlated with the number of outliers in neural representations.
That said, it turns out that the results look similar for other homologies.
\begin{figure}[!ht]
    \centering
    \subfloat[$H_1$]{
        \label{fig:char-vit-points-1}
        \includegraphics[width=0.475\linewidth]{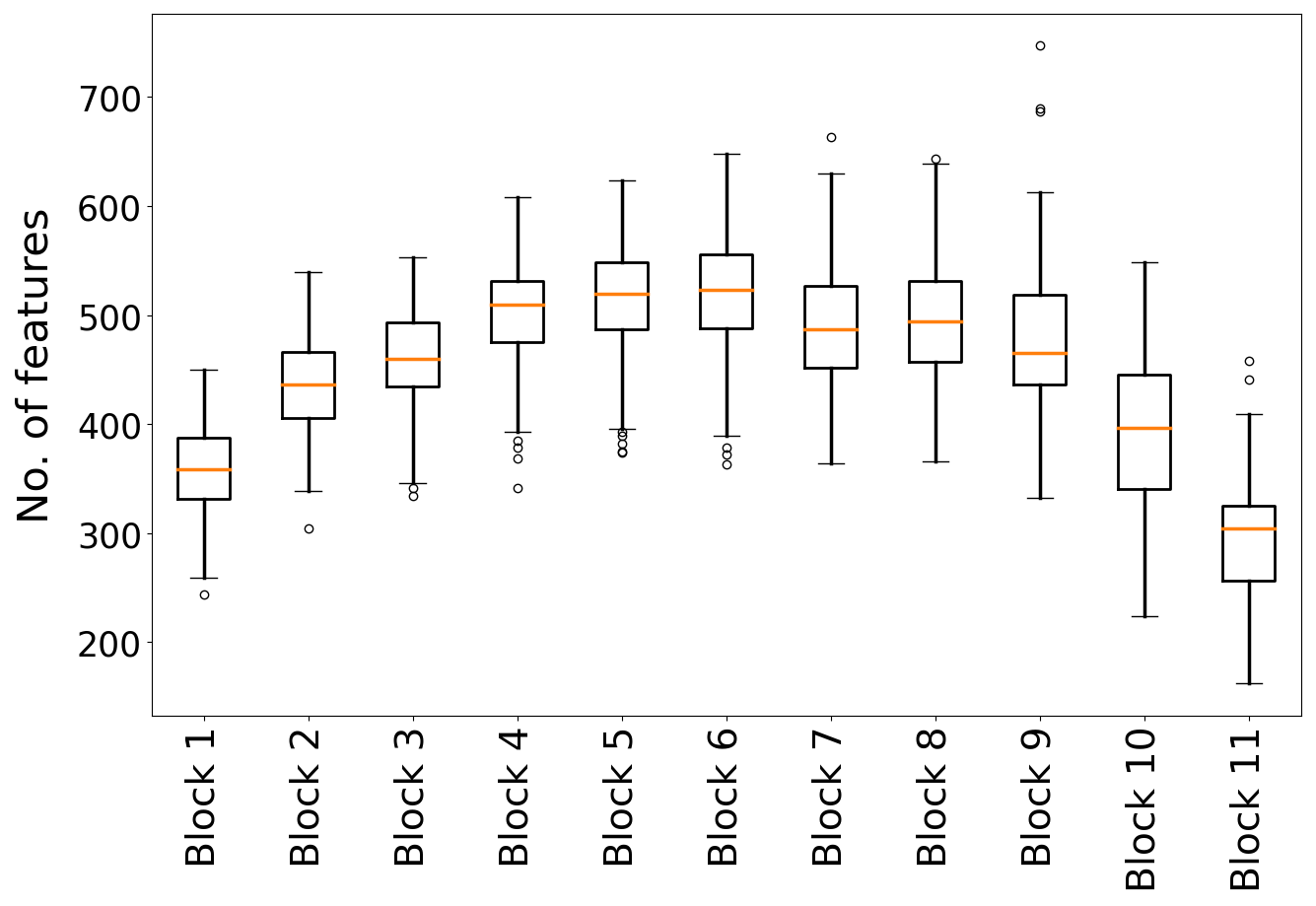}
    }
    \subfloat[$H_2$]{
        \label{fig:char-vit-points-2}
        \includegraphics[width=0.475\linewidth]{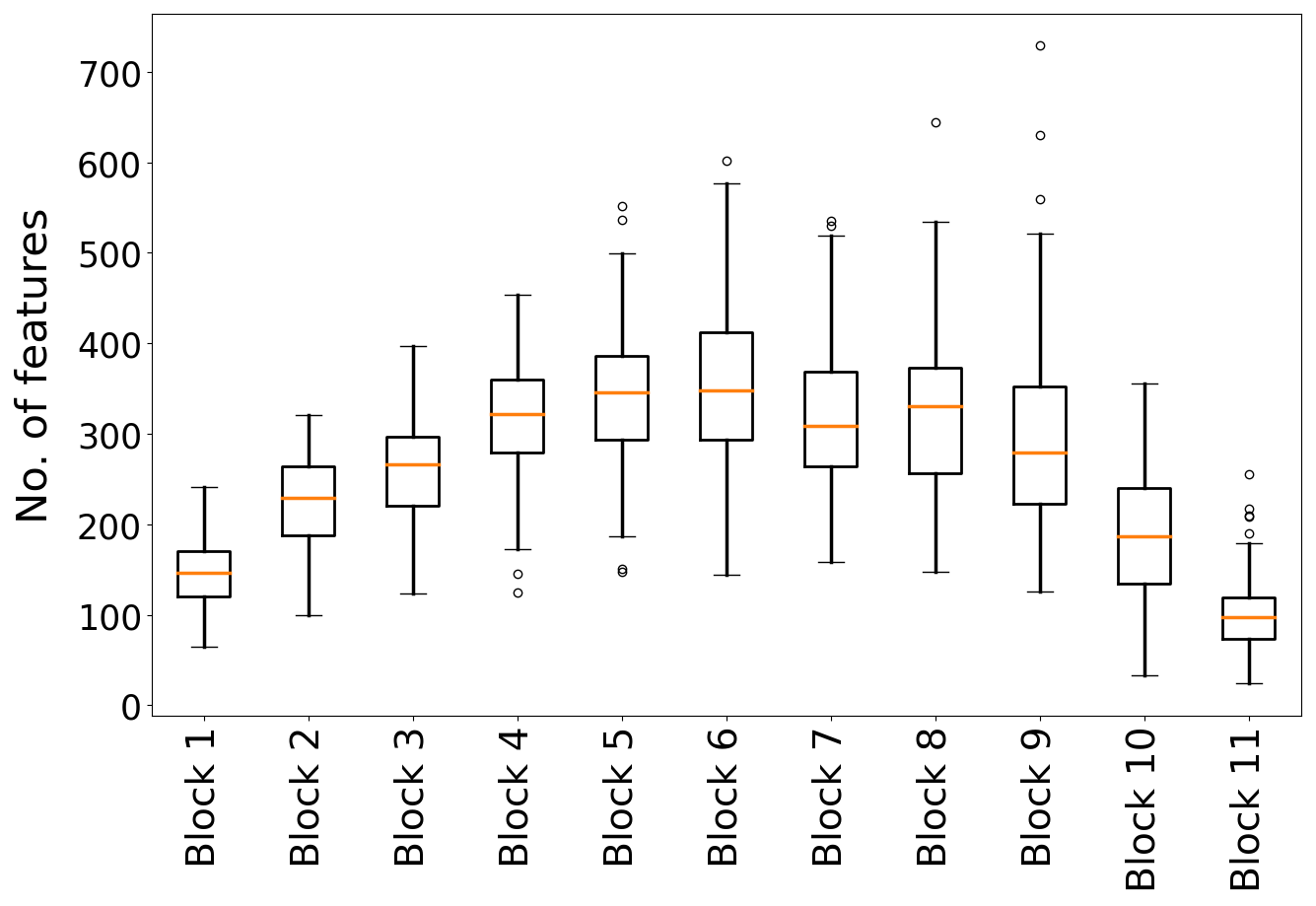}
    }
  \caption{\centering Boxplots of the number of $H_0$ and $H_2$ homological features homologies for the ViT network.}
  \label{fig:char-vit-points}
\end{figure}
We see this on \Cref{fig:char-vit-points}.
Specifically, for the $H_1$ homology (\Cref{fig:char-vit-points-1}), we observe an increase in the number of homological features from \texttt{Block 1} to \texttt{Block 4}.
Later, from \texttt{Block 4} to \texttt{Block 8}, we observe that the median number remains fixed at around 500 points.
We then see a significant drop in the last three blocks up to around 300 features in \texttt{Block 11}.
The interquartile range remains similar for all blocks.
This is in contrast to $H_2$ homology (\Cref{fig:char-vit-points-2}) where the interquartile ranges are bigger in blocks at the center of the network than in blocks at the beginning and end of the network.
Nevertheless, for $H_2$ homology, we also observe an increase in the number of homological features in the initial blocks and a decline in the final blocks.

\begin{figure}[!ht]
    \centering
    \subfloat[Death]{
        \label{fig:char-vit-death-2}
        \includegraphics[width=0.475\linewidth]{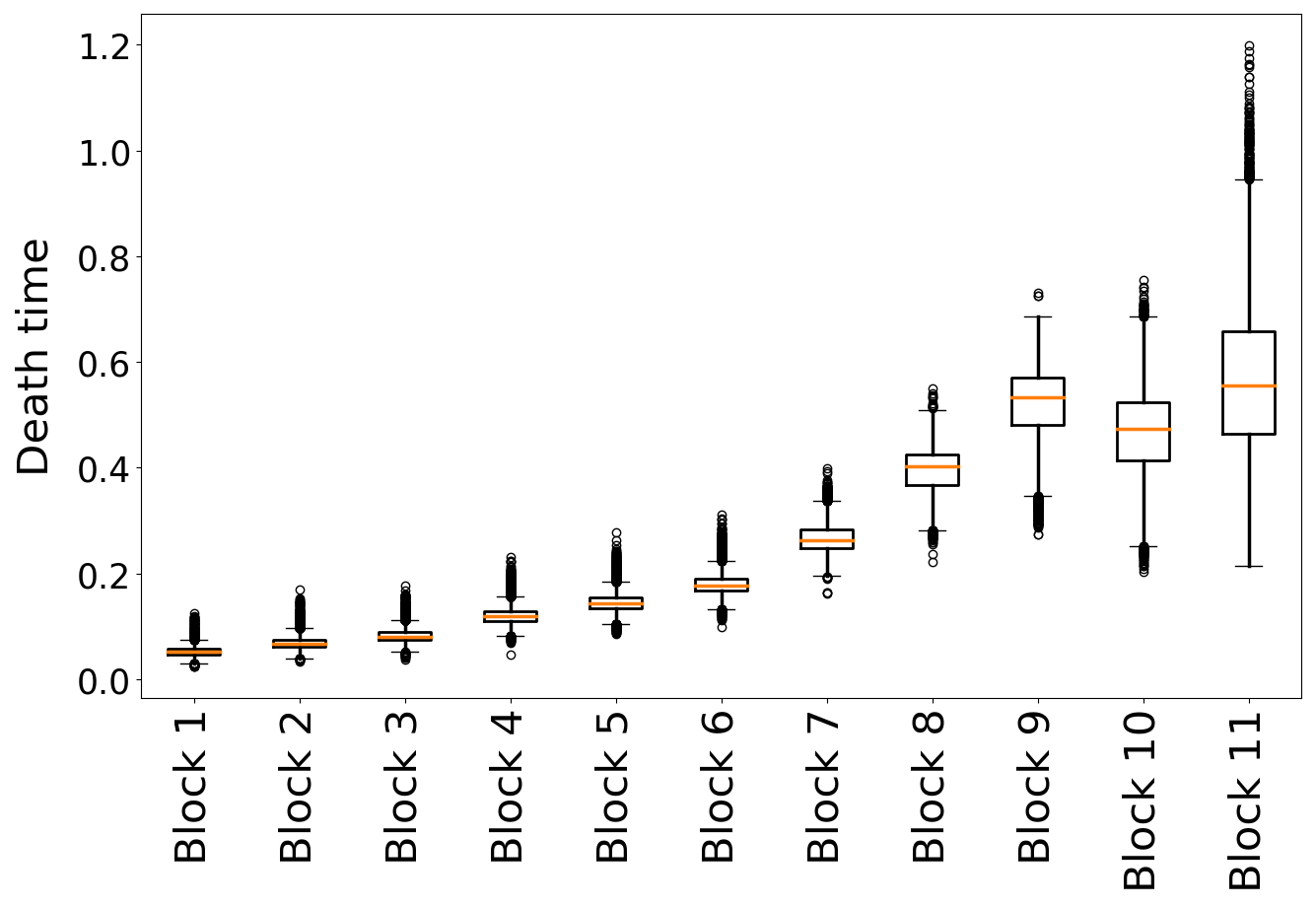}
    }
    \subfloat[Alive time]{
        \label{fig:char-vit-alive-2}
        \includegraphics[width=0.475\linewidth]{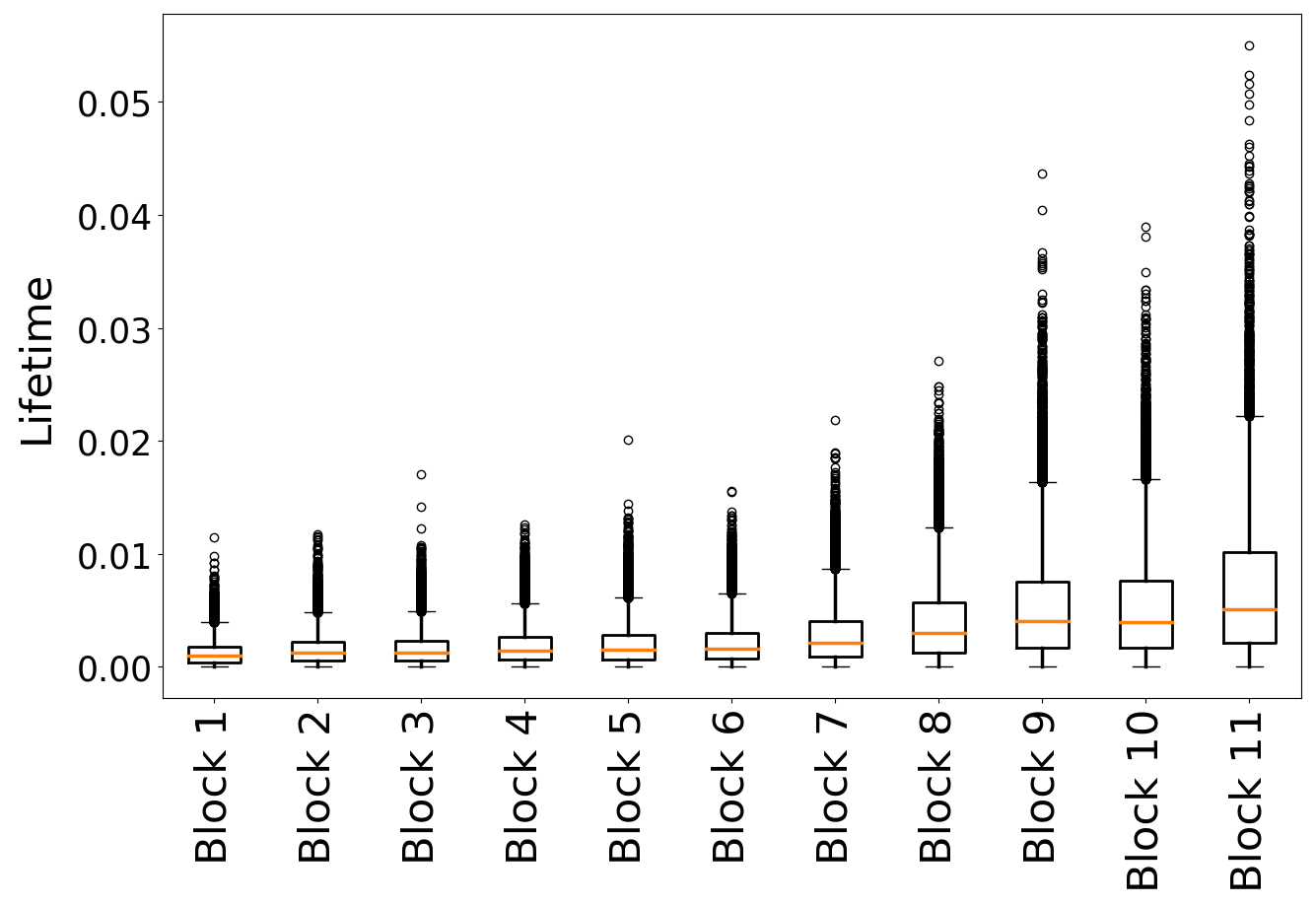}
    }
  \caption{\centering Boxplots of the times of death and life for $H_2$ homology in the ViT network.}
  \label{fig:char-vit-death_alive}
\end{figure}
To further investigate homological features in ViT, we study their death time and the lifetime in each block.
These results are reported in \Cref{fig:char-vit-death_alive}.
First, we observe boxplots of the time of death of $H_2$ homologies (\Cref{fig:char-vit-death-2}).
These results are similar to those observed for other homology groups.
The $H_2$ features die very quickly in the early blocks and the variance in their time of thread is minimal, reflected by interquartile ranges.
However, the interquartile ranges change with the network depth.
In particular, we observe that the times of deaths, as well as their variance, increase till at least \texttt{Block 9}, with the most significant increase occurring between \texttt{Block 7} and \texttt{Block 9}.
We also observe $H_2$ features have a much larger range in \texttt{Block 11} than the preceding blocks. 
This is yet another significant change in the last layer before the multilayer perceptron that classifies the inputs.
It is also worth noting, that unlike in convolutional networks, outliers in ViT representations occur on both tails of the distribution.

Boxplots of lifetimes for $H_2$ features are displayed on \Cref{fig:char-vit-alive-2}.
Here, we see an increase in the lifetime of homological features as the network depth progresses.
We also see that the lifetime is close to 0 for most blocks, and especially for the initial ones.
This distribution has a long tail, similarly to other models.

To sum up, we notice that ViT behaves differently than other architectures that we analyzed in this work.
We observe very characteristic patterns in its number of homological features.
Furthermore, outliers on both tails form a more symmetrical distribution.
One exception to this pattern is the distribution of lifetimes.
Here, however, one-sided tail distribution is expected, as most features live for only a short time.


\subsection{Analysis of bottleneck distances between classes}

In the next set of experiments, we will analyze the structure in the bottleneck distances between classes in the selected models.
We will perform this analysis on test and train datasets.
To this end, we subsample elements from the train dataset to match the number of test elements, as we observed in \Cref{sec:number_homologies} that this is important to obtain correct results.
\begin{figure}[!ht]
    \centering
    \subfloat[VGG19]{
        \label{fig:char-distances-vgg19}
        \includegraphics[width=0.47\linewidth]{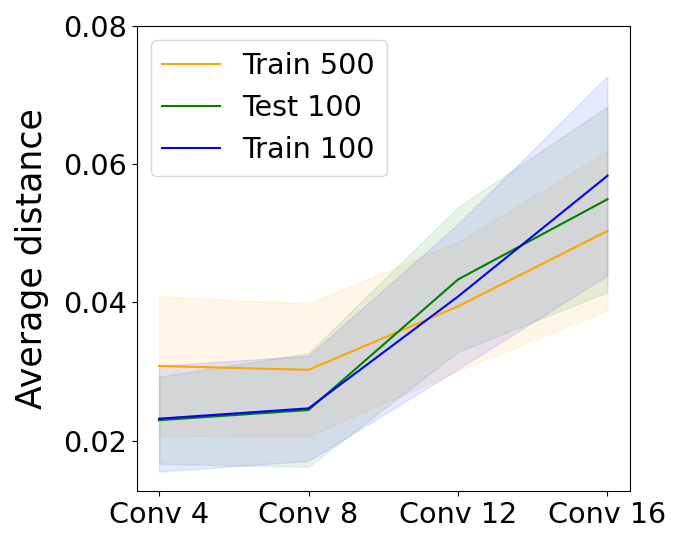}
    }
    \subfloat[ResNet18]{
        \label{fig:char-distances-resnet18}
        \includegraphics[width=0.47\linewidth]{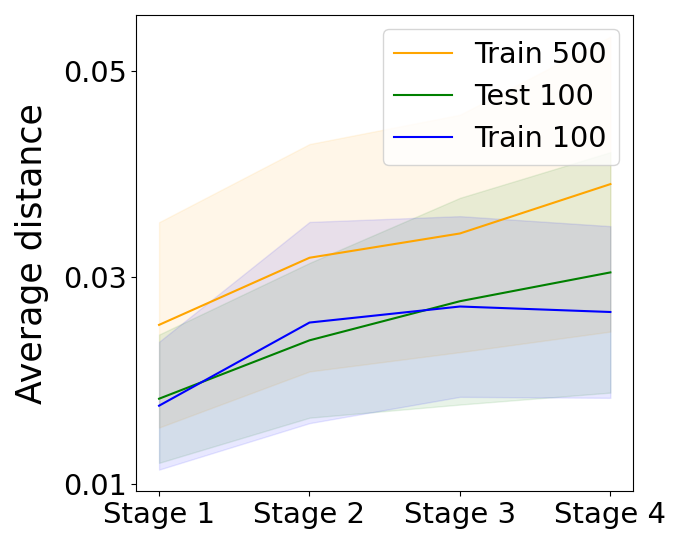}
    }
    \\
    \subfloat[ResNet50]{
        \label{fig:char-distances-resnet50}
        \includegraphics[width=0.47\linewidth]{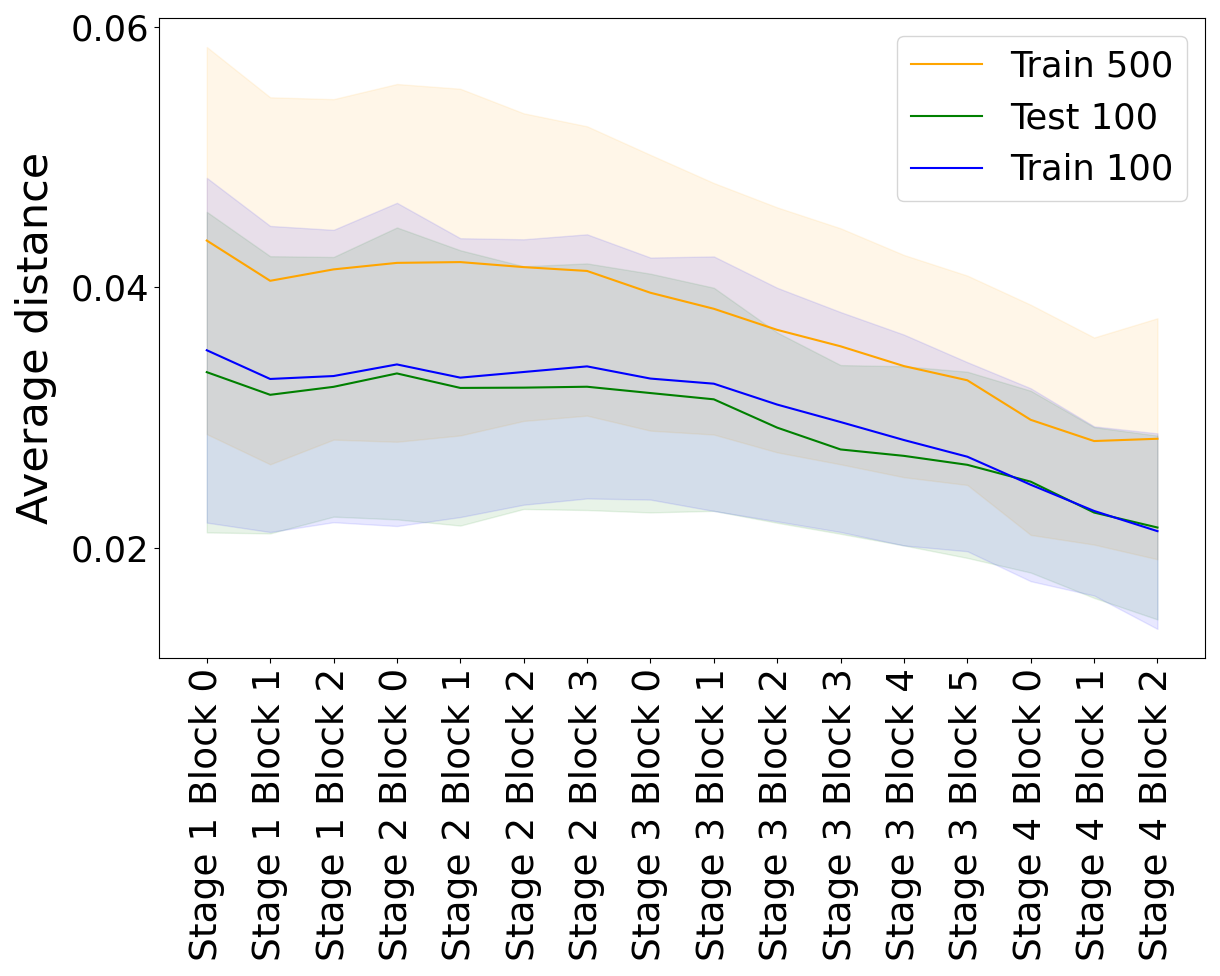}
    }
    \subfloat[ViT]{
        \label{fig:char-distances-vit}
        \includegraphics[width=0.47\linewidth]{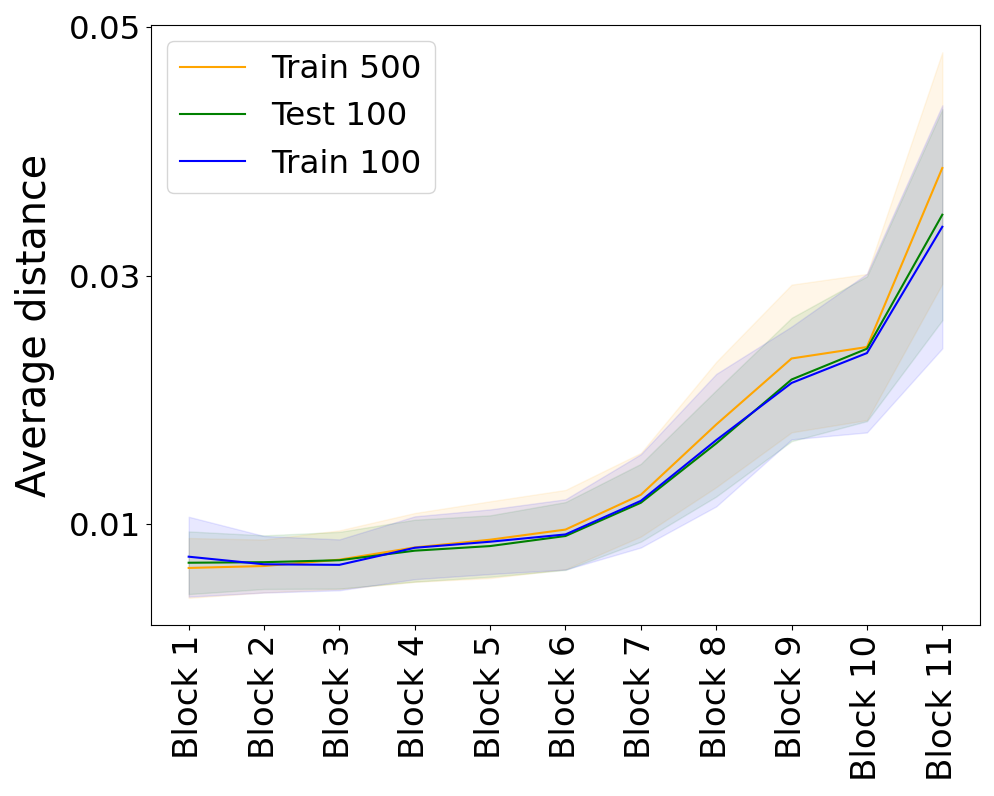}
    }
  \caption{\centering Bottleneck distances between class-specific persistent diagrams for selected networks and layers.}
  \label{fig:char-distances}
\end{figure}

On \Cref{fig:char-distances}, we report the average and standard deviation of bottleneck distances between class-specific persistent diagrams.
The distances were calculated from representations collected from selected network layers.
While we observe some differences between investigated model architectures, in each case, test and train subsets with the matching number of inputs (100) have almost identical average bottleneck distances between classes.
For VGG19 (\Cref{fig:char-distances-vgg19}), the average distance increases with successive convolutional layers. 
That said, there are differences between series that we calculated using 100 and 500 train inputs.
We see that in \texttt{Conv 4} and \texttt{Conv 8} the distance is larger for diagrams calculated using 500 train inputs.
Subsequently, in \texttt{Conv 12}, it matches estimates from other subsets, and then, in \texttt{Conv 16}, becomes smaller.
These differences look marginal, but even minor differences can create some clustering structures.
For ResNet18 (\Cref{fig:char-distances-resnet18}), we also observe that bottleneck distances increase in deeper stages.
The series with 500 train inputs has much larger values here than the other series.
The same observation is valid also for ResNet50 (\Cref{fig:char-distances-resnet50}).
Surprisingly, however, in ResNet50, average bottleneck distances decrease in later stages, instead of increasing.
For both ResNets, the distances for the rest 100 and train 100 series with the matching number of inputs remain almost identical.
Finally, in ViT (\Cref{fig:char-distances-vit}) we observe a slightly different situation with distances in all series being nearly identical.

The exact values for the test and train datasets may indicate that homologies for these two different subsets are not distinct.
This similarity seems to be a good prognostic, as it suggests that the investigated models did not learn by memorizing the training data, but more likely inferred how to extract useful features from images and use this knowledge on the test inputs.
The difference in bottleneck distances when comparing two subsets with different number of elements is related to the properties of the bottleneck distance  (\Cref{sec:number_homologies}).

\begin{figure}[!ht]
    \centering
    \subfloat[ResNet50, \texttt{Stage 3 Block 3}]{
        \label{fig:char-distances_2d-resnet50}
        \includegraphics[width=0.35\linewidth]{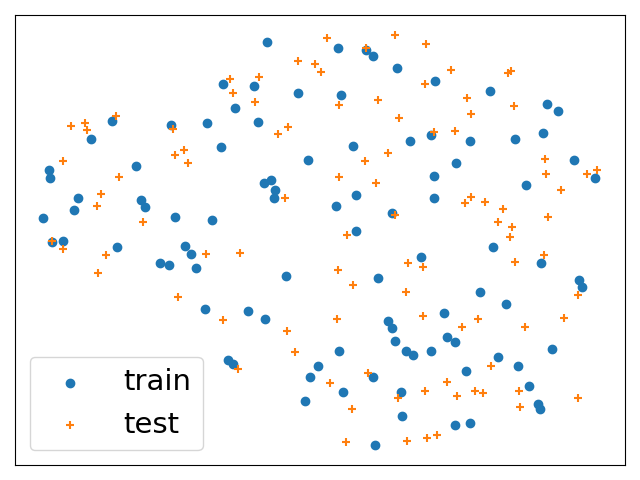}
    }
    \subfloat[ViT, \texttt{Block 7}]{
        \label{fig:char-distances_2d-vit}
        \includegraphics[width=0.35\linewidth]{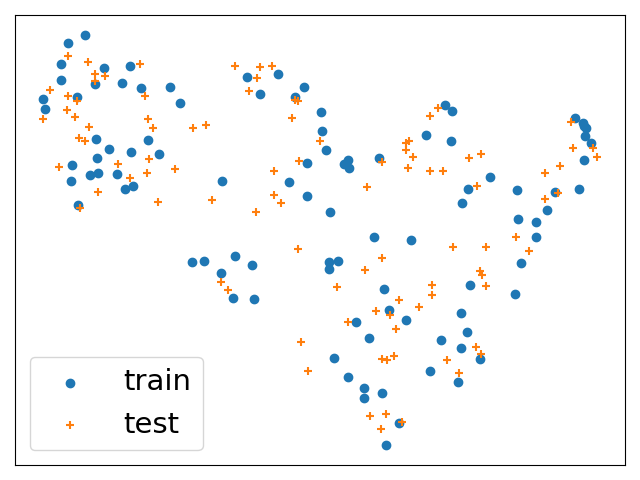}
    }
  \caption{\centering Two-dimensional visualization of bottleneck distances between persistence diagrams for the $H_1$ homology.}
  \label{fig:char-distances_2d}
\end{figure}

To further uncover only possible differences between representations of the test and train inputs, we conducted a 2D visualization of the distributions of class-specific persistent homologies in train and test data.
To this end, we used the UMAP algorithm to embed bottleneck distances between class-specific persistence diagrams.
Outlined visualizations (\Cref{fig:char-distances_2d}) indicate that the representations from both subsets form a single cluster.
We can therefore conclude that for ResNet50 and ViT, the representations for the train and test subsets are topologically similar. 
Reported results are for $H_1$ homology and two deep neural networks, but we observe analogous results for other homologies and architectures.

\subsection{Summary}

To summarize, we noticed many intriguing traits in our analysis of representations in selected architectures. 
Firstly, $H_0$ homology behaves differently than other homologies, mainly because connected components live for a long time, while other homologies die shortly after birth.
Secondly, homology often changes significantly at the last layers of the network, while changes in the middle layers are usually slow and monotonous.
Lastly, the observed results imply a lack of apparent differences between the train and test datasets with respect to the topology of neural representations.
This phenomenon suggests that the networks process known and unknown images in a similar manner, at least from the perspective of the topology of neural representations.

The rest of the observed are specific to some of the models. 
For VGG19 we observe rapid changes in topology between layers and a larger number of homological features than in other architectures. 
ResNet18 has some features similar to VGG19, but we also observe differences, the most important of which is an increase in the number of features in \texttt{Stage 4}.
Comparison between ResNet50 and ResNet18, reveals that the former behaves similarly to the latter, but with a slower rate of change between layers.
That said, there is one notable difference between the two models, visible in the bottleneck distances.
In ResNet18 these distances tend to increase with network depth, while in ResNet50 they instead decrease (\Cref{fig:char-distances}).

Our last analyzed model, namely ViT, behaves in many ways differently than the VGG and ResNet networks.
In ViT, the number of homological features changes in a way not seen in the other models.
Along with that, the death time of ViT's points creates a symmetric distribution, in contrast to convolutional models where we observe single-tailed distributions.
We also do not observe differences in bottleneck distances when different numbers of points are to generate persistent diagrams.
Such differences were seen in other models.
For the transformer architecture, we also observe more long-living $H_1$ and $H_2$ features.
We have some potential hypotheses as to why this happens.
For example, ViT is the only finetuned model in our experiments.
The other models were trained from scratch.
ViT model gives the best accuracy on the test dataset.
ViT also has non-convolutional architecture.
An interesting line of future research could therefore focus on investigating which of these reasons, if any, is responsible for the unique topological structure we observe in ViT.

\newpage
\section{Effects of finetuning on topology of representations}
\label{the_influcence}

\begin{figure}[!ht]
    \centering
    \subfloat[random]{
        \label{fig:rpf-resnet50-diagrams-random}
        \includegraphics[width=0.306\linewidth]{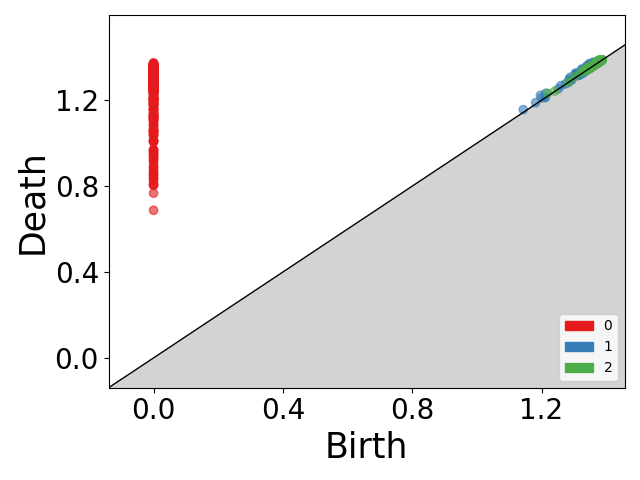}
    }
    \subfloat[pre-trained]{
        \label{fig:rpf-resnet50-diagrams-pretrained}
        \includegraphics[width=0.306\linewidth]{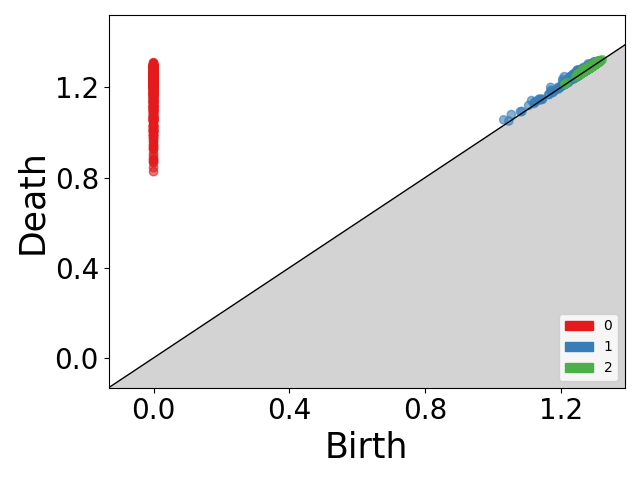}
    }
    \subfloat[finetuned]{
        \label{fig:rpf-resnet50-diagrams-finetuned}
        \includegraphics[width=0.306\linewidth]{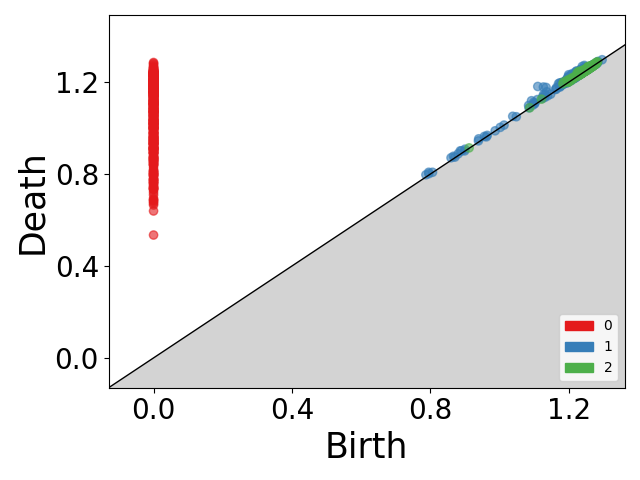}
    }
  \caption{\centering Persistence diagrams for the mouse class in random, pre-trained, and finetuned Resnet50 model. Representations were extracted from the \texttt{Stage 3 Block 3} layer.}
  \label{fig:rpf-resnet50-diagrams}
\end{figure}

We will now explore topological differences in neural representations that result from finetuning a neural network to solve a new classification task.
To this end, we calculated class-specific persistence diagrams for the Resnet50 models with random weights, weights pre-trained on the ImageNet dataset, and the same weights after finetuning on the CIFAR100 dataset. 
We report representative persistence diagrams from the middle layer on \Cref{fig:rpf-resnet50-diagrams}.
Note that there are clear differences between these diagrams, i.e., the number of points on diagrams and their placement differ for each model.
For $H_0$ homology points on the diagrams for random weights (\Cref{fig:rpf-resnet50-diagrams-random}), concentrate in a larger range compared to the pre-trained model (\Cref{fig:rpf-resnet50-diagrams-pretrained}) and a smaller range compared to the finetuned model (\Cref{fig:rpf-resnet50-diagrams-finetuned}).
In the finetuned model we also observe a different placement of $H_1$ features than in the other two models (points are in the 0.8-1.2 interval on the Birth axis; in random and pre-trained model diagrams points concentrate around 1.2 on the Birth axis).

\begin{figure}[!htbp]
    \centering
    \subfloat[\texttt{Stage 1 Block 1}, $H_0$]{
        \label{fig:rpf-resnet50-stage1_block1_h0}
        \includegraphics[width=0.30\linewidth]{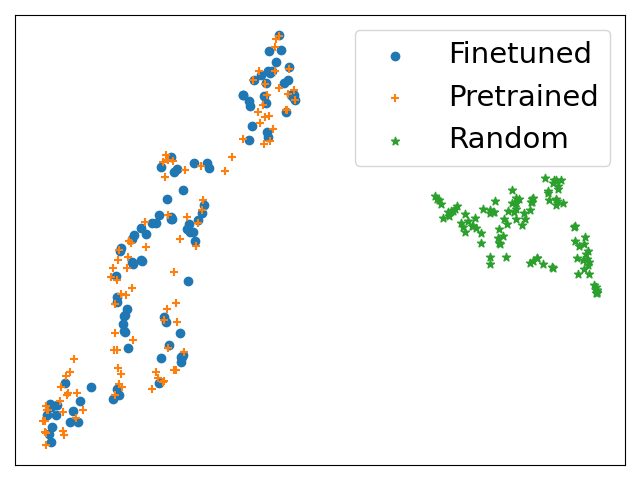}
    }
    \subfloat[\texttt{Stage 1 Block 1}, $H_1$]{
        \label{fig:rpf-resnet50-stage1_block1_h1}
        \includegraphics[width=0.30\linewidth]{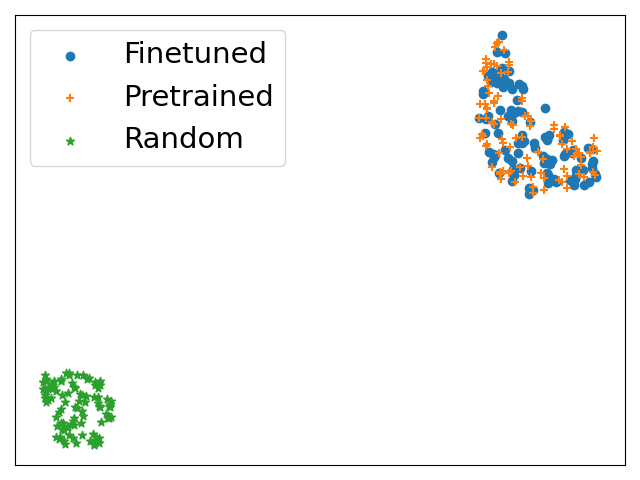}
    }
    \subfloat[\texttt{Stage 1 Block 1}, $H_2$]{
        \label{fig:rpf-resnet50-stage1_block1_h2}
        \includegraphics[width=0.30\linewidth]{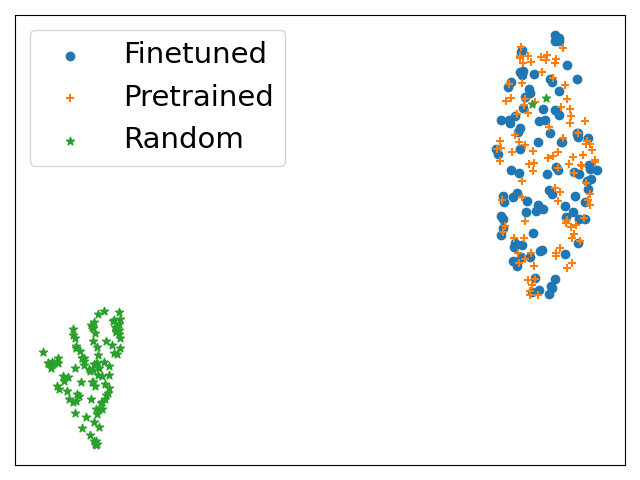}
    }
    \\
    \subfloat[\texttt{Stage 2 Block 2}, $H_0$]{
        \label{fig:rpf-resnet50-stage2_block2_h0}
        \includegraphics[width=0.30\linewidth]{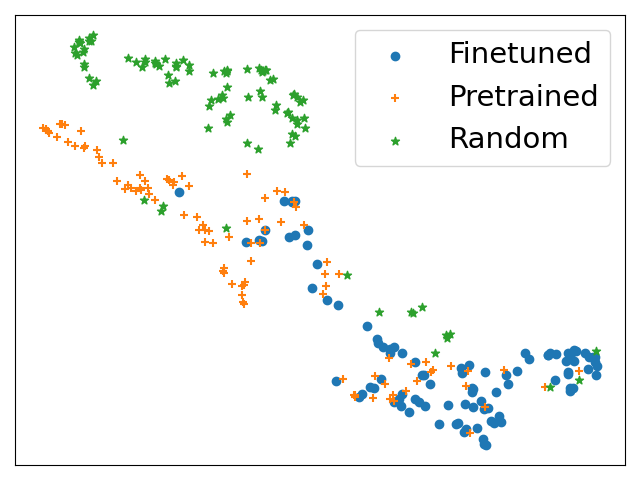}
    }
    \subfloat[\texttt{Stage 2 Block 2}, $H_1$]{
        \label{fig:rpf-resnet50-stage2_block2_h1}
        \includegraphics[width=0.30\linewidth]{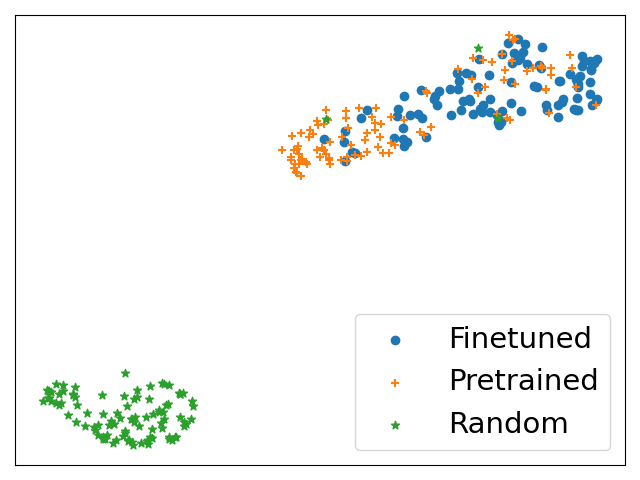}
    }
    \subfloat[\texttt{Stage 2 Block 2}, $H_2$]{
        \label{fig:rpf-resnet50-stage2_block2_h2}
        \includegraphics[width=0.30\linewidth]{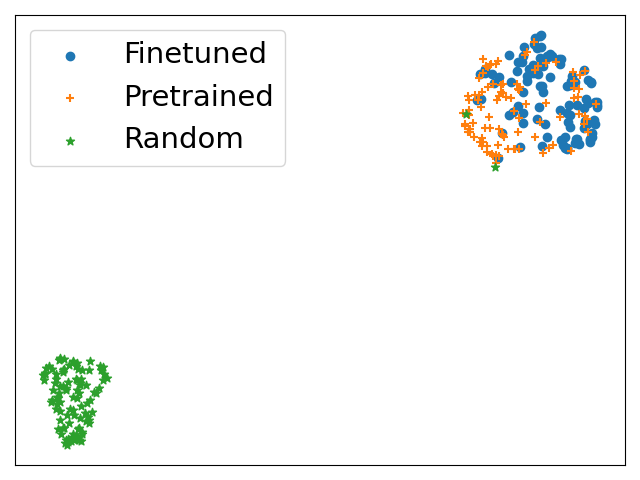}
    }
    \\
    \subfloat[\texttt{Stage 3 Block 3}, $H_0$]{
        \label{fig:rpf-resnet50-stage3_block3_h0}
        \includegraphics[width=0.30\linewidth]{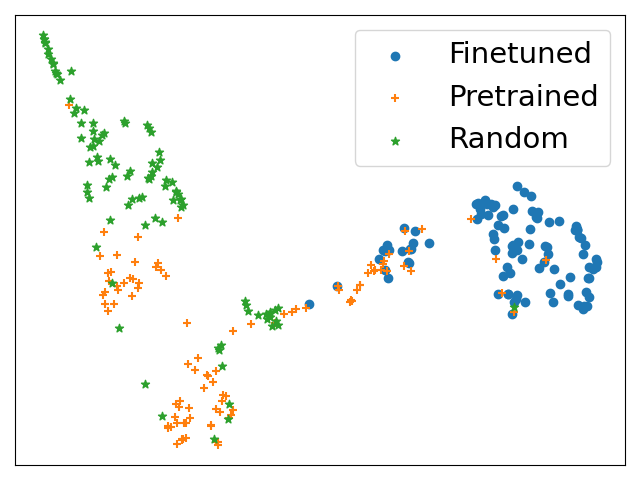}
    }
    \subfloat[\texttt{Stage 3 Block 3}, $H_1$]{
        \label{fig:rpf-resnet50-stage3_block3_h1}
        \includegraphics[width=0.30\linewidth]{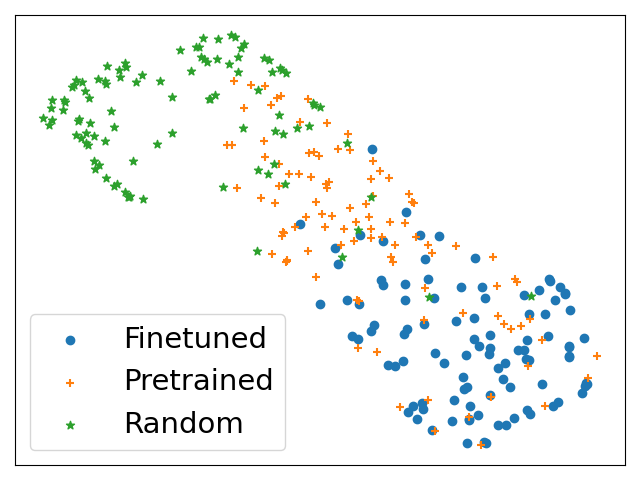}
    }
    \subfloat[\texttt{Stage 3 Block 3}, $H_2$]{
        \label{fig:rpf-resnet50-stage3_block3_h2}
        \includegraphics[width=0.30\linewidth]{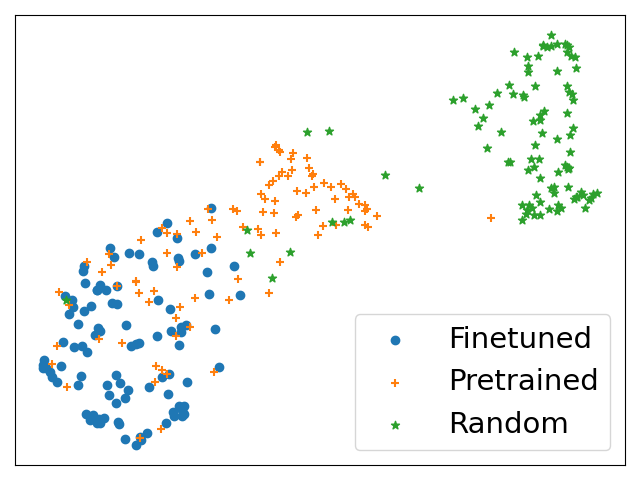}
    }
    \\
    \subfloat[\texttt{Stage 4 Block 2}, $H_0$]{
        \label{fig:rpf-resnet50-stage4_block2_h0}
        \includegraphics[width=0.30\linewidth]{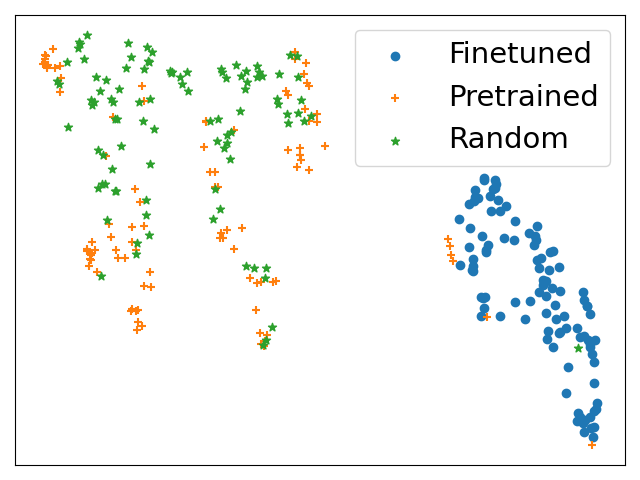}
    }
    \subfloat[\texttt{Stage 4 Block 2}, $H_1$]{
        \label{fig:rpf-resnet50-stage4_block2_h1}
        \includegraphics[width=0.30\linewidth]{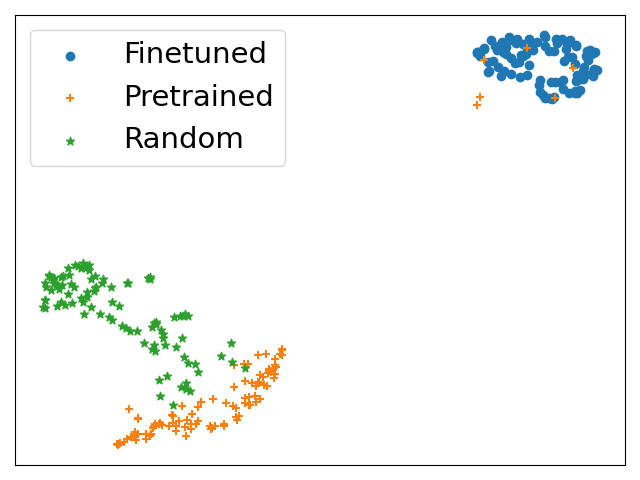}
    }
    \subfloat[\texttt{Stage 4 Block 2}, $H_2$]{
        \label{fig:rpf-resnet50-stage4_block2_h2}
        \includegraphics[width=0.30\linewidth]{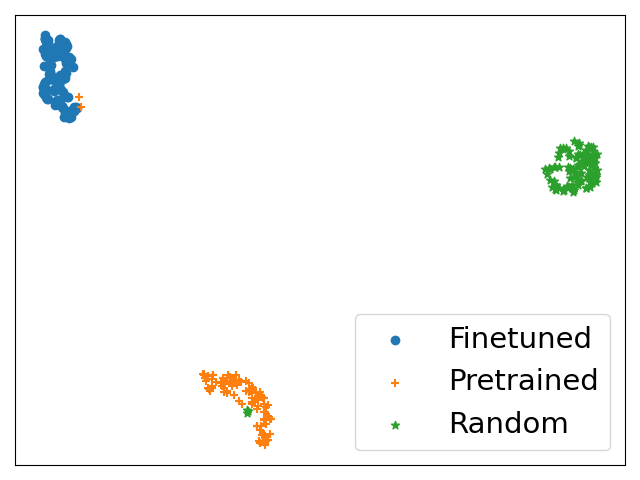}
    }
  \caption{\centering Two-dimensional visualization of the distribution of persistence diagrams for random, pre-trained, and finetuned ResNet50 models.}
  \label{fig:rpf-resnet50}
\end{figure}
Two-dimensional embeddings of the calculated bottleneck distances and reducing dimensionality to 2D are reported on \Cref{fig:rpf-resnet50}. 
We extracted neural activations from a middle block in each ResNet stage except for \texttt{Stage 4} where we extracted activations from the last block. 
We observe that points form clusters in all diagrams.
Depending on the stage and homology group, there are three, two, or, in one case (\Cref{fig:rpf-resnet50-stage3_block3_h1}) one cluster.
Also, even in the one-cluster diagram, points are grouped according to the source model.

Starting from \texttt{Stage 1 Block 1} (\Cref{fig:rpf-resnet50-stage1_block1_h0,fig:rpf-resnet50-stage1_block1_h1,fig:rpf-resnet50-stage1_block1_h2}) we see two clusters, one containing points from the model with random weights and another containing points from the pre-trained and the finetuned model. 
This observation indicates a similarity in neural representations (at this network depth) between pre-trained and finetuned models, and differences from a random weight model. 
The following plots come from \texttt{Stage 2 Block 2} (\Cref{fig:rpf-resnet50-stage2_block2_h0,fig:rpf-resnet50-stage2_block2_h1,fig:rpf-resnet50-stage2_block2_h2}).
Here, we observe a different situation: results from the $H_0$ homology differ from the rest of the analyzed homologies. 
Specifically, we observe three clusters for $H_0$ homology (\Cref{fig:rpf-resnet50-stage2_block2_h0}).
Each cluster contains mainly points from a single model. 
For $H_1$ and $H_2$ homologies (\Cref{fig:rpf-resnet50-stage2_block2_h2} and \Cref{fig:rpf-resnet50-stage2_block2_h2}) we observe a situation similar to the one observed in \texttt{Stage 1 Block 1}, i.e. there are two clusters.
That said in contrast to \texttt{Stage 1}, points from the finetuning model are starting to separate from the initially mixed cluster.
In other words, we see that as the computation in the model progresses, differences in neural representations between fine-tuned and pre-trained networks become larger.
Activations coming from \texttt{Stage 3 Block 3} (\Cref{fig:rpf-resnet50-stage3_block3_h0,fig:rpf-resnet50-stage3_block3_h1,fig:rpf-resnet50-stage3_block3_h2}) represents the middle of the network depth.
For $H_0$ homology (\Cref{fig:rpf-resnet50-stage3_block3_h0}), we observe here a similar case as in the previous stage.
When looking at $H_1$ homology (\Cref{fig:rpf-resnet50-stage3_block3_h1}), we see one cluster with points from different models clearly separated.
For the $H_2$ homology (\Cref{fig:rpf-resnet50-stage3_block3_h2}), we see a situation similar to those observed in the previous stage, with points from the finetuned and the pre-trained model starting to form distinct clusters.
This splitting of clusters continues up to the \texttt{Stage 4 Block 2}, where in $H_0$ homology (\Cref{fig:rpf-resnet50-stage4_block2_h0}), the finetuned model is separated from the two other models.
Additionally, we notice that points from the pre-trained model are scattered into several groups in the vicinity of the cluster corresponding to the random weights model. 
A different situation appears in $H_1$ and $H_2$ homology groups (\Cref{fig:rpf-resnet50-stage4_block2_h1,fig:rpf-resnet50-stage4_block2_h2}).
For $H_2$, we see 3 distinct clusters, each containing points from a single model.
For $H_1$ we technically observe two clusters, but with a strong separation between the pre-trained and the random model.
Looking at these charts, we see that as the network's processing progresses, there are bigger and bigger differences between representations in random, pretrained, and finetuned models.

\begin{figure}[!htbp]
    \centering
    \subfloat[\texttt{Block 2}, $H_0$]{
        \label{fig:rpf-vit-block2_h0}
        \includegraphics[width=0.306\linewidth]{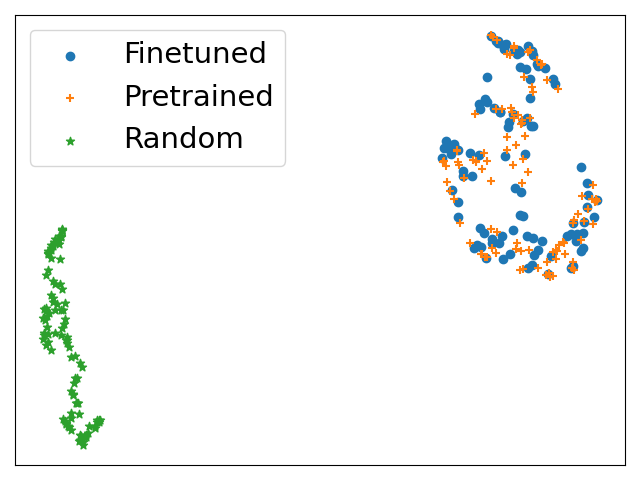}
    }
    \subfloat[\texttt{Block 2}, $H_1$]{
        \label{fig:rpf-vit-block2_h1}
        \includegraphics[width=0.306\linewidth]{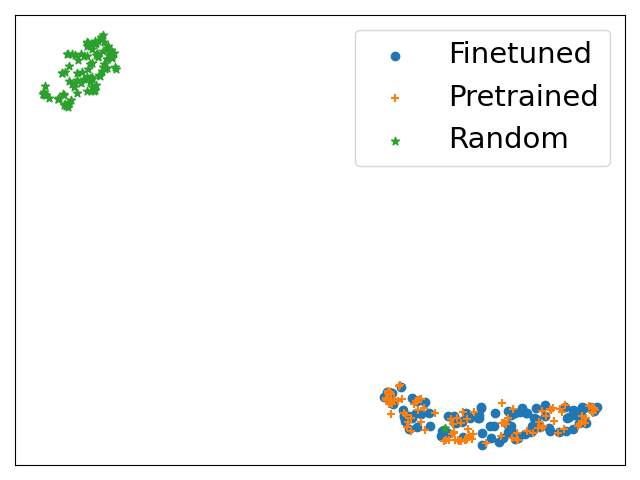}
    }
    \subfloat[\texttt{Block 2}, $H_2$]{
        \label{fig:rpf-vit-block2_h2}
        \includegraphics[width=0.306\linewidth]{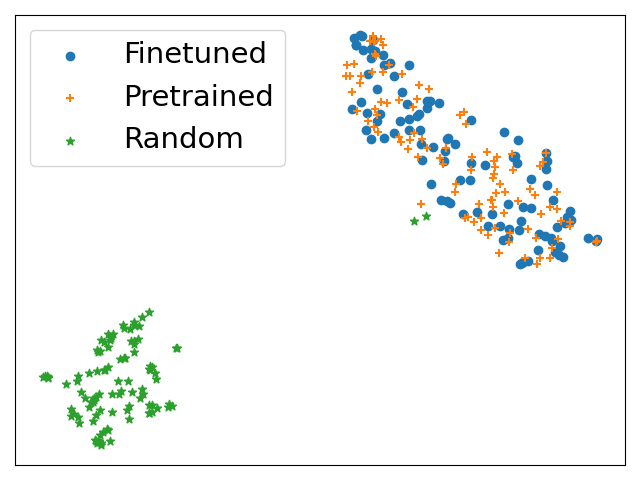}
    }
    \\
    \subfloat[\texttt{Block 5}, $H_0$]{
        \label{fig:rpf-vit-block5_h0}
        \includegraphics[width=0.306\linewidth]{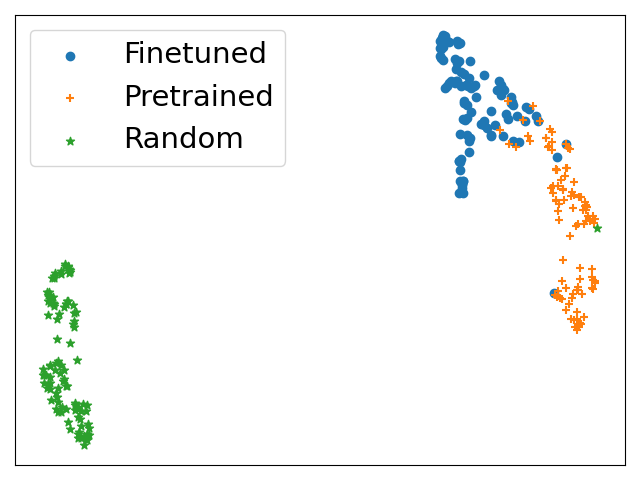}
    }
    \subfloat[\texttt{Block 5}, $H_2$]{
        \label{fig:rpf-vit-block5_h1}
        \includegraphics[width=0.306\linewidth]{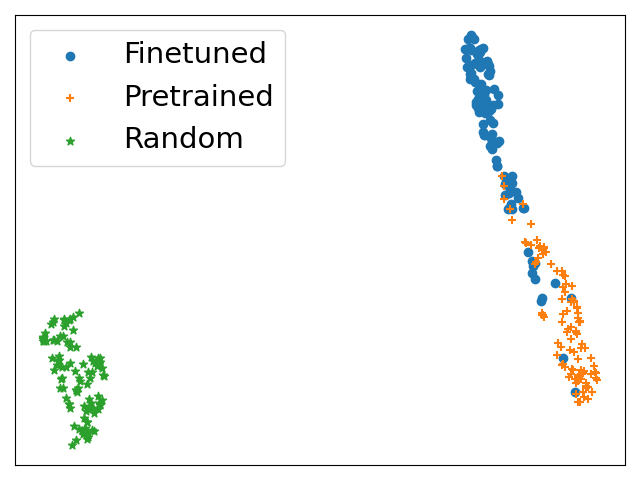}
    }
    \subfloat[\texttt{Block 5}, $H_2$]{
        \label{fig:rpf-vit-block5_h2}
        \includegraphics[width=0.306\linewidth]{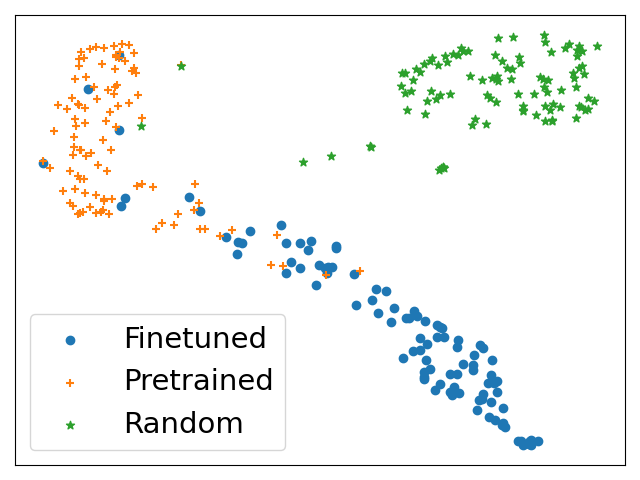}
    }
    \\
    \subfloat[\texttt{Block 7}, $H_0$]{
        \label{fig:rpf-vit-block7_h0}
        \includegraphics[width=0.306\linewidth]{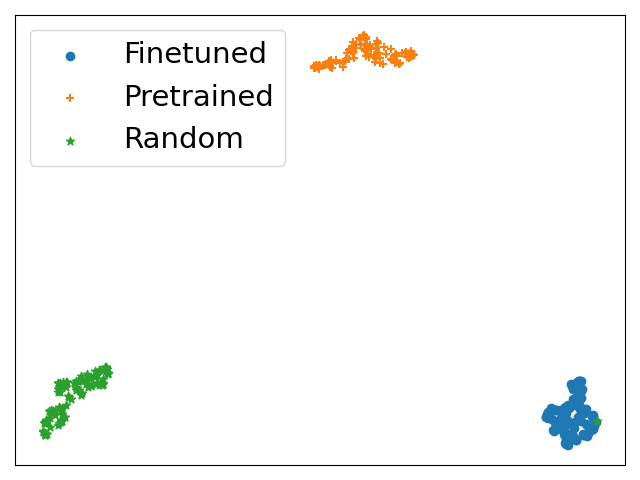}
    }
    \subfloat[\texttt{Block 7}, $H_1$]{
        \label{fig:rpf-vit-block7_h1}
        \includegraphics[width=0.306\linewidth]{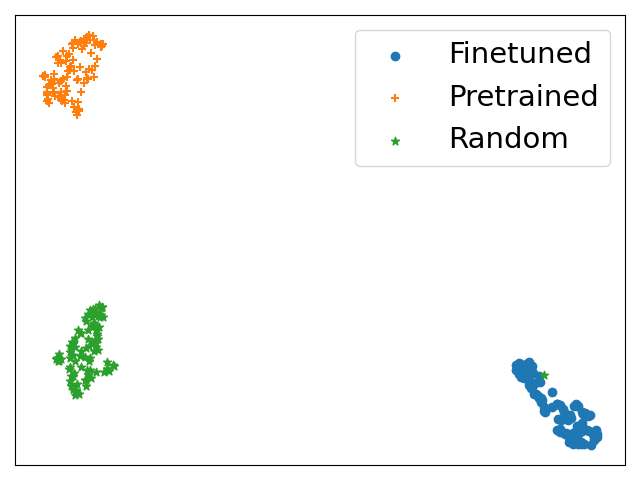}
    }
    \subfloat[\texttt{Block 7}, $H_2$]{
        \label{fig:rpf-vit-block7_h2}
        \includegraphics[width=0.306\linewidth]{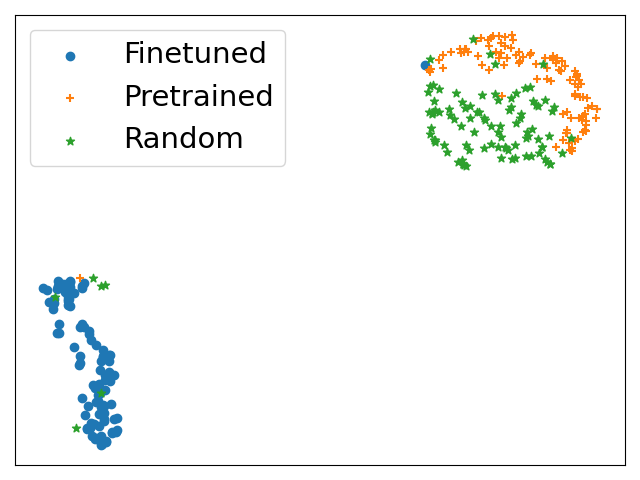}
    }
    \\
    \subfloat[\texttt{Block 11}, $H_0$]{
        \label{fig:rpf-vit-block11_h0}
        \includegraphics[width=0.306\linewidth]{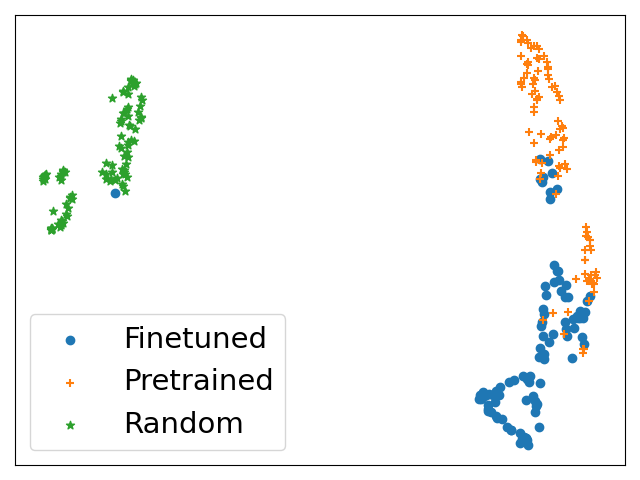}
    }
    \subfloat[\texttt{Block 11}, $H_1$]{
        \label{fig:rpf-vit-block11_h1}
        \includegraphics[width=0.306\linewidth]{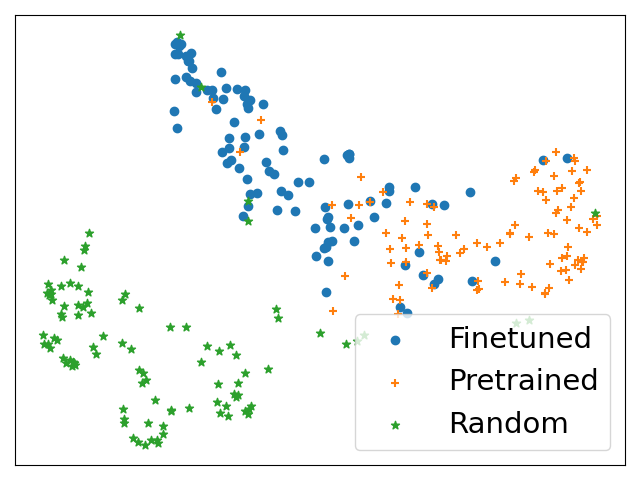}
    }
    \subfloat[\texttt{Block 11}, $H_2$]{
        \label{fig:rpf-vit-block11_h2}
        \includegraphics[width=0.306\linewidth]{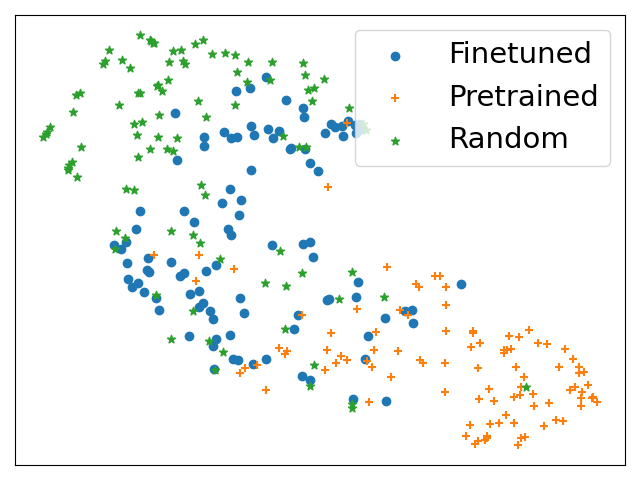}
    }
  \caption{\centering Two-dimensional visualization of the distribution of persistence diagrams for random, pre-trained, and finetuned ViT models.}
  \label{fig:rpf-vit}
\end{figure}
We conducted the same experiment for the Vision Transformer model.
The results are reported on \Cref{fig:rpf-vit}.
For \texttt{Block 2}, located at the beginning of the model, we see the same behavior as in Resnet50:
clusters with points from the random weights model and a second cluster with points from the other two models.
The difference from Resnet50 is that we observe this structure in all homologies (\Cref{fig:rpf-vit-block2_h0,fig:rpf-vit-block2_h1,fig:rpf-vit-block2_h2}).
For \texttt{Block 5} (\Cref{fig:rpf-vit-block5_h0,fig:rpf-vit-block5_h1,fig:rpf-vit-block5_h2}), we see that the cluster containing points from the finetuned and the pre-trained model starts to separate into two groups. 
Once again, this is observed for all homologies. 
This separation is finalized in \texttt{Block 7} for $H_0$ and $H_1$ homologies (\Cref{fig:rpf-vit-block7_h0,fig:rpf-vit-block7_h1}), where we see a clear segregation of the representations from the three models.
For $H_2$ (\Cref{fig:rpf-vit-block7_h2}), we observe a different structure, with points from the finetuned and the random weights models forming a single cluster.
However, inside the cluster points are well segregated.
Finally, we observe three clusters for the $H_0$ homology in the last  ViT block.
However, there is some mix between representations from the pre-trained and the finetuned model (\Cref{fig:rpf-vit-block11_h0}).
For $H_1$ and $H_2$ homologies we observe one cluster with segregated points (\Cref{fig:rpf-vit-block11_h1,fig:rpf-vit-block11_h2}).

After analyzing two different architectures, we observed that activations from the finetuned and pre-trained models are topologically similar at the initial layers.
However, this changes in deeper layers, where finetuned and pre-trained models become different from each other. 
We also observed changes at the last layers, where clusters are not as much separated as in the previous layers.
That said, even when representations from the three models form one cluster they are clearly segregated in that cluster.
These results suggest that during model finetuning most of changes that affect neural representations happen from the middle layers of the network.
This finding is consistent with the results observed in \citet{kornblith2019similarity}, where authors state that early network layers learn more similar representations, whereas later layers learn more distinct representations.
As expected, a model with random has different topological features than pre-trained and finetuned models.
These results seem logical, as the model with random weights differs significantly from the two others, which have weights that minimize the training loss,
Also, as the finetuned model evolved from the pre-trained one, they are more similar to each other,  and more often create a single cluster.
Differences between these models are also visible on persistence diagrams that are reported on \Cref{fig:rpf-resnet50-diagrams}.

\newpage
\section{Where networks change homology most rapidly}

In the last experiment, we will show how the topology of neural representations changes across the network depth.
We computed the average bottleneck distance between the input of selected layers and their corresponding, or output of another layer.
We presented these results on heatmaps that illustrate the magnitude of the input-output change. In this way we can quantify how rapidly network layers change the topology of neural representations.

\begin{figure}[!ht]
    \centering
    \subfloat[$H_0$]{
        \label{fig:change-vgg19-heatmap-0}
        \includegraphics[width=0.308\linewidth]{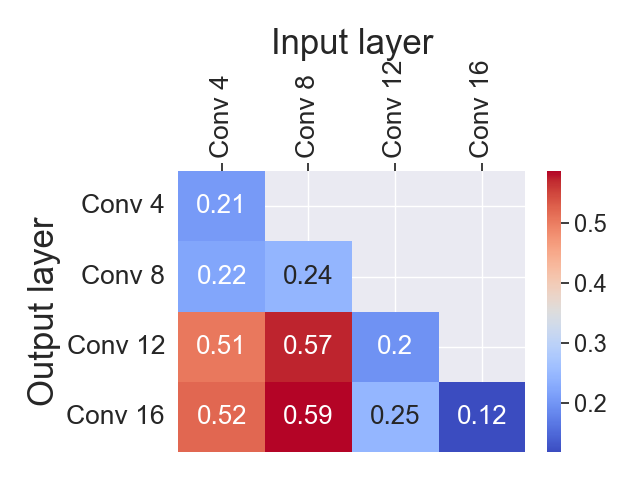}
    }
    \subfloat[$H_1$]{
        \label{fig:change-vgg19-heatmap-1}
        \includegraphics[width=0.308\linewidth]{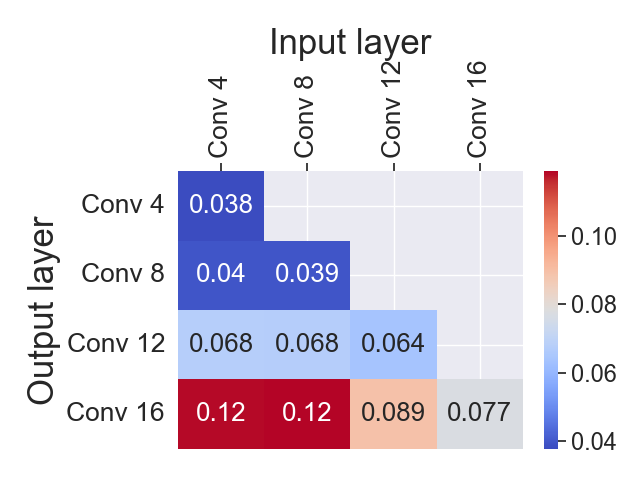}
    }
    \subfloat[$H_2$]{
        \label{fig:change-vgg19-heatmap-2}
        \includegraphics[width=0.308\linewidth]{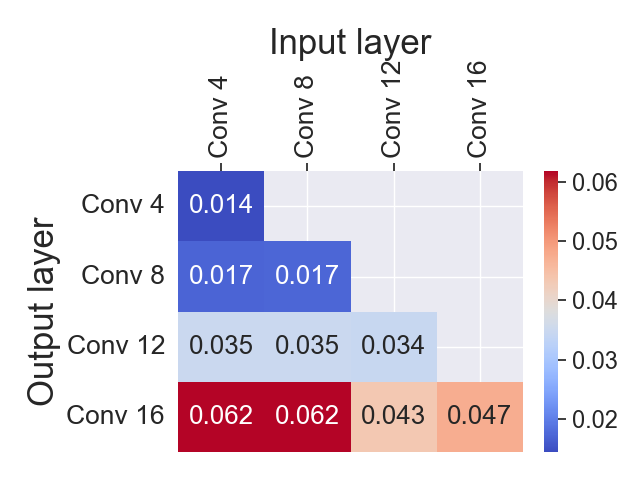}
    }
  \caption{\centering Heatmaps with average bottleneck distances between input and output of selected VGG19 layers.}
  \label{fig:change-vgg19-heatmap}
\end{figure}

Heatmaps for the VGG19 network are presented on \Cref{fig:change-vgg19-heatmap}.
In the results for $H_0$ homology (\Cref{fig:change-vgg19-heatmap-0}) we observe that the smallest distances occur when we compare input and output representations of the same layer (diagonal line).
We see that values on the diagonal are roughly equal for most layers, except for the \texttt{Conv 16}, where the distance is about two times smaller.
Bottleneck distances between representations from different layers are much larger, especially between the input of \texttt{Conv 8} and the output of \texttt{Conv 16}.
A slightly different picture arise for $H_1$ and $H_2$ homologies (\Cref{fig:change-vgg19-heatmap-1,fig:change-vgg19-heatmap-2}).
We see here that distances between \texttt{Conv 4} and \texttt{Conv 8} are relatively small, compared to distances in \texttt{Conv 12} or between \texttt{Conv 4} and \texttt{Conv 16}.
We also observe that homology changes in $H_1$ and $H_2$ follow a common pattern.
When looking at the diagonal, we see that distances increase in consecutive layers.
Another observation for $H_1$ and $H_2$ is that the distance from \texttt{Conv 4} to \texttt{Conv 16} is identical to the distance from \texttt{Conv 8} to \texttt{Conv 16}, suggesting that topological changes are somewhat divergent.
Overall, from these heatmaps we can deduce the following: $H_0$ homology differs from the other two homologies; topological changes are occurring across the whole network, but for $H_1$ and $H_2$ they are bigger in the deeper layers. 
Finally, the similarity of homologies decreases with the distance between the layers.

\begin{figure}[!ht]
    \centering
    \subfloat[$H_0$]{
        \label{fig:change-resnet18-heatmap-0}
        \includegraphics[width=0.308\linewidth]{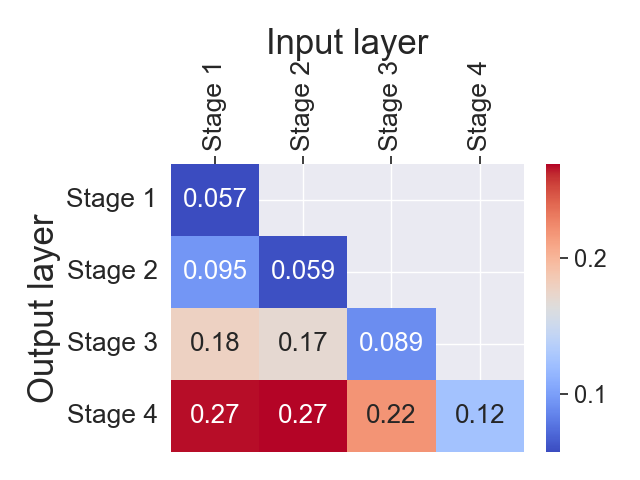}
    }
    \subfloat[$H_1$]{
        \label{fig:change-resnet18-heatmap-1}
        \includegraphics[width=0.308\linewidth]{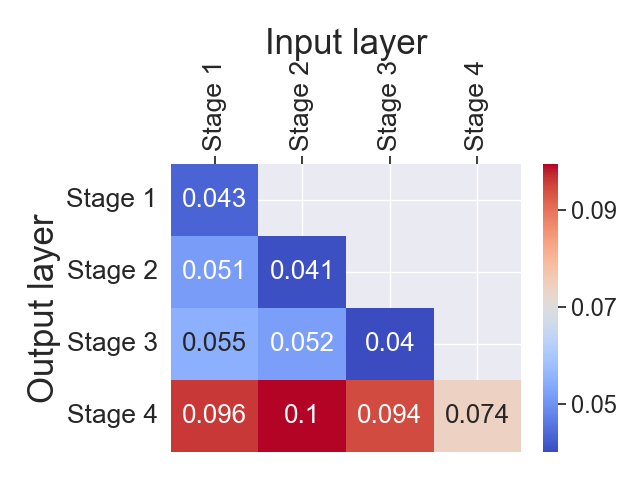}
    }
    \subfloat[$H_2$]{
        \label{fig:change-resnet18-heatmap-2}
        \includegraphics[width=0.308\linewidth]{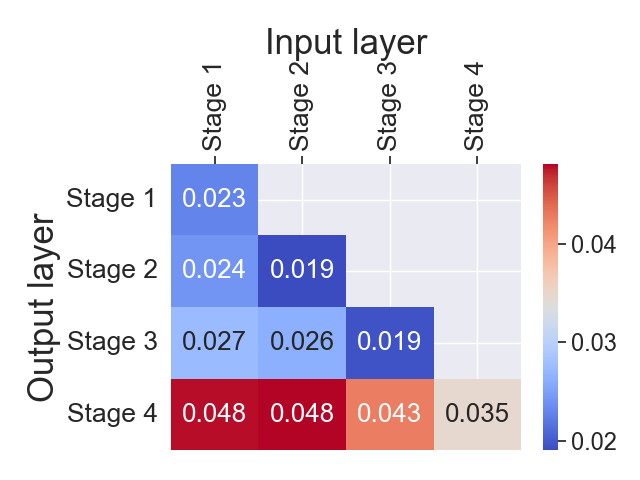}
    }
  \caption{\centering Heatmaps with average bottleneck distance between input and output of selected ResNet18 layers.}
  \label{fig:change-resnet18-heatmap}
\end{figure}
The distribution of distances in the ResNet18 model is displayed on \Cref{fig:change-resnet18-heatmap}.
In the residual model we observe that for all homologies representations change most rapidly in \texttt{Stage 4} (comparing input and output in each stage). 
Moreover, comparison of representations from different layers reveal that, the largest difference is between \texttt{Stage 2} and \texttt{Stage 4}.
Focusing on $H_0$ homology (\Cref{fig:change-resnet18-heatmap-0}), the input-output distances in consecutive stages steadily increase.
In contrast to VGG19, the distance is greater in the last layer than in the previous layers (in VGG19 we observe a decline).
Additionally, we observe that distances from \texttt{Stage 1} to \texttt{Stage 4} and from \texttt{Stage 2} to \texttt{Stage 4} are identical, which means that changes between \texttt{Stage 1} and \texttt{Stage 2} transform persistent homology in a different way than the changes in subsequent stages.
Looking at other homologies (\Cref{fig:change-resnet18-heatmap-1,fig:change-resnet18-heatmap-2}), we see that most of the homology evolution happens in \texttt{Stage 4}.
We also see that the distance between the input and output of \texttt{Stage 1} is slightly bigger than the one in \texttt{Stage 2} and \texttt{Stage 3}, which is a different behavior than what we observed for $H_1$ and $H_2$ in VGG19.

\begin{figure}[!ht]
    \centering
    \subfloat[$H_0$]{
        \label{fig:change-resnet50-heatmap-0}
        \includegraphics[width=0.308\linewidth]{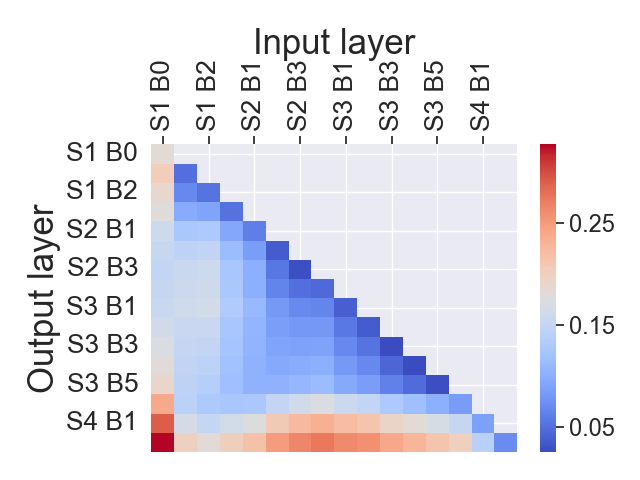}
    }
    \subfloat[$H_1$]{
        \label{fig:change-resnet50-heatmap-1}
        \includegraphics[width=0.308\linewidth]{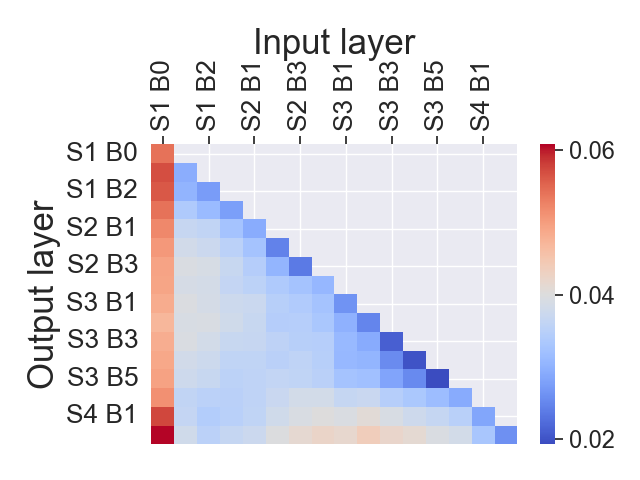}
    }
    \subfloat[$H_2$]{
        \label{fig:change-resnet50-heatmap-2}
        \includegraphics[width=0.308\linewidth]{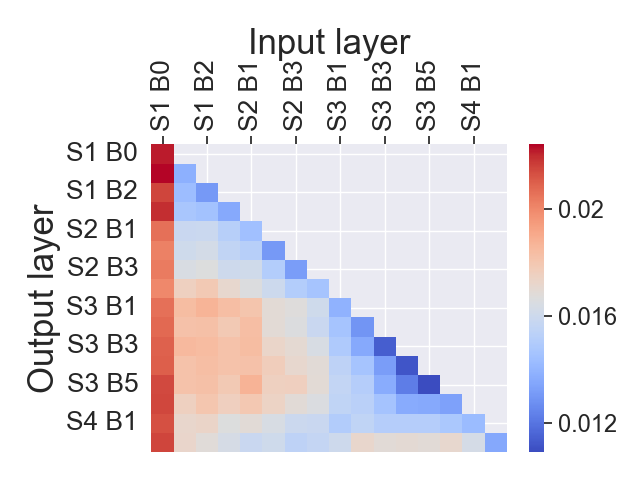}
    }
  \caption{\centering Heatmaps with average bottleneck distance between input and output of selected ResNet50 layers. On axis \texttt{Sn Bk} denotes \texttt{Stage n Block k}.}
  \label{fig:change-resnet50-heatmap}
\end{figure}
Results for ResNet50 are reported on \Cref{fig:change-resnet50-heatmap}. 
In this model we see a more complex pattern.
Note, however, that our analysis in this case is more fine-grained, i.e., covers more layers
Overall, evolution of representations in ResNet50 appears to proceed slower than in ResNet18 and VGG19. 
For $H_0$ homology (\Cref{fig:change-resnet50-heatmap-0}), we see that in \texttt{Stage 1 Block 0} topology changes significantly, especially compared to other layers.
Along with that, we notice a large distance between the blocks of \texttt{Stage 3} and \texttt{Stage 4 Block 2}. 
Interestingly, we observe that in some cases the similarity between two more distant layers is larger than between two closer ones.
This occurs, for example, in \texttt{Stage 1 Block 1}, which is closer to \texttt{Stage 4 Block 0} than \texttt{Stage 3}.
We also see that the distances between input and output representations in \texttt{Stage 3 Block 3-5} are smaller than in other layers.
Distances for $H_1$ homology follow a pattern similar to the one observed for $H_0$ (\Cref{fig:change-resnet50-heatmap-1}).
The large distance between \texttt{Stage 1 Block 0} and all other layers is pronounced even more in this homology.
In $H_2$ homology, the overall pattern is similar to that at $H_0$ and $H_1$ (\Cref{fig:change-resnet50-heatmap-2}).
That said, there is a marked increase in distances between \texttt{Stage 1} and \texttt{Stage 3} layers.

\begin{figure}[!ht]
    \centering
    \subfloat[$H_0$]{
        \label{fig:change-vit-heatmap-0}
        \includegraphics[width=0.308\linewidth]{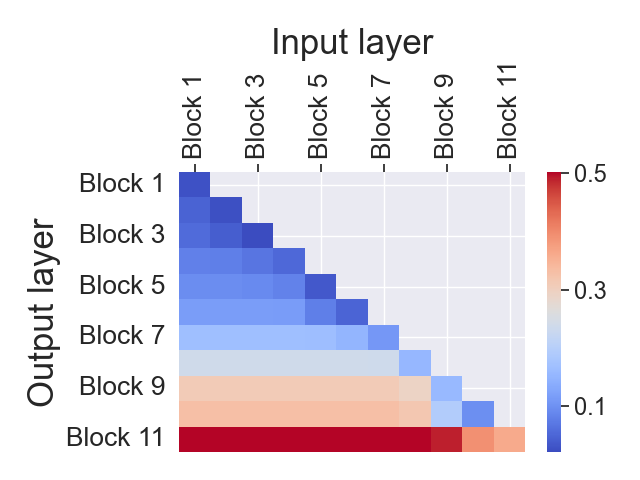}
    }
    \subfloat[$H_1$]{
        \label{fig:change-vit-heatmap-1}
        \includegraphics[width=0.308\linewidth]{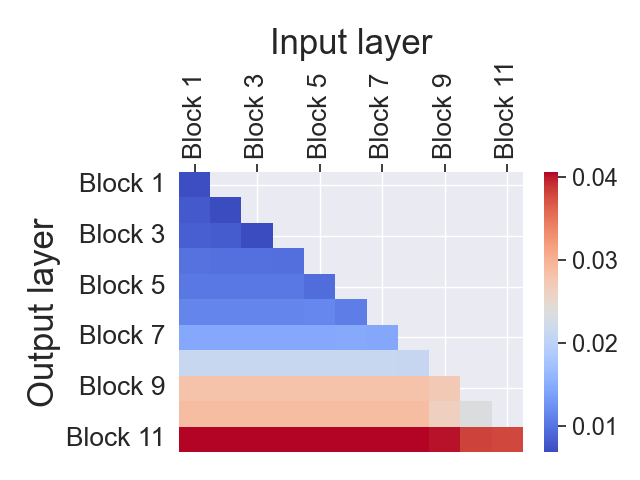}
    }
    \subfloat[$H_2$]{
        \label{fig:change-vit-heatmap-2}
        \includegraphics[width=0.308\linewidth]{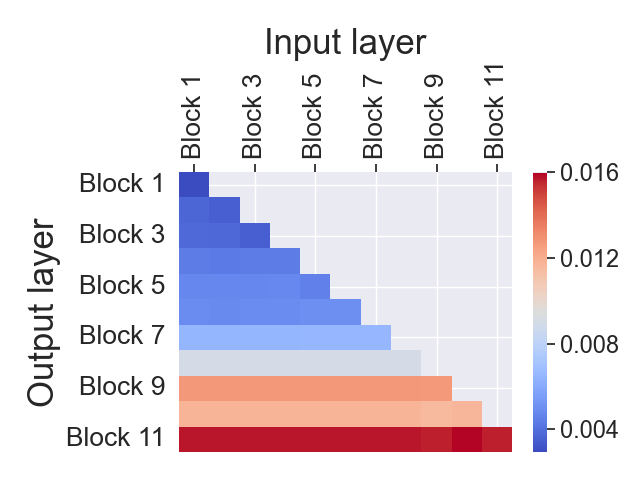}
    }
  \caption{\centering Heatmaps with average bottleneck distance between input and output of selected ViT layers.}
  \label{fig:change-vit-heatmap}
\end{figure}
After analyzing convolutional models, we now focus on ViT.
The bottleneck distances for the transformer are shown on \Cref{fig:change-vit-heatmap}.
Here, we observe a clearly different situation than in convolutional models. 
The distances between the first few blocks are small, and the distances between the initial blocks and the final blocks are particularly high.
Another pattern visible on heatmaps is that the distance between \texttt{Block 1} and \texttt{Block 11} is almost identical to that between \texttt{Block 8} and \texttt{Block 11}.
This shows that each of the blocks performs a different topological transformation.
Looking at the diagonal, we see that most of the topology evolution happens in the last three layers.
Additionally, input-output bottleneck distances grow with successive blocks.
Moving on to the remaining homologies, we notice mostly some minor differences.
The visible differences between homologies occur from \texttt{Block 9} to \texttt{Block 11}.
In general, various homologies create a similar pattern.
Interestingly, in the $H_2$ homology (\Cref{fig:change-vit-heatmap-2}), the distance depends mostly on the target (output) layer and has a nearly constant value regardless of the selection of input representations.
Such a pattern is an exception and does not occur for other homology groups and models.

To summarize the results of our analysis, we find that homology evolves differently in convolutional and transformer-based models.
This observation is consistent with findings in \Cref{characterization} and is one more trait that distinguishes ViT model from the rest.
There are also some prevalent patterns in topology evolution.
For example, the rate of change in the last few layers tends to be higher than in the initial layers (the exception is ResNet50, where change in the first layer is the largest).
Observations also suggest that the similarity between closer layers is not consistently larger than between more distant layers, again with the exception of ResNet50.
Still, in most cases, two layers located further apart in the network architecture tend to be more unsimilar than two closer layers.

\chapter{Related work}

Investigation of neural representations is an area of active research efforts.
In this chapter we will describe selected previous works that touch on investigating neural network via lenses of representations.

We will start our discussion with methods that analyze neural representations without using TDA techniques.
One of important methods that leverage neural representations to analyze deep neural networks is DORA (Data-agnostic Representation Analysis), introduced by \citet{bykov2022dora}.
The authors devised a distance measure between neural representations called Extreme-Activation (EA).
They tested this method in real applications, such as outlier detection.
Their results demonstrate that the EA distance is easy to interpret and is a good tool for detecting outliers.
That said, it has some limitations.
For example, it assumes that the malicious behaviors exhibited by neural networks are not systematic.
Another recent algorithm created to analyze DNNs, called Inverse Recognition (INVERT), was developed by \citet{bykov2024labeling}.
Its purpose is to label neural representations with concepts or compositions of concepts that humans understand.
Inverse Recognition tries to find concepts that maximize an AUC similarity with the representations.
Moreover, it provides statistical tests for confirming that the selected concept is not random.
Inverse Recognition, along with DORA, contributes to advancements in Explainable Artificial Intelligence.

Techniques discussed above analyze neural representations in a point-wise manner.
That said, they treat a representation of a single input from the data set as a point of analysis.
A different approach was employed in \cite{jamroz2023neural}, where authors performed class-wise analysis.
They used a non-parametric hierarchical Bayesian model to calculate class-conditional densities of neural representations.
In this way they found two modes of class-fitting manifested by distinct distribution of representations.
They also found that the two types of classes have different degree of input memorization and adversarial robustness.
The discovery of a group of classes with different features and characteristics influenced work in this thesis as we also embraced class-wise analysis.

Moving to TDA methods, there is prior work on using them to uncover structure in neural representations.
An example method used in this context is the Mapper algorithm.
The Mapper algorithm transforms a set of data points into a graph in a way that retains the structure of the original data.
An example algorithm that uses Mapper to analyze neural representations is TopoAct \citep{rathore2021topoact}.
TopoAct is a visual exploration system that studies topological summaries (e.g., branches, loops) of activation vectors.
Focus on extraction of topological features distinguishes it from standard visualization methods, like t-SNE or UMAP.
TopoAct can analyze deep neural networks that solve different tasks, such as image or text classification.
Because of that, it provides valuable insights into how various input data are processed, and which representations are similar.
The Mapper algorithm was also used by \citet{purvine2023experimental}, where the focus is on convolutional neural networks.
\citeauthor{purvine2023experimental} used Mapper to visualize how neural representations are organized in network layers.
In this way they demonstrated that there is a separation between classes in the last layers of CNN.
\citeauthor{purvine2023experimental} also use a metric called Sliced Wasserstein distance, which is based on persistence homology between layers or models.
The authors found that the same layers of two independently trained networks have smaller distances than different layers within a single network.
They also point out that the obtained results are training invariant. which indicates the presence of homological structures in neural networks.
Consequently, the results are potentially relevant to model interpretation.

Another important study that uses TDA to analyze how topology changes across the model is the work by \citet{naitzat2020topology}.
The authors train many multilayer perceptron models on synthetic and real datasets and then investigate topology in their representations.
The results of these experiments lead to important conclusions. 
First, the topology of data simplifies as it is transformed by a neural network.
This simplification occurs no matter how complicated input data is.
Next, using ReLU activation function leads to faster simplification of the topology of neural representations than using hyperbolic tangent activation.
This faster topology simplification is potentially related to ReLU being a nonhomeomorphic function.
The last conclusion reached by \citeauthor{naitzat2020topology} is that shallow neural networks transform data in a distinct manner.
Specifically, shallow models often operate on changing geometry of data and change topology in the final layers.
In contrast, deep models change topology across the whole depth. 
Conclusions from this work partially align with those obtained in our experiments, where we observe that the most interesting are middle and final layers, while the geometry of the data changes mostly in the initial layers.
This insight is visible, e.g., in \Cref{fig:rpf-vit}, where class separation is not visible in the first few blocks.
That said, contrary to \citeauthor{naitzat2020topology} findings, we found that topology simplification does not occur in ViT (\Cref{fig:char-vit-points}) and for ResNets (\Cref{fig:char-resnet50-points}).
These models, however, were not used in the \citeauthor{naitzat2020topology} study, and are much more complicated.
To sum up, the authors performed an interesting analysis and demonstrated that TDA can provide answers on how neural networks transform data.

Another work focusing on the comparison of neural network representations was carried out by \citet{barannikov2021representation}.
They proposed Representation Topology Divergence (RTD), a topological method for calculating dissimilarity between two point clouds of equal size with a one-to-one correspondence between points.
The method is related to R-Cross-Barcode, which was defined by \citeauthor{barannikov2021representation} and is a good alternative to classic representational similarity methods that do not use TDA, like SVCCA \citep{raghu2017svcca} or CKA \citep{kornblith2019similarity}.
Representation Topology Divergence was surprisingly well correlated with disagreement of neural network predictions, suggesting the need for further investigation.
The method itself is related to topological complexes that were used also in this thesis.
Another work exploring neural network activations with TDA methods is \citet{wheeler2021activation}.
This work presents a summary of the information obtained from a network layer by using persistent homology and activation landscapes.
The authors discovered that topological complexity does not monotonically decrease along with network depth, which agrees with our observations for ViT and ResNets.
Their second finding is that a better-trained network tends to have more topologically complex representations.
In contrast to our work, the studies conducted by \citeauthor{wheeler2021activation} were limited to experiments on MLPs. 
Consequently, their findings should be validated on more complex architectures.
\citet{kushnareva2022betti} leverages Betti numbers in text classification tasks.
Specifically, they computed Betti numbers from the matrix of attention weights in the BERT model.
Next, they demonstrated that text classification learned on the Betti numbers matches the results obtained with the BERT model.
This article shows that TDA methods are also suitable for large architectures, like BERT, and that homological information obtained from neural representations contains a lot of information about the underlying learning task.

Some of the previous works on TDA methods for neural networks use bottleneck distance.
An illustrative example is provided by \cite{perez2021characterizing}.
In contrast to our thesis, this work did not employ neural representations but instead obtained persistent homology by creating a directed graph from neural network weights.
\citeauthor{perez2021characterizing} conducted experiments on some popular datasets concentrating on calculating bottleneck distances for such persistence diagrams.
The main result of this study is that TDA methods abstract away individual network values, giving more abstract information about the network.
Additionally, the authors observed that models with the same architecture, but trained on different data, give very similar results.
Analysis of model weights is a popular approach to investigation of the neural networks and
\citeauthor{perez2021characterizing} is one of many that employs this methodology. 
Another work that follows this idea is \citet{watanabe2022topological}, where authors focused on analyzing persistent diagrams, particularly times of birth and death.
They conducted their research on CNNs, yet still calculated Vietoris-Rips Complexes from MLP weights at the classification head of the network.
This is another work that demonstrates the potential of TDA methods in the investigation of neural networks.

The articles discussed above concentrate on checking what information we can obtain from neural representations or model weights.
This thesis explores a similar line of research.
Nevertheless, there are multiple other studies using TDA methods in more application-specific scenarios.
As an illustration, consider the work of \cite{rieck2018neural}, in which the authors develop a complexity measure for neural network architectures, called neural persistence.
The technique uses model weights and persistence homology to provide reliable information on neural network structure.
One of the results of using the aforementioned complexity measure is the difference in the proposed metric between models trained with and without dropout.
The difference was much larger than the one between models with and without batch normalization.
Such experiments can provide insights into the benefits of different network designs.
The authors also propose an early stopping method, which achieves results similar to those of the classic one.
Still, it works without needing a validation dataset, which can be useful when the amount of data is limited and obtaining a larger dataset is troublesome.

An early-stopping method is not the only use for TDA methods.
An algorithm for neural network pruning was proposed by \citet{watanabe2020deep}.
Their persistent-homology-based pruning method (PHPM) uses persistent homology calculated from model weights.
It tends to outperform the global magnitude pruning algorithm, a standard method of pruning neural networks.
Another application of TDA methods is detecting artificially generated text.
In the era of large language models, the ability to detect text generated by such models is increasingly important.
A TDA-based method of detecting such text was proposed by \cite{kushnareva2021artificial}.
To detect artificially generated text, authors collected attention maps used by transformer models.
Next, they calculated topological information about these attention maps, such as Betti numbers and the mean lengths of persistence bars.
Later they trained a linear classifier on top of these topological features.
This approach outperformed other state-of-the-art detectors.
Additionally, their TDA method worked fine for previously unseen models, like GPT, which is not seen in other algorithms.

TDA methods were also used in the detection of adversarial examples.
In \citet{gebhart2017adversary}, TDA methods for detecting adversarial samples were introduced and tested on the MNIST dataset, achieving an accuracy of 98\%.
This technique consists of analyzing a graph obtained from network layers using persistent homology.
The last work we want to mention \citep{zheng2021topological} focused on the detection of trojaned neural networks.
The authors use persistence homology and show that in trojaned networks death times of topological features are different than in non-attacked models.
Their method exhibits greater accuracy in the detection of trojaned models than other popular algorithms.

To conclude, there is a substantial body of research that analyzes neural representations with TDA or uses some of the TDA methods in particular applications. 
However, previous works often focus on very simple architectures or, in contrast to this work, operate on model weights, instead of neural representations.
The drawback of such a weight-centric approach is its limitation in transferring methods to models with different architectures.
Also, the aforementioned works do not focus on the evolution of neural representations that occurs during the transformation of data by the model.
These two areas create a research gap that was addressed in this thesis.

\chapter{Summary}

This thesis employs TDA algorithms to examine neural representations in selected neural architectures for image classification tasks.
We use these algorithms to analyze neural representations in class-conditional settings, with an eye towards determining whether groups of classes have different homology.
Our study shows that classes from the same model tend to have similar topological characteristics, at least in bottleneck distance lenses, and form a single cluster.
This similarity also occurs between representations coming from the train and test datasets.
We also found many differences and similarities between classic CNNs, residual networks, and transformers.
With regard to the traits that occur in the majority of investigated models, an important observation is that most intriguing topology changes happen in deeper layers.
Additionally, we observe that different model architectures lead to different homology, and more similar architectures have more similar topological features.
Overall, we performed many experiments to ascertain the efficacy of persistent homology in characterizing neural representations derived from various stages of neural networks.
The methodologies utilized in this thesis can easily be applied to other neural architectures for computer vision, since they all operate on neural representations.

Experiments performed in this work are of two kings
The first one aims to examine the behavior of persistence homology in analysis of neural representations.
To this end, we performed two experiments.
One explores how the number of elements in neural representations impacts persistent homology.
We demonstrated that the use of varying numbers of dataset elements leads to significant differences in the results, as both the bottleneck distance and the number of homological features vary.
Consequently, this forces us to compare complexes that were created using the same number of inputs if possible.
The other experiment checks how important it is to sift out the outliers. 
While this initially seemed essential to obtain correct results, it turned out that removing outliers from the model representations had a negligible impact on persistent homology, and this step could be omitted.
The only architecture analyzed in this work where removing outliers can potentially change the results is ViT.
This is because we observed many more outliers with this architecture.
All this being said, using algorithms for outlier removal still seems correct, and we have used them in all experiments for all models.

In the second set of experiments, we focus on finding interesting features in various models and architectures.
Firstly, we calculate and analyze how persistence diagrams change in multiple layers and networks.
Our findings indicate that each homology in each architecture exhibits distinctive characteristics.
For instance, in the case of deeper models (e.g., ResNet50 and ViT), topological changes occur relatively slowly.
In ViT we also observe many specific trends, such as longer-living homologies.
Another experiment checks how similar a finetuned model is to a pre-trained one.
It offers some interesting insights into the mechanisms underlying finetuning.
We observe that first-layer representations from the investigated models create one cluster when their persistence homologies are visualized in 2D.
For all examined architectures this changes in deeper layers where the topology of fine-tuned models starts to depart from the pre-trained model.
This behavior suggests that finetuning affects deeper layers of the model.
Moreover, this experiment shows that neural representations coming from models trained in different tasks are topologically different.
In the last experiment, we checked where topology of neural representations changes most rapidly.
Our results show that topology changes differently in each architecture and that ViT is unique compared to other models.
However, we observe a number of commonalities across all models, including the observation that the change in persistent homology is typically largest in the final layers.

After performing the aforementioned experiments, we can identify avenues for further investigation: the topic has not yet been fully explored, and there are still areas that haven't been touched.
In the next steps, we can explore a broader collection of datasets, to eliminate the impact of the dataset on observed homological features.
Additionally, we can explore more architectures, particularly outside computer vision problems.
For example, we can explore architectures like GPT and diffusion models, as both of them are currently very popular, and are state-of-the-art models for multiple tasks.
Another possibility is to try to find differences between models with the same architecture, but different activation functions or normalization methods.
Yet another avenue for future research is using different TDA algorithms, like the Mapper algorithm mentioned in related work section.
We can also move beyond class-wise analysis as in this work we did not find classes that have significantly different homology or separate groups of classes.
Instead of class-wise computation, we can perform global analysis, which may have the potential to uncover further structure in neural representations. 

\backmatter 

 \bibliography{main}
\end{document}